%% file: main.tex
\RequirePackage[OT1]{fontenc}
\documentclass[journal]{IEEEtran}

\usepackage{epsfig}
\usepackage{multirow}
\usepackage{booktabs}
\usepackage{array}
\usepackage{tabularx}
\usepackage{subcaption}
\usepackage[group-separator={,}]{siunitx}

\usepackage[table]{xcolor}
\definecolor{road}{rgb}{.502,.251,.502}
\definecolor{sidewalk}{rgb}{.957,.137,.910}
\definecolor{building}{rgb}{.275,.275,.275}
\definecolor{wall}{rgb}{.4,.4,.612}
\definecolor{fence}{rgb}{.745,.6,.6}
\definecolor{pole}{rgb}{.6,.6,.6}
\definecolor{tlight}{rgb}{.980,.667,.118}
\definecolor{tsign}{rgb}{.863,.863,0}
\definecolor{vegetation}{rgb}{.420,.557,.137}
\definecolor{terrain}{rgb}{.596,.984,.596}
\definecolor{sky}{rgb}{0,.510,.706}
\definecolor{person}{rgb}{.863,.078,.235}
\definecolor{rider}{rgb}{1,0,0}
\definecolor{car}{rgb}{0,0,.557}
\definecolor{truck}{rgb}{0,0,.275}
\definecolor{bus}{rgb}{0,.235,.392}
\definecolor{train}{rgb}{0,.314,.392}
\definecolor{motorbike}{rgb}{0,0,.902}
\definecolor{bicycle}{rgb}{.467,.043,.125}
\definecolor{unlabelled}{rgb}{0,0,0}

\usepackage{arydshln}
\usepackage{cite}
\usepackage[resetlabels]{multibib}
\newcites{appendix}{References}

\usepackage{amsmath}
\usepackage{amssymb}
\usepackage{mathtools}
\usepackage{url}

\usepackage{xspace}
\makeatletter
\DeclareRobustCommand\onedot{\futurelet\@let@token\@onedot}
\def\@onedot{\ifx\@let@token.\else.\null\fi\xspace}

\def\eg{\emph{e.g}\onedot} 
\def\ie{\emph{i.e}\onedot}

\def\etal{\emph{et al}\onedot}
\makeatother

\DeclareMathOperator{\ASR}{ASR}
\DeclareMathOperator{\mASR}{mASR}
\DeclareMathOperator{\IoU}{IoU}

\begin{document}
\title{Road Scenes Segmentation Across Different Domains by Disentangling Latent Representations}

\author{Francesco Barbato,~\IEEEmembership{Student Member,~IEEE,}
		Umberto Michieli,~\IEEEmembership{Graduate Student Member,~IEEE,}\\
        Marco Toldo
        and~Pietro Zanuttigh,~\IEEEmembership{Member,~IEEE.}%
		\thanks{F. Barbato, U. Michieli, M. Toldo and P. Zanuttigh are with University of Padova. Corresponding e-mail: umberto.michieli@dei.unipd.it}%
        \thanks{Our work was in part supported by the Italian Ministry for Education (MIUR) under the “Departments of Excellence” initiative (Law 232/2016) and by the SID project “Semantic Segmentation in the Wild”}}%

\markboth{Journal of \LaTeX\ Class Files,~Vol.~14, No.~8, August~2015}%
{Shell \MakeLowercase{\textit{et al.}}: Bare Demo of IEEEtran.cls for IEEE Journals}

{\maketitle}

\begin{abstract}
Deep learning models obtain impressive accuracy in road scenes understanding, however they need a large quantity of labeled samples for their training. Additionally, such models do not generalize well to %
environments where the statistical properties of data do not perfectly match those of training scenes,
and this can be a significant problem for intelligent vehicles. Hence, domain adaptation approaches have been introduced to transfer knowledge acquired on a label-abundant source domain to a related label-scarce target domain. In this work, we design and carefully analyze multiple latent space-shaping regularization strategies that work together to reduce the domain shift. More in detail, we devise a feature clustering strategy to increase domain alignment, a feature perpendicularity constraint to space apart features belonging to different semantic classes, including those not present in the current batch, and a feature norm alignment strategy to separate active and inactive channels. In addition, we propose a novel evaluation %
metric to capture the relative performance %
of an adapted model with respect to supervised training. We validate our framework in driving scenarios, considering both synthetic-to-real and real-to-real adaptation, outperforming previous feature-level state-of-the-art methods on multiple road scenes benchmarks. %
\end{abstract}

\IEEEpeerreviewmaketitle

\input{sections/introduction}

\input{sections/related}

\input{sections/problem_setup}
\input{sections/method}

\input{sections/implementation_details}
\input{sections/mASR}
\input{sections/results}

\input{sections/ablation}

\input{sections/conclusions}

\bibliographystyle{IEEEtran}
\bibliography{strings,refs}
\vspace{-1cm}
\begin{IEEEbiography}[{\includegraphics[width=1in,height=1in,clip,keepaspectratio]{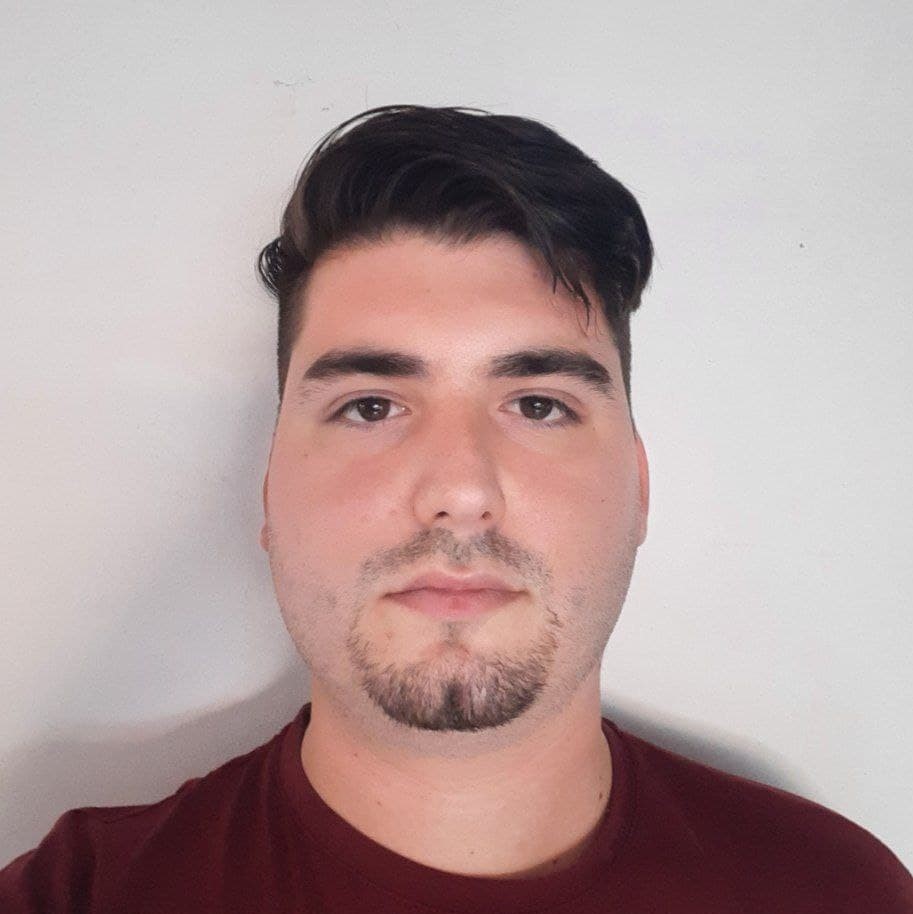}}]{Francesco Barbato}
received the M.Sc.\ degree in Telecommunication Engineering from the University of Padova in 2020. He is currently a Ph.D. Student the same University. His research focuses on domain adaptation and continual learning applied to computer vision tasks, particularly to semantic segmentation for autonomous vehicles.
\end{IEEEbiography}
\vspace{-1cm}
\begin{IEEEbiography}[{\includegraphics[width=1in,height=1in,clip,keepaspectratio]{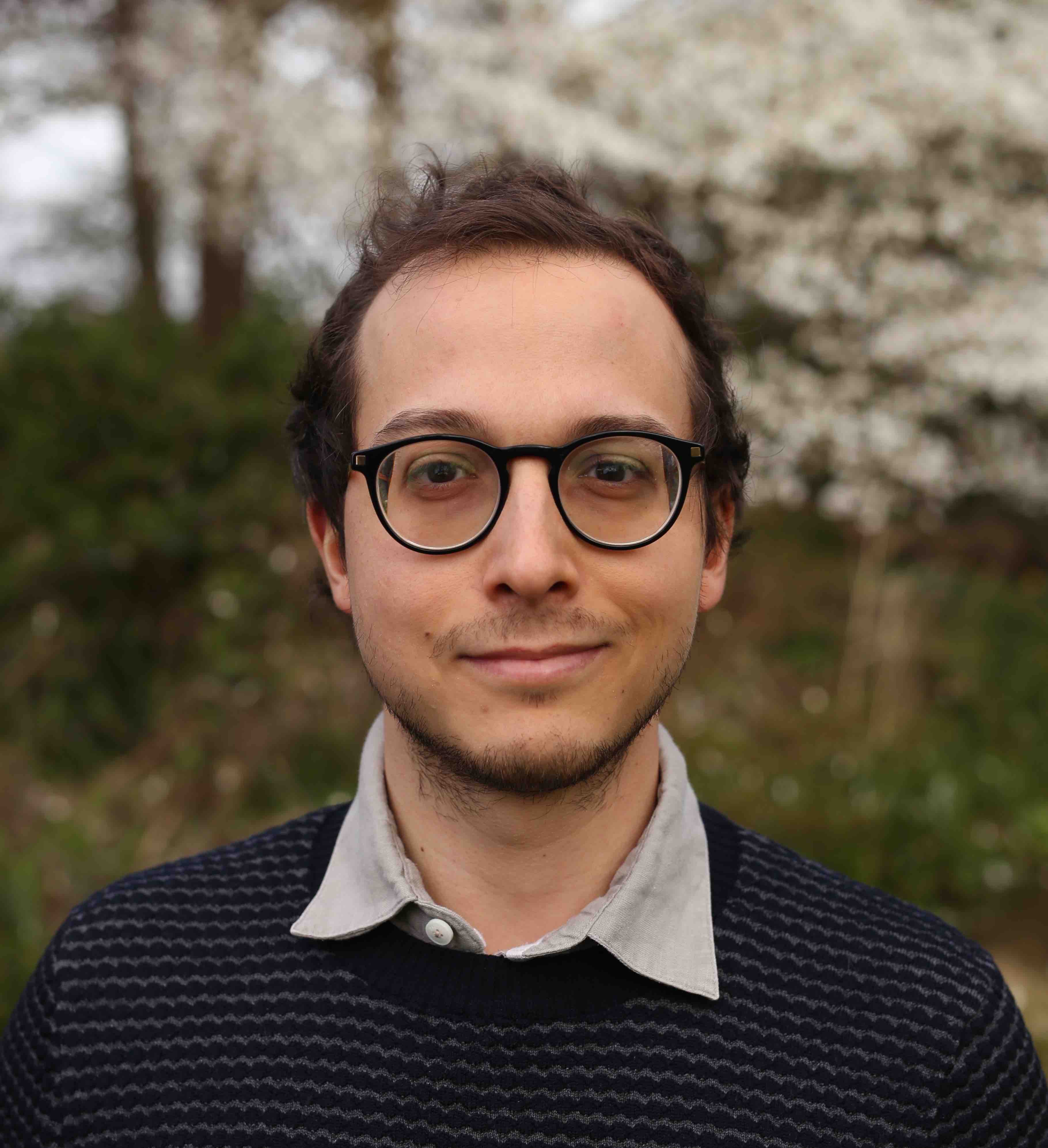}}]{Umberto Michieli}
received the M.Sc.\ degree in Telecommunication Engineering from the University of Padova in 2018. He is currently a final-year Ph.D.\ student at the same University. In 2018, he spent 6 months as a Visiting Researcher with the Technische Universit\"at Dresden. In 2020 he interned as Research Engineer for 8 months at Samsung Research UK. His research focuses on transfer learning techniques for semantic segmentation, in particular on domain adaptation and on incremental learning.
\end{IEEEbiography}
\vspace{-1cm}
\begin{IEEEbiography}[{\includegraphics[width=0.9\textwidth, trim=2cm 4cm 1.8cm 0, clip]{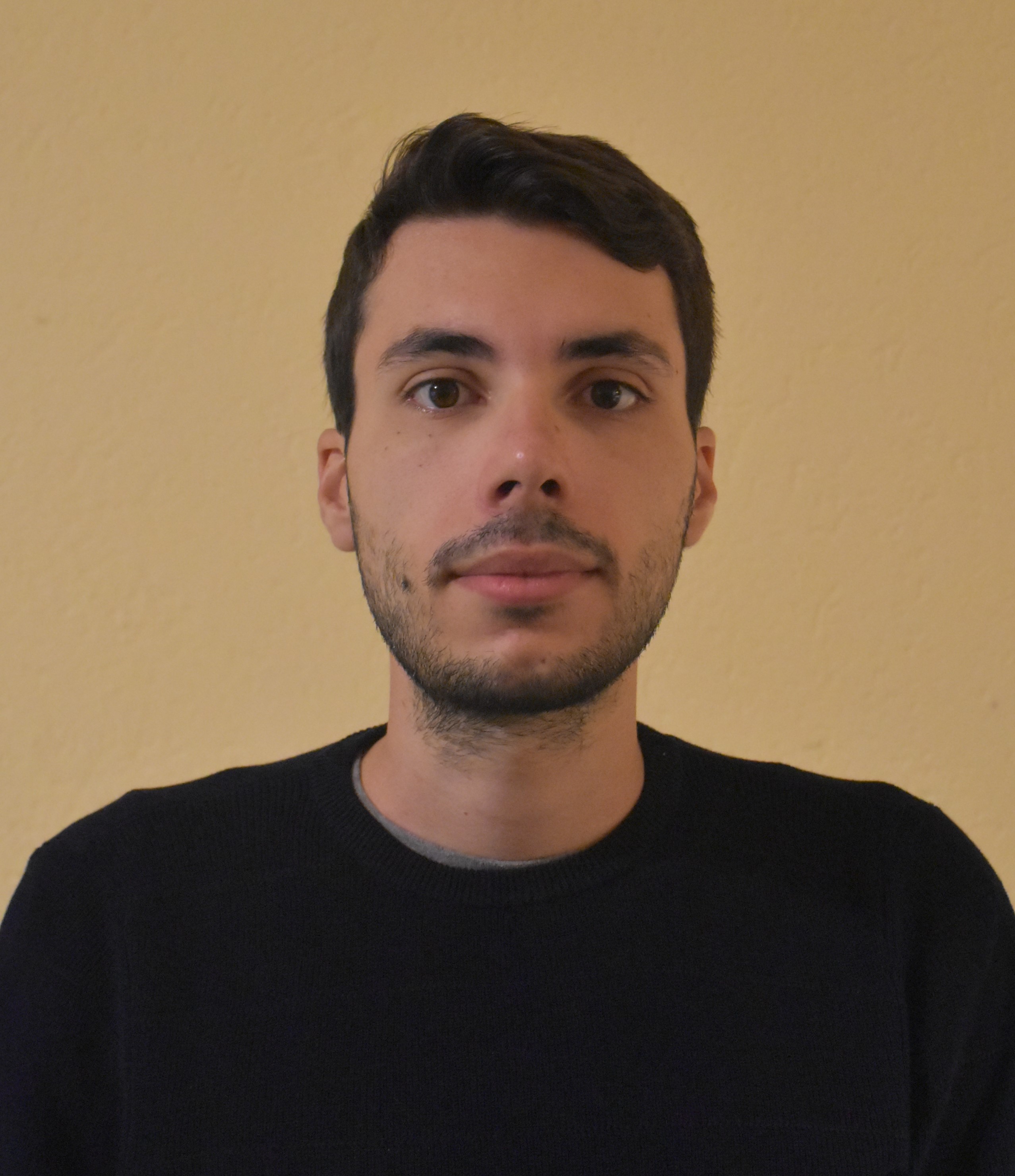}}]{Marco Toldo}
Marco Toldo received the M.Sc.\ degree in ICT for Internet and Multimedia in 2019 at the University of Padova. At present, he is doing his Ph.D. at the Department of Information Engineering of the same university. He is also doing an internship as Research Engineer at Samsung Research UK. His research involves domain adaptation and continual learning applied to computer vision.
\end{IEEEbiography}
\vspace{-1cm}
\begin{IEEEbiography}[{\includegraphics[width=1in,height=1in,clip,keepaspectratio]{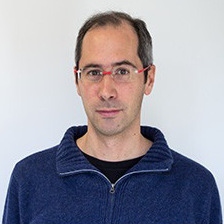}}]{Pietro Zanuttigh}
received a Master degree in Computer Engineering at the University of Padova in 2003 where he also got the Ph.D. degree in 2007. Currently he is an associate professor at the Department of Information Engineering. He works in the computer vision field, with a special  focus on domain adaptation and incremental learning in semantic segmentation, 3D acquisition with ToF sensors, depth data processing, sensor fusion and hand gesture recognition.
\end{IEEEbiography}

\clearpage

\setcounter{equation}{0}
\setcounter{figure}{0}
\setcounter{table}{0}
\setcounter{section}{0}
\setcounter{page}{1}
\renewcommand{\thesection}{S\arabic{section}}
\renewcommand{\theequation}{S\arabic{equation}}
\renewcommand{\thefigure}{S\arabic{figure}}
\renewcommand{\thetable}{S\arabic{table}}

\title{Supplementary Materials for:\\
Road Scenes Segmentation Across Different Domains by Disentangling Latent Representations}
{\maketitle}

\input{sections/appendixA}

\bibliographystyleappendix{IEEEtran}
\bibliographyappendix{strings,auxrefs}

\end{document}

%% file: sections/introduction.tex
\section{Introduction}
\label{sec:intro}

One of the key components of a self-driving vehicle is the capability of understanding the surrounding environment from sensory input data. Semantic segmentation enables profound scene understanding, where all pixels of the input images are assigned to a semantic category corresponding to key elements to be detected, such as the road, other vehicles or traffic lights and signs.
Nowadays, such task is commonly tackled with Deep Convolutional Neural Networks (DCNNs), which have achieved outstanding
results in image understanding tasks,
provided that a sufficiently large number of labeled examples are available from the target input domain distribution.
On the other side, the annotation of thousands of images of road scenes is highly expensive, time-consuming, error-prone and, possibly, worthless, since  the test data can show a domain shift with respect to the training labeled samples. Hence, recently, a strong requirement emerged for research and development of autonomous driving systems: namely, of being able to train DCNNs with a combination of labeled source samples (\eg, synthetic from ad-hoc simulators or driving video games)  and unlabeled target samples (\eg, real-world acquisitions from cameras mounted on cars), with the aim of getting high performance on data following the target distribution. The need for large quantities of labeled target data is superseded by data coming from a source domain where samples are abundantly available and annotations are faster and cheaper to generate. %

Unfortunately, DCNNs are prone to failure when they are shown an input domain distribution other than the training one (\textit{domain shift} phenomenon). In order to deal with this problem, various Unsupervised Domain Adaptation (UDA) techniques have been developed to adapt networks at different stages (the most common are the input, feature and output levels) ~\cite{toldo2020unsupervised}.%

Deep learning models for semantic segmentation are mostly based on encoder-decoder architectures, \ie, they build some concise latent representations of the inputs, which are highly correlated with the classifier output. As such, they are used in the subsequent classification process~\cite{bengio2013representation,girshick2014rich} that reconstructs the full resolution segmentation map. Nevertheless, a smaller number of UDA techniques for semantic segmentation work in the feature space because of its large dimensionality.
In this paper, we propose one such approach, comprised of a new set of strategies working at the latent space level, building on top of our previous conference work~\cite{barbato2021latent}. By employing a shaping objective in such place, our aims are to promote class-aware features extraction and features invariance between source and target domains. We remark that, while the general target behind each strategy is similar to the one of~\cite{barbato2021latent}, their actual implementation was overhauled, resulting in a significant performance increase.

Firstly, a clustering constraint groups feature vectors of each class tightly around their prototypical representation. 
Secondly, a perpendicularity objective over the class prototypes promotes disjoint filter activation sets across different semantic categories.  
Finally, a regularization-based norm alignment objective enforces consistent vector norms in the source and target domains, while jointly forcing progressively increasing norm values. This, in combination with the perpendicularity constraint, is able to reduce the entropy associated with the feature vector channel activations.

Importantly, the proposed techniques require the generation of accurate class prototypes and the imposition of a strong correlation between feature representations and predicted segmentation maps. Hence, we also propose a novel strategy to map semantic information from the labeling maps to the low resolution feature space (annotations downsampling). 

This paper moves from our previous work~\cite{barbato2021latent}, which %
already achieved state-of-the-art results on feature-level UDA in semantic segmentation. 
Compared to the conference version, this journal extension introduces several novel contributions.

First of all, the computation of prototypes and feature vector extraction have been refined. The first now considers the prototype trajectory evolution for a better estimation (Sec. \ref{sec:setup}-B), while the second exploits target information to reduce the domain shift (Sec. \ref{sec:setup}-C), additionally a class-weighting scheme is used in the source supervision %
(Sec. \ref{sec:setup}-A).

Then, each of the three proposed space-shaping constraints has been improved and additional ablation studies are shown both for our approach, LSR$^+$ %
(Sec. \ref{sec:ablation}), and for the proposed evaluation metric, mASR (Sec. \ref{sec:masr}). In particular, the clustering objective was modified to be more resilient to outliers (Sec. \ref{loss:clust}); the perpendicularity constraint now accounts for classes not present in the current batch (Sec. \ref{loss:perp}); the norm alignment now ignores low-activated channels %
(Sec. \ref{loss:norm}).

Finally, extensive experiments have been conducted on many road scenarios, expanding the set of experiments reported in~\cite{barbato2021latent}. The results are evaluated on $4$ backbones and $6$ setups. These include not only $2$ synthetic-to-real ones, commonly used in related works, but also  $4$ real-to-real settings addressing the critical issue of generalizing of autonomous driving systems across different cities and types of roads in different regions of the world. %
Additional results using the unlabeled \textit{Cityscapes} coarse set \cite{Cordts2016} are reported, showing significant performance gains when more unlabeled data are used (see Table \ref{table:results}).

%% file: sections/related.tex
\section{Related Works}
\label{sec:related}

\textbf{Semantic Segmentation of Road Scenes} is a very active research field. Semantic segmentation architectures like FCN \cite{long2015}, U-Net~\cite{ronneberger2015u}, PSPNet \cite{zhao2017}, and DeepLab \cite{chen2017rethinking,chen2018deeplab,chen2018encoder} have been recently applied to road scenes for two interconnected motivations: first, there is a large interest into the target application of self-driving vehicles and, second, there is the availability of large datasets (Cityscapes \cite{Cordts2016}, Mapillary \cite{neuhold2017mapillary}, IDD \cite{varma2019idd}, Cross-City \cite{chen2017nomore}, CamVid \cite{brostow2009semantic}) and simulators (CARLA \cite{Dosovitskiy17}, GTA5 \cite{Richter2016}, SYNTHIA \cite{ros2016}) that can be employed to train deep learning architectures.
 
\textbf{Unsupervised Domain Adaptation} consists in transferring knowledge extracted from a label-rich source domain to a completely unlabeled target domain.
The ultimate objective is to address the performance decline caused by domain shift, which %
negatively affects the generalization capabilities of deep neural networks.
The problem was initially studied for the classification task, but recently many works dealt with the unsupervised adaptation problem in relation to semantic segmentation. 
Although several methods have been proposed to tackle the adaptation task, they all share an underlying search for a form of domain distribution alignment over some representation space.
Some methods pursue distribution matching inside the input image space via style transfer or image generation techniques, others aims at bridging the statistical gap between source and target representations produced by the task model, whether manipulating some output representations, or operating inside a latent feature space \cite{toldo2020unsupervised}. 

Input-space adaptation has been commonly addressed resorting to image-to-image translation \cite{chen2019crdoco,hoffman2018, hoffman2016,MurezKKRK18,toldo2020,pizzati2020domain}. 
By transferring visual attributes across source and target samples, domain invariance is achieved in terms of visual appearance. 
Source supervision can thus be safely exploited in the shared image space, retaining consistent accuracy on source and target data. 
 
As concerns feature and output-space adaptation, adversarial learning has been largely employed to bridge the statistical domain gap \cite{du2019ssfdan, sankaranarayanan2018,tsai2018,Tsai2019, biasetton2019, michieli2020adversarial,  spadotto2020unsupervised}.
With the help of a domain discriminator, the task network is forced to provide statistically indistinguishable source and target representations, typically drawn from a latent feature space \cite{du2019ssfdan, sankaranarayanan2018,tsai2018} or in the form of probability maps at the output of the segmentation pipeline \cite{tsai2018, Tsai2019, biasetton2019, michieli2020adversarial,  spadotto2020unsupervised}. 
More recently, some works focusing on feature-level regularization have been proposed \cite{toldo2020clustering,barbato2021latent}.
In \cite{toldo2020clustering} a class-conditional domain alignment is achieved by means of a discriminative clustering module, paired with orthogonality constraints to enhance class separability. 
The approach of \cite{barbato2021latent}  relies on conditional clustering adaptation, enhanced by a perpendicularity objective over class prototypical representations and a novel norm alignment loss to improve class separability at the latent space. 
As an alternative form of feature-level adaptation, dropout regularization has been explored~\cite{Lee2019,Park2018,Saito2018ADR}; decision boundaries are pushed away from target high density regions in the latent space without direct supervision.

Output-space adaptation has been further pursued resorting to self-training~\cite{zou2018,Zou2019}, where the learning process is guided (in a self-supervised manner) by pseudo-labels extracted from target network predictions. 
Self-supervision has been proposed in a curriculum learning fashion as well \cite{zhang2017,zhang2020curriculum}. First, simple tasks that are less sensitive to domain shift are solved, by inferring some useful properties related to the target domain. 
Then, the extracted information is exploited to address more complex learning tasks (\eg, semantic segmentation).
Alternatively, some works introduce entropy minimization techniques~\cite{Chen2019,vu2019advent}, which force more confident network predictions over target data, thus encouraging the behavior shown in the supervised source domain.

\textbf{Latent Space Regularization} has been shown to ease the semantic segmentation tasks in different settings, such as UDA \cite{kang2019contrastive,tian2020domain}, continual learning \cite{michieli2021continual} and few-shot learning \cite{dong2018few,wang2019panet}.
The idea is to embed additional constraints on feature representations during the training process, enforcing a regular semantic structure on latent spaces of the deep neural classifier.
In UDA, where target semantic supervision is missing, regularization can be applied in class-conditional manner by relying on the exclusive supervision of source samples, while indirectly propagating its effect to target representations as well.
Such improved regularity has, in fact, shown to promote generalization properties, leading to statistical alignment between the source and target distributions when regularization is jointly applied over both domains \cite{toldo2020clustering, barbato2021latent}.

A multitude of feature clustering techniques based on the K-Means algorithm have been proposed \cite{kang2019contrastive,liang2019distant,wang2019unsupervised,tian2020domain} to address the adaptation task.
Those works are mainly focused on image classification and resort to a projection to a more easily-tractable lower-dimensional latent space where to perform pseudo labeling of the original target representations extracted by the task model \cite{liang2019distant, wang2019unsupervised, tian2020domain}.
In \cite{toldo2020clustering, barbato2021latent} the idea is further refined and applied to semantic segmentation by proposing an explicit clustering objective paired with orthogonality constraints to force feature vectors to cluster around the respective class prototypes. 
Feature-level orthogonality has been also explored in \cite{choi2020role} to limit the redundancy of the information encoded in feature representations. Approaches closer to our strategy are \cite{pinheiro2018unsupervised,wu2019improving}, where UDA is promoted via an orthogonality objective over class prototypes. Nonetheless, \cite{choi2020role,pinheiro2018unsupervised,wu2019improving} all limit their focus to the image classification task.

%% file: sections/problem_setup.tex
\section{Problem Setting}
\label{sec:setup}
In this section we overview our setup, detailing the mathematical notation used throughout the paper.
We start by identifying the input space as ${\mathcal{X} \! \subset \! \mathbb{R}^{H \times W \times 3}}$ and the corresponding label space as ${\mathcal{Y} \! \subset \! \mathcal{C}^{H \times W}}$, where $H$ and $W$ represent the image resolution and $\mathcal{C}$ the class-set. 
Furthermore, we assume to have a training set ${\mathcal{T} = \mathcal{T}^s \bigcup \mathcal{T}^t}$, where ${ \mathcal{T}^s = \{ (\mathbf{X}_{n}^s,  \mathbf{Y}_{n}^s) \}_{n=1}^{N_s} }$ contains labeled samples ${ (\mathbf{X}_{n}^s,  \mathbf{Y}_{n}^s) \in \mathcal{X}^s \times \mathcal{Y}^s }$ originated from a source domain, while an additional set of input samples ${ \mathcal{T}^t = \{ \mathbf{X}_{n}^t \}_{n=1}^{N_t} }$ is drawn from an unlabelled target domain ($\mathbf{X}_{n}^t \in \mathcal{X}^t$).
We adapt the knowledge of semantic segmentation learned on the source domain to the unsupervised target domain. Superscript $s$ identifies the source domain, while $t$ the target.

As done by most recent approaches for semantic segmentation, we assume a task model $S = D \circ E$  based on an encoder-decoder architecture, that is, made by the consecutive application of an encoder network $E$ (referred to as backbone, which acts as feature extractor) and a decoder network $D$, which actually performs the classification and produces the segmentation map. 
We denote the features extracted from an input image $\mathbf{X}$ as $E(\mathbf{X}) = \mathbf{F} \in \mathbb{R}^{H' \times W' \times K}_{0+}$, where $K$ refers to the number of channels and $H' \times W'$ to the low-dimensional (feature-level) spatial resolution.
Thanks to the topology of encoder-decoder DCNNs for semantic segmentation, classes are encoded into ideal latent representations, invariant with respect to the domain shift.
The strategies presented in Sec.~\ref {sec:method} enforce this goal by comparing the extracted features belonging to each class with the respective prototypical representations. %
In the following paragraphs, we present the techniques used to compute the prototypes and associate feature vectors to semantic classes.

\textbf{A.~~Weighted Histogram-Aware Downsampling.} 
Given that most of the spatial information of an image is maintained while it it processed by an encoder-decoder network, we can assume a tight relationship between any feature vector (\ie, the vector of features associated to a single spatial location within the feature tensor) and the semantic labeling of the corresponding image region. %

Hence, the extraction process begins with the identification of a way to propagate the labeling information to %
latent representations %
(decimation),
without losing the semantic content of the window (image region) corresponding to each feature vector. %
A naïve approach, which allows wrong mappings, would strongly affect the whole following procedure. 
Our solution is a non-linear pooling function, which instead of computing a simple subsampling %
(\eg, nearest neighbor%
) extracts a weighted frequency histogram over the labels of all the pixels in the window corresponding to a low-resolution feature location. The weights are inversely proportional to the class-frequency in the source training dataset.
Then, these metrics are used to select the most appropriate class for each image region, producing source feature-level label maps $\{\mathbf{I}^{s}_n\}^{N_{s}}_{n=1}$. The computation of the target counterparts ($\{\mathbf{I}^{t}_n\}^{N_{t}}_{n=1}$) is discussed in Sec.~\ref{sec:setup}-C and we remark that each $ \mathbf{I}^{s,t}_n \in \mathcal{C}^{H' \times W'}$.
In particular, we choose the label with the highest frequency peak in the windows, only if such peak is relevant enough, \ie, if all other peaks are smaller than $T_h$ times it (%
a similar approach is found in the orientation assignment step of the SIFT feature extractor \cite{790410}). Empirically, we set $T_h=0.5$.
Finally, we remark a useful side-effect of this technique: whenever a window cannot be uniquely assigned to a class (that is, it contains multiple labels) the procedure automatically assigns it to the \textit{void} class.

\textbf{B.~~Prototype Extraction.} The feature-level label maps ${\{\mathbf{I}^{s,t}_n\}^{N_{s,t}}_{n=1}}$ allow to identify the set $\mathcal{F}^{s,t}_c$ of feature vectors belonging to class $c \in \mathcal{C}$ in training batch $\mathcal{B}$:
\begin{equation}
\label{eq:feats}
\mathcal{F}^{s,t}_c \! = \! \left\{ \mathbf{F}_n^{s,t}[h,w] \in \mathbb{R}^{K}_{0^+} \mid \mathbf{I}^{s,t}_n[h,w] = c, \forall n \in \mathcal{B}\right\} ,
\end{equation}
where the couple $[h,w]$ denotes the spatial location ($0 \leq h < H'$  and $0 \leq w < W'$). 
The definition is further expanded into the set of all feature vectors in batch $\mathcal{B}$ by taking their union with the set $\mathcal{F}^{s,t}_v$ of samples belonging to class \textit{void}: $\mathcal{F}^{s,t} = ( \bigcup_{c} \mathcal{F}^{s,t}_c ) \cup\mathcal{F}^{s,t}_v$.
From these sets we can extract the batch-wise prototypes of each class (note that we use feature vectors exclusively from the source):
\begin{equation}
\label{eq:proto}
\mathbf{p}_c[i] = \frac{1}{|\mathcal{F}^s_c|}\sum\limits_{\mathbf{f} \in \mathcal{F}^s_c}{\mathbf{f}[i]} \;\;\;\;\; \forall i,\;\; 1 \leq i \leq K .
\end{equation}
Moreover, with the goal of obtaining more stable and reliable prototypes, and to reduce estimation noise, we consider the exponentially smoothed vectors: %
\begin{equation}
\label{eq:smooth}
\hat{\mathbf{p}}_c = \eta \mathbf{p}_c + (1-\eta)\hat{\mathbf{p}}_c' .
\end{equation}
The parameters are initialized with $\hat{\mathbf{p}}_c=\mathbf{0}$ and $\eta=0.8$ (empirically). In our notation, $\hat{\mathbf{p}}_c'$ represents the estimate at the previous optimization step, while $\mathbf{p}_c$ the one of the current. 
This way, by setting $\eta<1$, we can propagate the previous estimates to the current batch, allowing to consider classes absent from $\mathbf{Y}^{s}_n$ in the loss computation. %

\begin{figure}[t]
\centering
\includegraphics[width=0.45\textwidth]{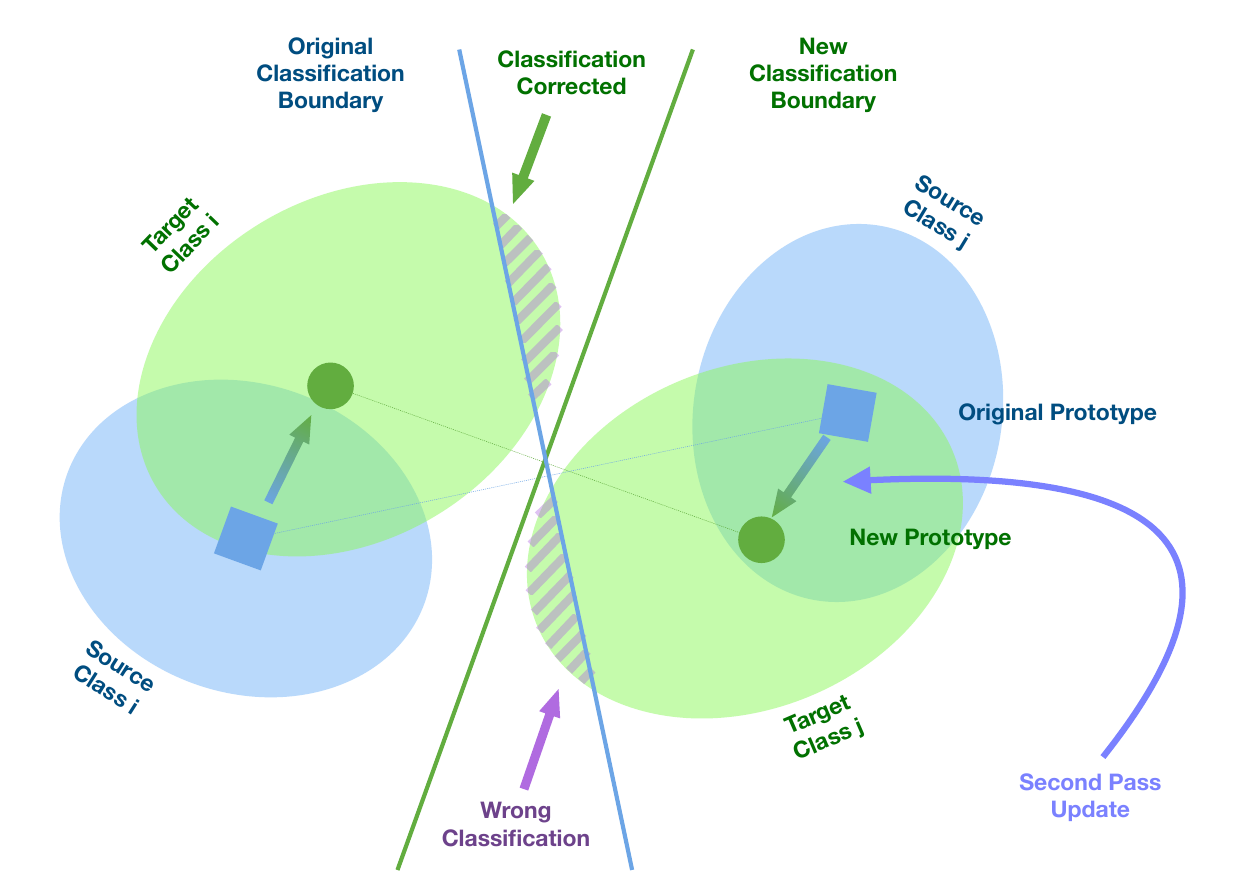}
\caption{Visual representation of our two-pass feature vector classification strategy. The initial source-based classification (in blue) can lead to erroneously classified target samples (purple shaded areas). This problem is tackled by computing target prototypes as the centroids of the partitioned vectors (notice the shift compared to the original source prototype), these prototypes are used as new classification centers (green boundary), producing a correct segmentation.}
\label{fig:doubleproto}
\end{figure}

\textbf{C.~~Feature pseudo-labeling.}
\label{subsec:twopass}
While the histogram strategy can be seamlessly extended to be used with pseudo-labels (\ie, network estimates for the unlabeled target samples, as was our strategy in the previous work \cite{barbato2021latent}), this approach can introduce instability in the training procedure. To avoid such issue, we devise a novel way of extracting the target feature-level label maps $\{\mathbf{I}^{t}_n\}^{N_{t}}_{n=1}$.\\
Our strategy exploits the euclidean distance in the latent space, computing a clustering of the feature vectors around their prototype (see Fig. \ref{fig:doubleproto}).
More in detail, we compute an initial classification exploiting the  prototypes computed over the source labeled data, 
which, due to the domain shift, will not be adequately representative of the target distribution: \\
\begin{equation}
\begin{aligned}
\mathcal{\widetilde{F}}_{c}^{t} &= \{\mathbf{F}_n^t[h,w] \;\; \text{if} \\
& \hspace{1.5cm} \sigma_c(-|| \mathbf{F}_n^t[h,w]-\mathbf{\hat{p}}_c ||) > 0.5 \; \forall n \in \mathcal{B} \} \\
\mathbf{p}^t_c[i] &= \frac{1}{|\mathcal{\widetilde{F}}_{c}^{t}|}\sum\limits_{\mathbf{f} \in \mathcal{\widetilde{F}}_{c}^{t}}{\mathbf{f}[i]} \;\;\;\;\; \forall i,\;\; 1 \leq i \leq K .
\end{aligned}
\end{equation}
Where $\sigma_c(\cdot)$ is the softmax function computed over the classes.
Then, we refine the classification keeping only those vectors that have a high classification confidence according to a probability distribution attained through a softmax function:
\begin{equation}
\begin{aligned}
\mathbf{I}^t_n[h,w] &= \begin{cases} c & \sigma_c(-|| \mathbf{F}_n^t[h,w]-\mathbf{p}^t_c ||) > 0.5 \\
								\textit{void}  & \text{otherwise}
								\end{cases}
\end{aligned}
\end{equation}

%% file: sections/method.tex
\section{Methodology}
\label{sec:method}

\begin{figure*}[t]
\centering
\includegraphics[width=0.95\textwidth]{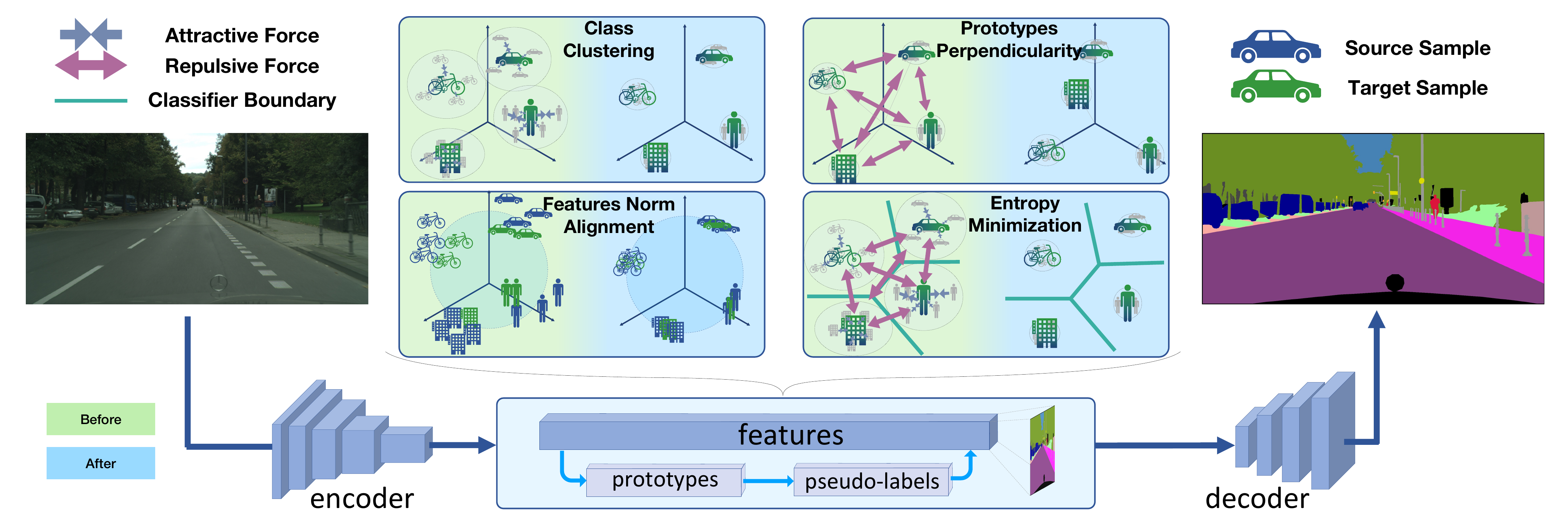}
\caption{Visual summary of our strategy. %
 Features are associated to semantic classes and prototypes are computed from them (\ref{sec:setup}). Class clustering (\ref{loss:clust}),  prototypes perpendicularity (\ref{loss:perp}) and norm alignment and enhancement (\ref{loss:norm}) are the three proposed space-shaping constraints. Additonally, we apply on top an entropy minimization objective \cite{Chen2019}.}
\label{fig:architecture}
\end{figure*}

The proposed approach is detailed in this section, highlighting the key differences with respect to our previous work. %
Our investigation moves from the fact that the discriminative effect acquired by the model with the source supervised cross-entropy objective may not be propagated to the target domain due to the distribution shift.
To tackle such problem, in~\cite{barbato2021latent} we proposed to use additional space-shaping objectives to increase the network generalization capability, therefore improving %
robustness to distribution shifts from the original source training data.
In particular, we added three feature-space shaping constraints
to the standard source-supervision ($\mathcal{L}_{CE}^s$), whose combined effect can be  expressed as:
\begin{equation}
\mathcal{L} = \mathcal{L}_{CE}^{s} + \lambda_C \cdot \mathcal{L}_C^{s,t} + \lambda_P \cdot \mathcal{L}_P^{s} + \lambda_N \cdot  \mathcal{L}_N^{s,t} .
\label{eq:tot}
\end{equation}
Here, $\mathcal{L}_C$ represents the clustering objective acting on the feature vectors (Sec.~\ref{loss:clust}), $\mathcal{L}_P$ the perpendicularity constraint applied to class prototypes (Sec.~\ref{loss:perp}) and $\mathcal{L}_N$ the norm alignment goal (Sec.~\ref{loss:norm}). %
To simplify the notation, Eq.\ (\ref{eq:tot}) contains each loss component with $s,t$ superscripts to indicate the sum of the loss on source and target samples.
To further improve the performance and to show how the proposed techniques  can be used on top of existing strategies, we also add to the optimization target  the entropy minimization loss introduced by Chen \etal \cite{Chen2019}, obtaining: $\mathcal{L}^+ = \mathcal{L} + \lambda_{EM} \cdot \mathcal{L}_{EM}$. By doing so, we also show that our space-shaping objectives provide a different and complementary %
effect on the feature vectors when compared to the entropy minimization constraint. 
An overview of the proposed strategy is presented in Fig.~\ref{fig:architecture}.

\subsection{Clustering of Latent Representations}
\label{loss:clust}
Due to the distribution discrepancy between source and target domains,
feature vectors originating from them will be misaligned. %
This inevitably causes some incorrect classifications of target representations, in turn degrading the segmentation accuracy in the target domain.
We introduce our first loss, a clustering objective over the latent space, to mitigate this problem, seeking for class-conditional alignment of feature distribution. %
We do so by exploiting the prototypical representations discussed in Sec.~\ref{sec:setup} %
and forcing the feature vectors from source and target representations to tightly cluster around %
them: representations are adapted into a common class-wise distribution and the disciminativeness of the latent space is increased.

Differently from the previous work, we define the clustering objective as the L1 distance between feature vectors and their associated class-prototypes. %
This results in a more stable training evolution and lower error rate in clustering, thanks to the outlier-rejecting properties of the L1 norm. In particular, due to the quadratic nature of the L2 loss, outliers with distances greater than $1$ have a strong push towards the clusters. 
On the other hand, the L1 loss is stronger than L2 for close samples, which are more representative of each class, and is significantly gentler than L2 for distant outliers.
The loss can be expressed mathematically as:
\begin{equation}
 \mathcal{L}_C^{s,t} = 
 \frac{1}{|\mathcal{C}|} \sum\limits_{c \in \mathcal{C}} 
 \frac{1}{|\mathcal{F}^{s,t}_c|} 
 \sum\limits_{\mathbf{f} \in \mathcal{F}^{s,t}_c}
\frac{1}{K}\sum_{k=1}^{K}|\hat{\mathbf{p}}_c[k]-\mathbf{f}[k]|,
\end{equation}%
This loss has multiple targets: the first is the increased clustering of the latent representations thanks to label supervision, which reduces the tendency to erroneous predictions.
The second one is to perform self-supervised clustering on target samples using our two-pass pseudo-labeling strategy (see Sec.~\ref{sec:setup}-C). %
Finally, it leads to better prototype estimates, due to the fact that forcing tighter clusters will lead to more stable %
batch-wise centroids, which in turn will get closer to the moving-averaged prototypes.

\subsection{Perpendicularity of Latent Representations}
\label{loss:perp}

A prototype perpendicularity loss is further proposed to aid the latent space regularization brought by the clustering objective. 
Our goal is to induce compact and domain-aligned feature clusters, in order to boost the accuracy of network segmentation maps.
As a direct consequence, the margin between classification boundaries and feature clusters is expanded, thus decreasing the probability that target high-density regions are traversed by such boundaries.
We directly encourage a class-wise orthogonality property, not only increasing the distance among class clusters, but also reducing class cross-talk by discouraging shared channels activations in distinct categories.

In the loss, we encode the perpendicularity score exploiting the definition of euclidean space inner product: ${\mathbf{j}\cdot\mathbf{k} \! = \! ||\mathbf{j}||\;||\mathbf{k}|| \!\cos\theta}$, where $\theta$ is the angle internal to the two vectors $\mathbf{j}$ and $\mathbf{k}$.
To maximize $\theta$ we just need to minimize the vectors' normalized product (recall that $\mathbf{j}, \mathbf{k} \in \mathbb{R}^{K}_{0^+}$).
Therefore, the cross-perpendicularity between prototypes is encoded as:
\begin{equation}
\mathcal{L}_P^{s} = \frac{1}{|\mathcal{C}|(|\mathcal{C}|-1)}\sum\limits_{c_i,c_j \in \mathcal{C}, i \neq j}{\frac{\mathbf{\hat{p}}_{c_i}}{||\mathbf{\hat{p}}_{c_i}||}\cdot\frac{\mathbf{\hat{p}}_{c_j}}{||\mathbf{\hat{p}}_{c_j}||}}.
\label{eq:perp}
\end{equation}
Eq.~\eqref{eq:perp} computes the sum of the cosines over the set of all couples %
of non-void classes.
The influence of the orthogonality objective indirectly reaches all feature vectors, as prototypical representations and single feature instances share a strong geometric bound promoted by $\mathcal{L}_C^{s,t}$.
What we ultimately achieve is thus to enforce a perpendicularity constraint among instances of different clusters, with a homogeneous action over all latent representations from the same semantic class.
In other words, the angular gap among distinct semantic categories in the feature space is enlarged, by inducing disjoint patterns of activated feature channels between distinct classes.

In contrast to our previous paper \cite{barbato2021latent}, we compute the loss on the exponentially smoothed version of the prototypes, \ie, from Eq. (\ref{eq:smooth}). This guarantees that the space will be more evenly occupied by the classes, since all directions are considered in the computation of the loss, instead of considering only the ones in the current batch.

\subsection{Latent Norm Alignment Constraint}
\label{loss:norm}

This loss term is computed exploiting %
source and target feature vector norms.
More in detail, we enforce norm consistency between the latent representations extracted from the two domains.
This has two objectives:
firstly, we aim at an improved classification confidence on target predictions, as done by adaptation techniques using entropy minimization in the output space \cite{vu2019advent}.
Secondly, we assist the perpendicularity constraint by reducing the number of domain-specific feature channels used by the network for  classification. 
Thirdly, we reduce the number channels enabled only on one of the domains, which would lead to norm misalignment.
Moreover, to reduce the possible decrease in norm value during the alignment process, we introduce a regularization term that promotes norm increase.
Differently from \cite{barbato2021latent}, here the norm objective is encoded as a relative difference with a regularization term inversely proportional to the norm value. This allows to obtain a value-independent loss where norm values higher than the target are less discouraged. %
Moreover, we introduce a \textit{norm filtering} strategy to reduce the negative effects a careless increase in norm could imply. In particular, we suppress %
low channel activations, stopping the gradient flow through them and preventing the norm alignment procedure to increase their value, %
in contrast to what source supervision indicates.
Formally, we define the loss term as: %
\begin{equation}
\mathcal{L}_N^{s,t} \! =\! \frac{1}{|\mathcal{F}^{s,t}_{*}|} \sum\limits_{\mathbf{f} \in \mathcal{F}^{s,t}_{*} } \frac{{\left| (\bar{f}_s+\Delta_f) \! - \! ||\mathbf{f}|| \right|}}{\bar{f}_s},
\label{eq:norm}
\end{equation}
where $\bar{f}_s$ is the average source vector norm (extracted in the previous optimization step),
 $\Delta_f$ dictates the regularization strength (experimentally tuned to $0.1$) and $\mathcal{F}^{s,t}_{*}$ is a thresholded version of $\mathcal{F}^{s,t}$ where we set to $0$ the low-activated channels of each feature vector, stopping the gradient propagation:
\begin{equation}
\begin{aligned}
\mathcal{F}^{s,t}_{*} &= \{\phi(\mathbf{f}) \;\;\; \forall \mathbf{f} \in \mathcal{F}^{s,t}\}, \\
\phi(\mathbf{f})_i &= \begin{cases}
\mathbf{f}_i & \mathbf{f}_i \geq \frac{1}{K}\sum\limits_{j=1}^{K}{\mathbf{f}_i}, \\
0 & \text{otherwise.}
\end{cases}
\end{aligned}
\end{equation}
This objective is applied in a completely unsupervised manner, the vector norms are forced to align to the same value regardless of their class.
In this way we remove the bias generated by heterogeneous pixel-class distribution in semantic labels, that, for example, lead features of the most frequent classes to have larger norms than  average. %
The constraint of Eq.~(\ref{eq:norm}) forces the inter-class alignment step, 
\ie, it promotes %
gradual alignment of the norms towards a target common to all categories, while discouraging the value of such target to decrease too rapidly.
An additional benefit of rescaling the loss by the norm target is that the loss gradients will be limited in magnitude and, therefore, more stable.

%% file: sections/implementation_details.tex
\section{Implementation Details}
\label{sec:implementation}

\textbf{Training Data.} We evaluated our approach (LSR$^+$, Latent Space Regularization) on road scenes segmentation in various  synthetic-to-real and real-to-real unsupervised adaptation tasks. 
For the  source domains we used the synthetic datasets \textit{GTAV}~\cite{Richter2016} and \textit{SYNTHIA}~\cite{ros2016}.
The first contains $24,966$  labeled images at a resolution of $1914\times1052$ px, produced with the rendering engine of the \textit{GTAV} videogame, while the second contains $9,500$  labeled images at a resolution of $1280\times760$ px, rendered with a custom software.
For the target domain, we selected the real world dataset \textit{Cityscapes}~\cite{Cordts2016}. It contains $5,000$  labeled images at a resolution of $2048\times1024$ px and an additional set of $20,000$ coarsely labeled samples, acquired in European cities. When considering only unlabeled samples the two versions are equivalent and can be merged (obtaining a dataset we refer as \textit{CS-full}) increasing the adaptation process (as we show in Table~\ref{table:results}).
In the real-to-real setup we used the \textit{Cross-City} benchmark, where the \textit{Cityscapes} dataset takes the role of source domain, while the \textit{Cross-City} dataset~\cite{chen2017no} takes the role of target.
Such dataset is comprised of $12,800$ ($4\times3,200$) high resolution ($2048\times1024$ px) images taken in four major cities (\textit{Rome}, \textit{Rio}, \textit{Tokyo}, \textit{Taipei}).

We trained the model in a closed-set~\cite{toldo2020unsupervised} setting, \ie, with the same source and target class sets.
More in detail, we used the $19$, $16$ and $13$  classes in common for \textit{GTAV}, \textit{SYNTHIA} and \textit{Cross-City}, respectively.
\textit{GTAV}, \textit{Cityscapes} and \textit{Cross-City} images have been rescaled for training to $1280\times720$ px, $1280\times640$ px and $1280\times640$ px, respectively, while the resolution of \textit{SYNTHIA} images has not been changed. %

\textbf{Segmentation Network.} We employed the well-known~\cite{tsai2018, vu2019advent, Chen2019, tranheden2020dacs, toldo2020clustering} DeepLabV2 network \cite{chen2017rethinking,chen2018encoder,chen2018deeplab}, with ResNet101~\cite{he2016deep} as  backbone (using $2048$ channels in the last stage of the encoder) and a stride of $8$. %
We pre-train the network following the same procedure as our previous work~\cite{barbato2021latent},  employing the same data augmentation techniques used during adaptation.%

\textbf{Network Training.} We optimize the model with SGD (using a 
momentum of  $0.9$ and a weight decay regularization of $5 \times 10^{-4}$).
 The learning  rate starts from $2.5 \times 10^{-4}$ and uses a polynomial decay of power $0.9$  over $250k$ steps, as in~\cite{Chen2019}.
We used for validation a subset of the original training set to tune  the hyper-parameters of our loss components.
To tackle overfitting we used some data augmentation strategies: random left-right flipping; white point re-balancing $\propto \mathcal{U}([-75,75])$%
; color jittering $\propto \mathcal{U}([-25,25])$ (the last two applied independently  in the R,G and B channels) and random Gaussian blur~\cite{zou2018,Chen2019}.  
We perform training on a  NVIDIA Titan RTX, using a batch size of $2$ ($1$ source and $1$ target samples) for $24,750$ steps (\ie, $10$ epochs of the \textit{Cityscapes} train set). We also exploited the validation set for early stopping. %

The code developed for this work is publicly available at the following link: \url{https://github.com/LTTM/LSR}.

%% file: sections/mASR.tex
\section{Mean Adapted-to-Supervised Ratio metric}
\label{sec:masr}

In this section we introduce a novel measure, called mASR (\textit{mean Adapted-to-Supervised Ratio}), in order to better evaluate the domain adaptation task with respect to the usual mIoU. %

The idea behind the new metric sparks from realizing that the mIoU is missing a key component to evaluate  an adaptation method: \ie, it does not account for the starting accuracy on the different classes in supervised training.
In particular, the objective of domain adaptation is to transfer the knowledge learned on a source dataset to a target one, trying to get as close as possible to the results attainable through supervised learning on the target domain. %
We design mASR to capture the relative accuracy of an adapted architecture with respect to its target supervised counterpart, which we identify as a reasonable upper bound. %
Therefore mASR focuses less on the absolute-term performance and more on the relative accuracy obtained by an adapted architecture when compared to traditional supervised training. 

We compare the per-class IoU score of the adapted network for each  $c\in \mathcal{C}$ ($\IoU^c_{adapt}$) with the results of supervised training on target data ($\IoU^c_{sup}$) and we compute mASR by:
\begin{equation}
\mASR = \frac{1}{|\mathcal{C}|} \sum\limits_{c \in \mathcal{C}}{\ASR^c}, \;\;\; \ASR^c \overset{def}{=} \frac{\IoU^c_{adapt}}{\IoU^c_{sup}} \cdot 100 .
\end{equation}
In mASR, the contribution corresponding to each semantic category is inversely proportional to the accuracy  of the segmentation model on it in the supervised scenario, thus giving more relevance to the most challenging classes and producing a more class-agnostic adaptation score. Furthermore, notice how the most challenging classes in  driving scenarios are typically associated to small objects like traffic lights or pedestrians and bicycles, that are very critical for the autonomous navigation.
In this metric, higher means better and when the adapted network has the same performance as supervised training the score is $100\%$.

As an example, the mASR scores reported in the last two columns of Table~\ref{table:results} allow to identify at a glance the algorithms that more faithfully match the target performance. 

\begin{figure}[t]
\centering
\includegraphics[width=0.45\textwidth]{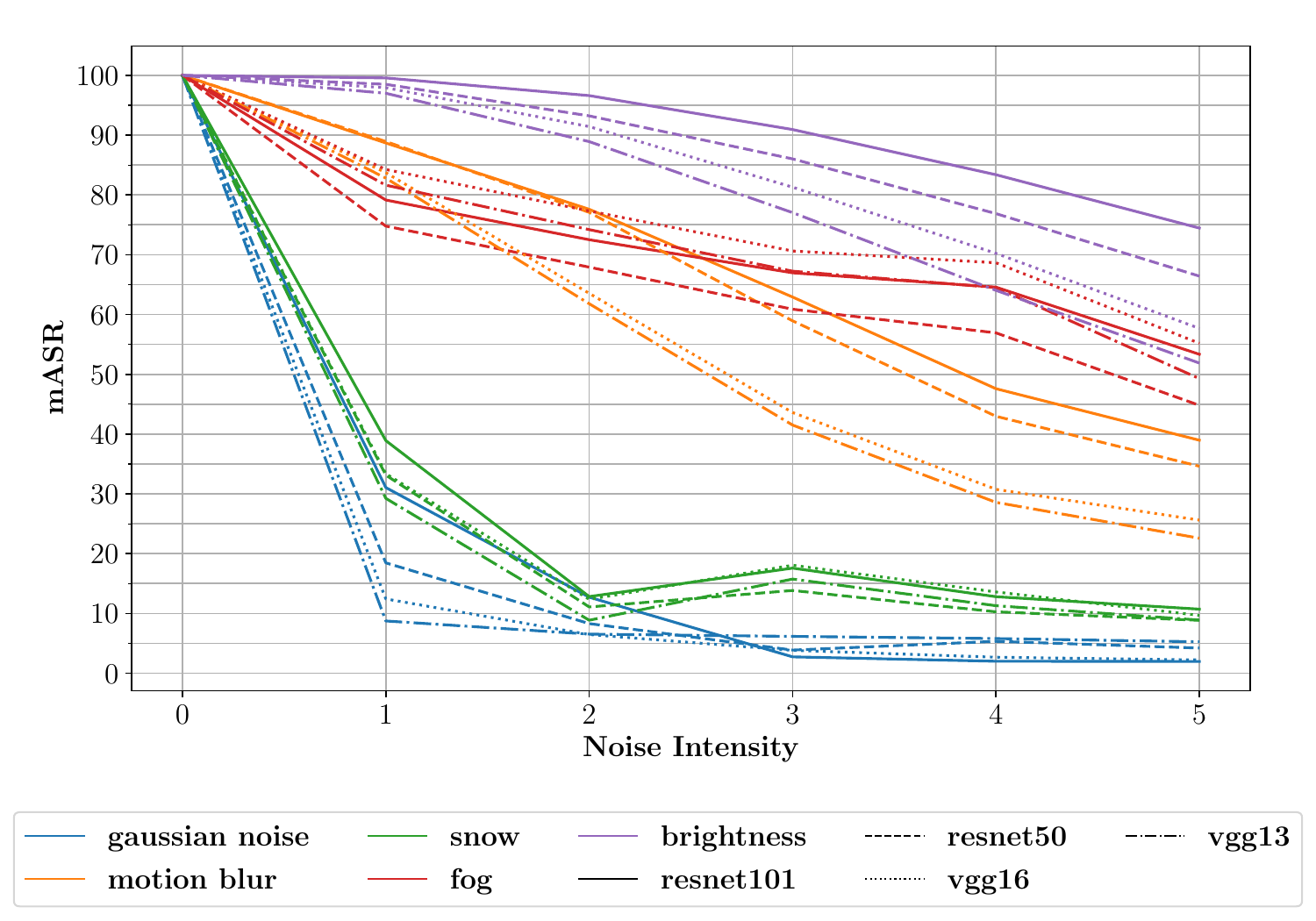}
\caption{mASR score as a function of the injected noise intensity.  %
}
\label{fig:mASR}
\end{figure}

To validate the new metric, we used as reference the supervised training on the \textit{Cityscapes} dataset and compared it with the training on corrupted versions of the same dataset using the introduced mASR metric to evaluate the relative performances and so, indirectly, the domain shift introduced by the perturbations. %
In Fig.~\ref{fig:mASR} we identified $5$ types of perturbations which are likely to be encountered by an agent moving outdoor (\ie, Gaussian noise, motion blur, snow, fog, brightness) and we set $5$ levels of noise intensity as defined by \cite{hendrycks2019benchmarking}. As expected, the higher is the noise intensity and the lower is the adaptation score computed by mASR. Furthermore, we can also have a hint of the most detrimental types of noise for adapting source knowledge: namely, Gaussian noise, snow, motion blur. This can help us identify which set of samples we should consider more in order to obtain a reliable model capable of handling these situations.
On the other hand, brightness and fog influence less the final scores.

%% file: sections/results.tex
\newcommand{\phantomDigit}[1]{\makebox{\hphantom{0}#1}}
\newcommand{\Cdash}{\hspace*{.8em}\hbox{-}}
\newcommand{\lenA}{.45em}
\newcommand{\lenB}{.8em}
\newcommand{\lenC}{1.5em}
\newcommand{\lenD}{2em}
\newcolumntype{P}[1]{>{\fontsize{8}{8}\selectfont\centering\arraybackslash}p{#1}}
\newcolumntype{Q}[1]{>{\fontsize{8}{8}\selectfont}p{#1}}
\newcolumntype{Y}[1]{>{\fontsize{8}{8}\selectfont\centering\arraybackslash}X{#1}}
\begin{table*}[h!]
\caption{Comparison of adaptation strategies in terms of IoU, mIoU and mASR  (Sec.~\ref{sec:results}). Best in \textbf{bold}, runner-up \underline{underlined}. mIoU\textsuperscript{1} and mASR\textsuperscript{1} restricted to 13 classes, ignoring the classes with same superscript.} %
\label{table:results}
\linespread{0.7}\selectfont\centering
\begin{tabularx}{\textwidth}{P{\lenA}P{\lenA}YQ{\lenA}Q{\lenA}Q{\lenA}Q{\lenA}Q{\lenA}Q{\lenA}Q{\lenA}Q{\lenA}Q{\lenA}Q{\lenA}Q{\lenA}Q{\lenA}Q{\lenA}Q{\lenA}Q{\lenA}Q{\lenA}Q{\lenA}Q{\lenA}Q{\lenB}P{\lenC}P{\lenC}P{\lenD}P{\lenD}}
\toprule[1pt]
\rotatebox{45}{Backbone} & \rotatebox{45}{Setup} & Configuration & \rotatebox{45}{Road} & \rotatebox{45}{Sidewalk} & \rotatebox{45}{Building} & \rotatebox{45}{Wall\textsuperscript{1}} & \rotatebox{45}{Fence\textsuperscript{1}} & \rotatebox{45}{Pole\textsuperscript{1}} & \rotatebox{45}{Traffic Light} & \rotatebox{45}{Traffic Sign} & \rotatebox{45}{Vegetation} & \rotatebox{45}{Terrain} & \rotatebox{45}{Sky} & \rotatebox{45}{Person} & \rotatebox{45}{Rider} & \rotatebox{45}{Car} & \rotatebox{45}{Truck} & \rotatebox{45}{Bus} & \rotatebox{45}{Train} & \rotatebox{45}{Motorbike} & \rotatebox{45}{Bicycle} & \makebox{mIoU} & \makebox{mIoU\textsuperscript{1}} & \makebox{mASR} & \makebox{mASR\textsuperscript{1}}\\
\noalign{\smallskip} 
\toprule[1pt]
\multirow{26}{*}{\rotatebox{90}{ResNet101}} & & Target only & 96.5 & 73.8 & 88.4 & 42.2 & 43.7 & 40.7 & 46.1 & 58.6 & 88.5 & 54.9 & 91.9 & 68.7 & 46.2 & 90.7 & 68.8 & 69.9 & 48.8 & 47.6 & 64.5 & 64.8 & - & 100 & 100\\
\cline{2-26}%
\noalign{\smallskip}
\footnotesize
& \multirow{10}{*}{\rotatebox{90}{From GTAV}} & Baseline~\cite{toldo2020clustering} & 71.4 & 15.3 & 74.0 & 21.1 & 14.4 & 22.8 & 33.9 & 18.6 & 80.7 & 20.9 & 68.5 & 56.6 & 27.1 & 67.4 & 32.8 & ~~5.6 & ~~7.7 & 28.4 & 33.8 & 36.9 & - & 54.0 & -\\
\cdashline{3-26}
\noalign{\smallskip}
& & ASN (feat)~\cite{tsai2018} & 83.7 & 27.6 & 75.5 & 20.3 & 19.9 & 27.4 & 28.3 & 27.4 & 79.0 & 28.4 & 70.1 & 55.1 & 20.2 & 72.9 & 22.5 & 35.7 & ~~\underline{8.3} & 20.6 & 23.0 & 39.0 & - & 56.9 & -\\
& & MinEnt~\cite{vu2019advent} & 84.4 & 18.7 & 80.6 & 23.8 & 23.2 & 28.4 & 36.9 & 23.4 & 83.2 & 25.2 & \underline{79.4} & 59.0 & \textbf{29.9} & 78.5 & 33.7 & 29.6 & ~~1.7 & 29.9 & 33.6 & 42.3 & - & 61.9 & -\\
& & SAPNet~\cite{li2020spatial} & 88.4 & \textbf{38.7} & 79.5 & \underline{29.4} & \textbf{24.7} & 27.3 & 32.6 & 20.4 & 82.2 & 32.9 & 73.3 & 55.5 & 26.9 & 82.4 & 31.8 & 41.8 & ~~2.4 & 26.5 & 24.1 & 43.2 & - & 63.1 & -\\
& & MaxSquareIW~\cite{Chen2019} & 87.7 & 25.2 & \textbf{82.9} & \textbf{30.9} & 24.0 & 29.0 & 35.4 & 24.2 & 84.2 & 38.2 & 79.2 & 59.0 & \underline{27.7} & 79.5 & 34.6 & 44.2 & ~~7.5 & \underline{31.1} & \underline{40.3} & 45.5 & - & 62.2 & -\\
& & UDA OCE~\cite{toldo2020clustering} & \textbf{89.4} & 30.7 & 82.1 & 23.0 & 22.0 & \underline{29.2} & \underline{37.6} & \textbf{31.7} & 83.9 & 37.9 & 78.3 & \underline{60.7} & 27.4 & \textbf{84.6} & \underline{37.6} & \underline{44.7} & ~~7.3 & 26.0 & 38.9 & 45.9 & - & 67.3 & -\\
& & LSR \cite{barbato2021latent} & 87.7 & \underline{32.6} & \underline{82.6} & 29.1 & 23.0 & 28.5 & 36.1 & \underline{28.5} & \textbf{84.8} & \underline{41.8} & \textbf{80.1} & 59.4 & 23.8 & 76.5 & \textbf{38.4} & \textbf{45.8} & ~~7.1 & 28.5 & 40.1 & \underline{46.0} & - & \underline{67.7} & -\\
& & LSR$^{+}$ (ours) & \underline{88.9} & 26.6 & 82.0 & 21.0 & \underline{24.4} & \textbf{30.1} & \textbf{41.1} & 27.0 & \underline{84.7} & \textbf{42.7} & \textbf{80.1} & \textbf{63.0} & 26.4 & \underline{83.1} & 30.4 & 44.3 & \textbf{16.8} & \textbf{35.8} & \textbf{42.4} & \textbf{46.9} & - & \textbf{69.5} & -\\
\cdashline{3-26}
& & LSR$^{+}$ on \textit{CS-full} & 89.3 & 28.7 & 82.1 & 25.2 & 27.5 & 31.9 & 40.3 & 33.2 & 84.7 & 38.7 & 81.2 & 63.2 & 27.2 & 85.2 & 34.7 & 43.9 & ~~9.8 & 37.2 & 47.7 & 48.0 & - & 71.3 & -\\

\cline{2-26}%
\noalign{\smallskip}
& \multirow{10}{*}{\rotatebox{90}{From SYNTHIA}} & Baseline~\cite{toldo2020clustering} & 17.7 & 15.0 & 74.3 & 10.1 & ~~0.1 & 25.5 & 6.3 & 10.2 & 75.5 & \Cdash & 77.9 & 57.1 & 19.2 & 31.2 & \Cdash & 31.2 & \Cdash & 10.0 & 20.1 & 30.1 & 34.3 & 41.7 & 44.6\\
\cdashline{3-26}
\noalign{\smallskip}
& & ASN (feat)~\cite{tsai2018} & 62.4 & 21.9 & 76.3 & \Cdash & \Cdash & \Cdash & 11.7 & 11.4 & 75.3 & \Cdash & 80.9 & 53.7 & 18.5 & 59.7 & \Cdash & 13.7 & \Cdash & \textbf{20.6} & 24.0 & - & 40.8 & - & 52.5\\
& & MinEnt~\cite{vu2019advent} & 73.5 & 29.2 & 77.1 & ~~7.7 & ~~\underline{0.2} & 27.0 & ~~7.1 & 11.4 & 76.7 & \Cdash & 82.1 & 57.2 & \underline{21.3} & 69.4 & \Cdash & 29.2 & \Cdash & 12.9 & 27.9 & 38.1 & 44.2 & 51.1 & 56.3\\
& & SAPNet~\cite{li2020spatial} & 81.7 & 33.5 & 75.9 & \Cdash & \Cdash & \Cdash & ~~7.0 & ~~6.3 & 74.8 & \Cdash & 78.9 & 52.1 & \underline{21.3} & 75.7 & \Cdash & 30.6 & \Cdash & 10.8 & 28.0 & - & 44.3 & - & 56.0\\
& & MaxSquareIW~\cite{Chen2019} & 78.9 & 33.5 & 75.3 & \underline{15.0} & ~~\textbf{0.3} & 27.5 & \underline{13.1} & \textbf{16.7} & 73.8 & \Cdash & 77.7 & 50.4 & 19.9 & 66.7 & \Cdash & \textbf{36.1} & \Cdash & 13.7 & 32.1 & 39.4 & 45.2 & 53.8 & 58.3\\
& & UDA OCE~\cite{toldo2020clustering} & \textbf{88.3} & \textbf{42.2} & 79.1 & ~~7.1 & ~~\underline{0.2} & 24.4 & \textbf{16.8} & \underline{16.5} & \underline{80.0} & \Cdash & \textbf{84.3} & 56.2 & 15.0 & \textbf{83.5} & \Cdash & 27.2 & \Cdash & ~~6.3 & 30.7 & 41.1 & \underline{48.2} & 54.3 & 60.9\\
& & LSR~\cite{barbato2021latent} & 81.0 & 36.9 & \textbf{79.5} & 13.4 & ~~\underline{0.2} & \underline{28.7} & ~~9.0 & 16.1 & 79.1 & \Cdash & 81.7 & \underline{57.9} & \textbf{21.6} & 77.2 & \Cdash & \underline{35.3} & \Cdash & 14.2 & \underline{35.4} & \underline{41.7} & 48.1 & \underline{56.5} & \underline{61.6}\\
& & LSR$^{+}$ (ours) & \underline{82.6} & \underline{38.4} & \textbf{80.6} & \textbf{15.5} & ~~\textbf{0.3} & \textbf{31.8} & ~~6.7 & 16.3 & \textbf{81.7} & \Cdash & \underline{82.5} & \textbf{58.4} & 20.2 & \underline{81.3} & \Cdash & 32.7 & \Cdash & \underline{15.3} & \textbf{36.7} & \textbf{42.6} & \textbf{48.7} & \textbf{57.7} & \textbf{62.1}\\
\cdashline{3-26}
& & LSR$^{+}$ on \textit{CS-full} & 89.4 & 47.9 & 79.4 & 13.9 & ~~0.4 & 29.5 & 10.0 & 16.5 & 79.5 & \Cdash & 83.3 & 57.7 & 17.0 & 84.3 & \Cdash & 37.7 & \Cdash & 21.5 & 28.6 & 43.5 & 50.2 & 58.8 & 64.2\\
\end{tabularx}
\end{table*}

\renewcommand{\lenA}{1.75em}
\renewcommand{\lenB}{2.2em}
\renewcommand{\lenC}{6em}
\begin{table*}[h!]
\caption{Quantitative results on the \textit{Cross-City} real-to-real benchmark. (r) indicates that the strategy was re-trained, starting from the official code. Best in \textbf{bold}, runner-up \underline{underlined}.}
\label{table:crosscity}
\linespread{0.7}\selectfont\centering
\begin{tabularx}{\textwidth}{P{\lenA}YQ{\lenA}Q{\lenA}Q{\lenA}Q{\lenA}Q{\lenA}Q{\lenA}Q{\lenA}Q{\lenA}Q{\lenA}Q{\lenA}Q{\lenA}Q{\lenA}Q{\lenB}P{\lenC}}
\toprule[1pt]
\rotatebox{20}{Target City} & Configuration & \rotatebox{20}{Road} & \rotatebox{20}{Sidewalk} & \rotatebox{20}{Building} & \rotatebox{20}{Traffic Light} & \rotatebox{20}{Traffic Sign} & \rotatebox{20}{Vegetation} & \rotatebox{20}{Sky} & \rotatebox{20}{Person} & \rotatebox{20}{Rider} & \rotatebox{20}{Car} & \rotatebox{20}{Bus} & \rotatebox{20}{Motorbike} & \rotatebox{20}{Bicycle} & \makebox{mIoU}\\

\noalign{\smallskip} 
\toprule[1pt]
\multirow{7}{*}{\rotatebox{90}{Rome}} & Source only~\cite{Chen2019} & 85.0 & 34.7 &  86.4 & 17.5 & 39.0 & 84.9 & 85.4 & 43.8 & 15.5 & 81.8 & 46.3 & 38.4 & \phantomDigit{4.8} & 51.0 \\
& \textit{Cross-City}~\cite{chen2017no} & 79.5 & 29.3 & 84.5 & \phantomDigit{0.0} & 22.2 & 80.6 & 82.8 & 29.5 & 13.0 & 71.7 & 37.5 & 25.9 & \phantomDigit{1.0} & 42.9 \\
& ASN (feat)~\cite{tsai2018} & 83.9 & 34.2 & \textbf{88.3} & 18.8 & 40.2 & \textbf{86.2} & \underline{93.1} & 47.8 & 21.7 & 80.9 & 47.8 & 48.3 & \phantomDigit{8.6} & 53.8 \\
& MaxSquareIW~\cite{Chen2019} (r) & \textbf{86.2} & \textbf{37.8} & 86.4 & 22.3 & 39.5 & 85.4 & 84.0 & \underline{49.5} & 21.2 & 82.7 & \textbf{55.3} & \underline{48.5} & \phantomDigit{\textbf{9.5}} & 54.5 \\
& UDA OCE~\cite{toldo2020clustering} (r) & \underline{85.6} & \underline{35.0} & 87.9 & \underline{23.1} & \underline{42.0} & \underline{85.9} & 89.2 & 49.3 & \underline{24.3} & \underline{82.8} & \underline{48.8} & \underline{48.5} & \phantomDigit{\underline{9.0}} & \underline{54.7} \\
& LSR$^{+}$ (ours) & 83.4 & 34.5 & \underline{88.1} & \textbf{29.0} & \textbf{44.5} & 85.5 & \textbf{93.9} & \textbf{51.9} & \textbf{31.3} & \textbf{83.2} & 44.7 & \textbf{51.5} & \phantomDigit{8.8} & \textbf{56.2} \\

\toprule[1pt]
\multirow{7}{*}{\rotatebox{90}{Rio}} & Source only~\cite{Chen2019} & 74.2 & 42.2 & 84.0 & 12.1 & 20.4 & 78.3 & \textbf{87.9} & 50.1 & 25.6 & \textbf{76.6} & \underline{40.0} & 27.6 & 17.0 & 48.9 \\
& \textit{Cross-City}~\cite{chen2017no} & 74.2 & 43.9 & 79.0 & \phantomDigit{2.4} & \phantomDigit{7.5} & 77.8 & 69.5 & 39.3 & 10.3 & 67.9 & \textbf{41.2} & 27.9 & 10.9 & 42.5 \\
& ASN (feat)~\cite{tsai2018} & 76.2 & 44.7 & \underline{84.6} & \phantomDigit{9.3} & \textbf{25.5} & \textbf{81.8} & \underline{87.3} & 55.3 & 32.7 & 74.3 & 28.9 & \textbf{43.0} & 27.6 & 51.6 \\
& MaxSquareIW~\cite{Chen2019} (r) & \textbf{79.5} & \underline{50.7} & 84.5 & \textbf{14.9} & 17.7 & 80.8 & 85.7 & 54.5 & 29.6 & \underline{75.1} & 37.0 & 40.6 & 24.5 & 51.9 \\
& UDA OCE~\cite{toldo2020clustering} (r)& \underline{78.9} & 48.5 & \textbf{85.3} & \underline{14.2} & \underline{24.4} & \underline{81.3} & 87.0 & \underline{55.9} & \underline{36.2} & 74.3 & 29.7 & \underline{41.8} & 27.9 & \textbf{52.7} \\
& LSR$^{+}$ (ours) & \textbf{79.5} & \textbf{52.2} & 83.7 & 10.2 & 23.1 & 79.3 & 82.3 & \textbf{59.8} & \textbf{40.0} & 75.0 & 23.0 & \textbf{43.0} & \textbf{29.0} & \underline{52.3} \\

\toprule[1pt]
\multirow{7}{*}{\rotatebox{90}{Tokyo}} & Source only~\cite{Chen2019} & 81.4 & 28.4 & \underline{78.1} & 14.5 & 19.6 & 81.4 & 86.5 & 51.9 & 22.0 & 70.4 & \underline{18.2} & \underline{22.3} & 46.4 & 47.8 \\
& \textit{Cross-City}~\cite{chen2017no} & 83.4 & \textbf{35.4} & 72.8 & 12.3 & 12.7 & 77.4 & 64.3 & 42.7 & 21.5 & 64.1 & \textbf{20.8} & \phantomDigit{8.9} & 40.3 & 42.8 \\
& ASN (feat)~\cite{tsai2018} & 81.5 & 26.0 & 77.8 & \textbf{17.8} & \textbf{26.8} & \underline{82.7} & \textbf{90.9} & \underline{55.8} & \textbf{38.0} & 72.1 & \phantomDigit{4.2} & \textbf{24.5} & 50.8 & \underline{49.9} \\
& MaxSquareIW~\cite{Chen2019} (r) & 84.1 & 32.9 & 76.7 & 11.3 & 23.8 & 82.3 & 87.4 & 55.3 & 30.0 & 72.0 & \phantomDigit{8.6} & 18.9 & 47.1 & 48.5 \\
& UDA OCE~\cite{toldo2020clustering} (r) & \textbf{85.0} & 33.3 & 77.9 & \phantomDigit{8.5} & \underline{25.5} & 82.5 & \underline{89.4} & \textbf{56.1} & 29.2 & \textbf{72.4} & \phantomDigit{2.1} & 12.3 & 41.9 & 47.4 \\
& LSR$^{+}$ (ours) & \underline{84.2} & \underline{34.6} & \textbf{78.2} & \underline{16.8} & 22.6 & \textbf{83.3} & 89.3 & 55.0 & \underline{33.2} & 72.0 & \phantomDigit{8.6} & 20.5 & \textbf{52.2} & \textbf{50.0}\\

\toprule[1pt]
\multirow{7}{*}{\rotatebox{90}{Taipei}} & Source only~\cite{Chen2019} & \textbf{82.6} & \textbf{33.0} & \underline{86.3} & 16.0 & \textbf{16.5} & \underline{78.3} & 83.3 & 26.5 & \phantomDigit{8.4} & 70.7 & 36.1 & 47.9 & 15.7 & 46.3 \\
& \textit{Cross-City}~\cite{chen2017no} & 78.6 & 28.6 & 80.0 & 13.1 & \phantomDigit{7.6} & 68.2 & 82.1 & 16.8 & \phantomDigit{9.4} & 60.4 & 34.0 & 26.5 & \phantomDigit{9.9} & 39.6 \\
& ASN (feat)~\cite{tsai2018} & 81.7 & 29.5 & 85.2 & \textbf{26.4} & \underline{15.6} & 76.7 & 91.7 & 31.0 & \textbf{12.5} & \underline{71.5} & 41.1 & 47.3 & 27.7 & \underline{49.1} \\
& MaxSquareIW~\cite{Chen2019} (r) & 80.9 & 31.3 & 83.3 & 12.9 & 13.4 & 75.4 & 89.5 & 31.8 & \phantomDigit{3.9} & 69.0 & \textbf{44.3} & \underline{49.4} & \textbf{33.3} & 47.6 \\
& UDA OCE~\cite{toldo2020clustering} (r) & 81.4 & 30.1 & 84.3 & 16.7 & 13.4 & 75.4 & \textbf{91.9} & \underline{32.5} & \phantomDigit{4.6} & 71.0 & \underline{41.4} & 48.0 & \textbf{33.3} & 48.0\\
& LSR$^{+}$ (ours) & \underline{81.8} & \underline{32.9} & \textbf{86.8} & \underline{19.1} & 14.2 & \textbf{79.3} & \underline{91.8} & \textbf{35.1} & \underline{11.6} & \textbf{72.8} & 33.8 & \textbf{58.7} & \underline{31.6} & \textbf{50.0} \\

\end{tabularx}
\end{table*}

\section{Experimental Evaluation}
\label{sec:results}

The qualitative and quantitative results achieved by the proposed approach (LSR$^+$) in various driving contexts  will be presented in this section, where it will be compared with several other feature-level approaches (\ie, \cite{tsai2018,li2020spatial,toldo2020clustering}), with some entropy minimization strategies (\ie, \cite{vu2019advent,Chen2019}) that have a similar effect on feature distribution, and finally with the conference version of our work \cite{barbato2021latent}. The key feature of these approaches is the training efficiency, indeed the addition of such constraints does not increase the computation complexity of the training, differently from hugely expensive generative networks or modified architectures.%

Our end-to-end method allows straightforward integration with other strategies, \eg, adversarial approaches at input or output level, or entropy minimization.
In order to verify such compatibility, we introduce an additional entropy-minimization loss~\cite{Chen2019} in our setup.
We start from considering two widely used synthetic-to-real benchmarks and a standard ResNet-101 as backbone architecture obtaining the results shown in Table~\ref{table:results}. Then, a real-to-real benchmark~\cite{chen2017no} has also been used (see  Table~\ref{table:crosscity}).
To further verify the robustness of our setup, in Table~\ref{table:backbones} we report some results using different backbones (\ie, ResNet50, VGG16 and VGG13).

\newcommand{\imgWidth}{0.142\textwidth}
\begin{figure*}
\newcolumntype{Y}{>{\centering\arraybackslash}X}
\centering
\begin{subfigure}{.6em}
\scriptsize\rotatebox{90}{~~~\textit{GTAV}$\rightarrow$\textit{Cityscapes}}
\end{subfigure}%
\begin{subfigure}{\textwidth-1em}
\hspace*{.2em}%
\begin{subfigure}{\imgWidth}
\includegraphics[width=\textwidth]{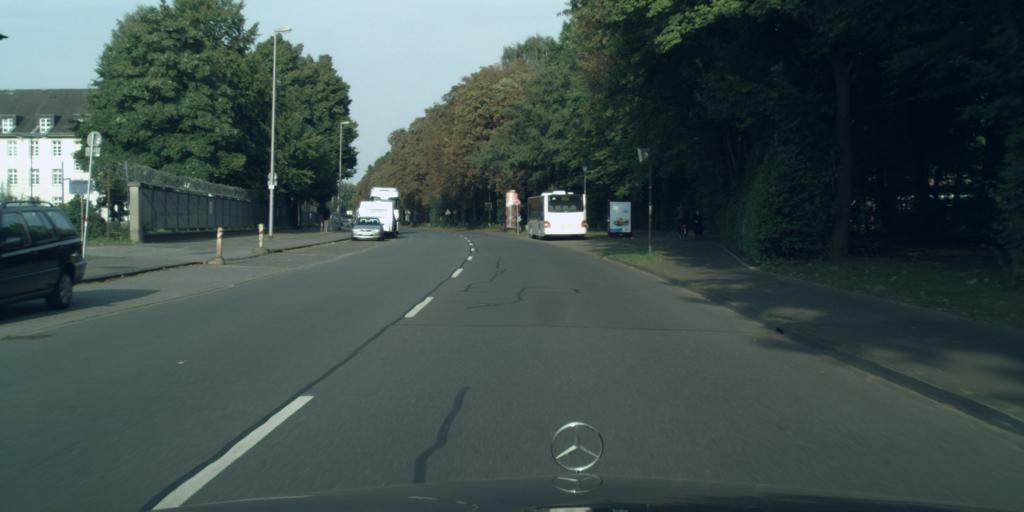}
\end{subfigure}%
\begin{subfigure}{\imgWidth}
\includegraphics[width=\textwidth]{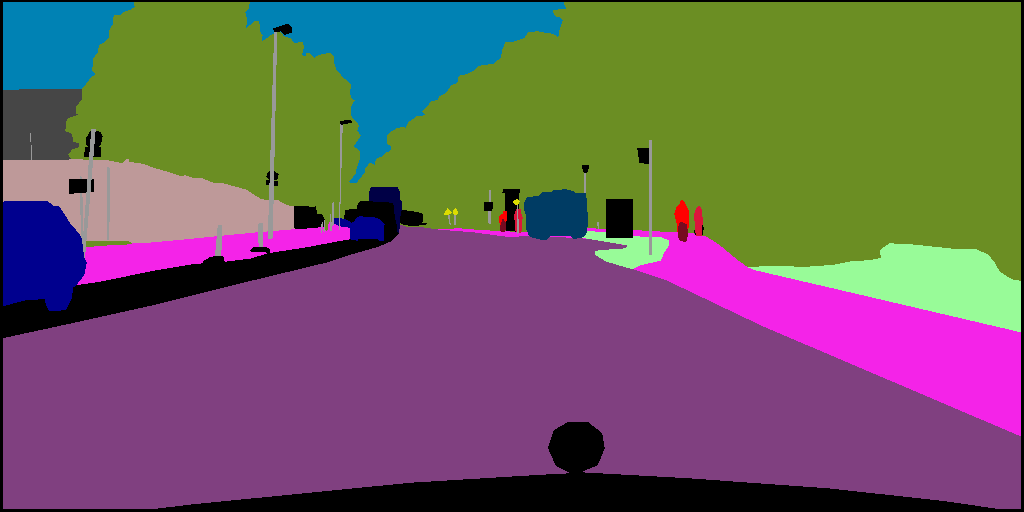}
\end{subfigure}%
\begin{subfigure}{\imgWidth}
\includegraphics[width=\textwidth]{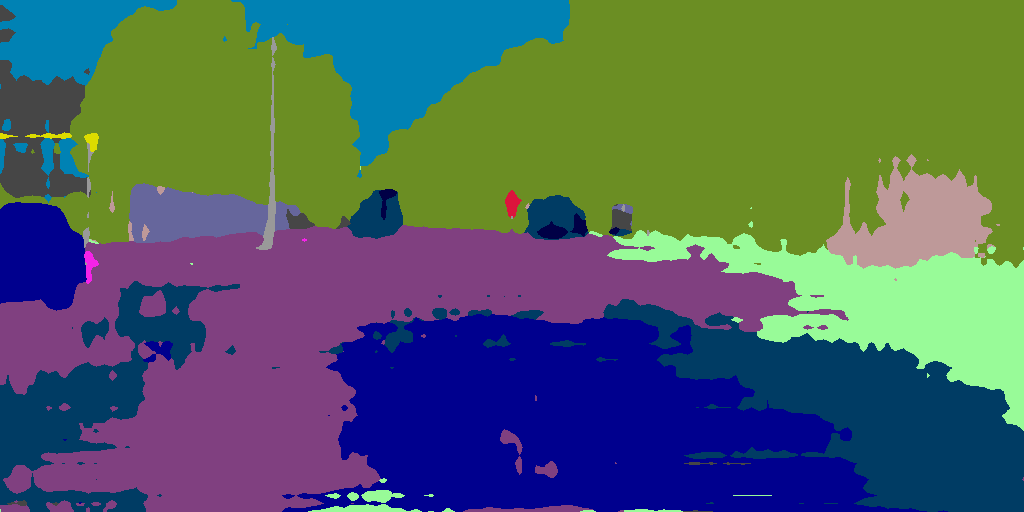}
\end{subfigure}%
\begin{subfigure}{\imgWidth}
\includegraphics[width=\textwidth]{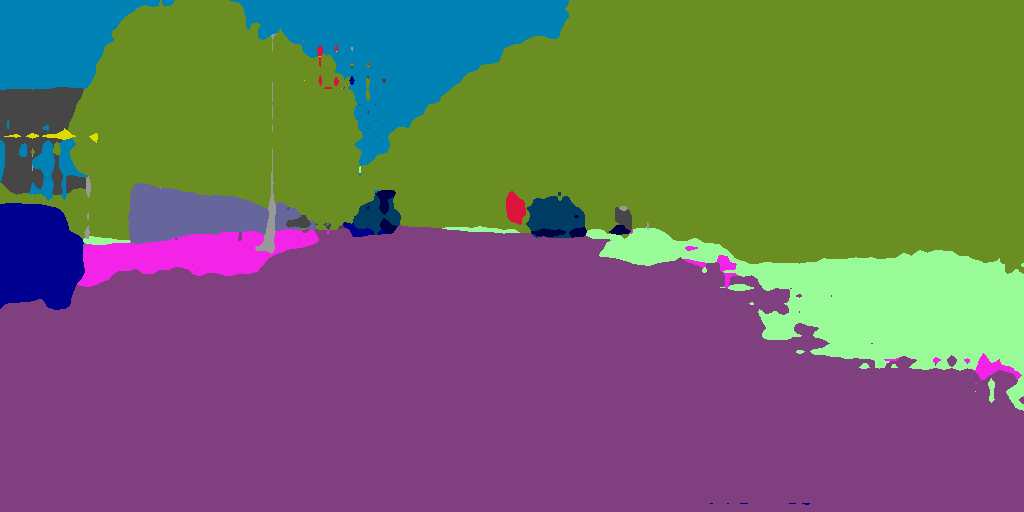}
\end{subfigure}%
\begin{subfigure}{\imgWidth}
\includegraphics[width=\textwidth]{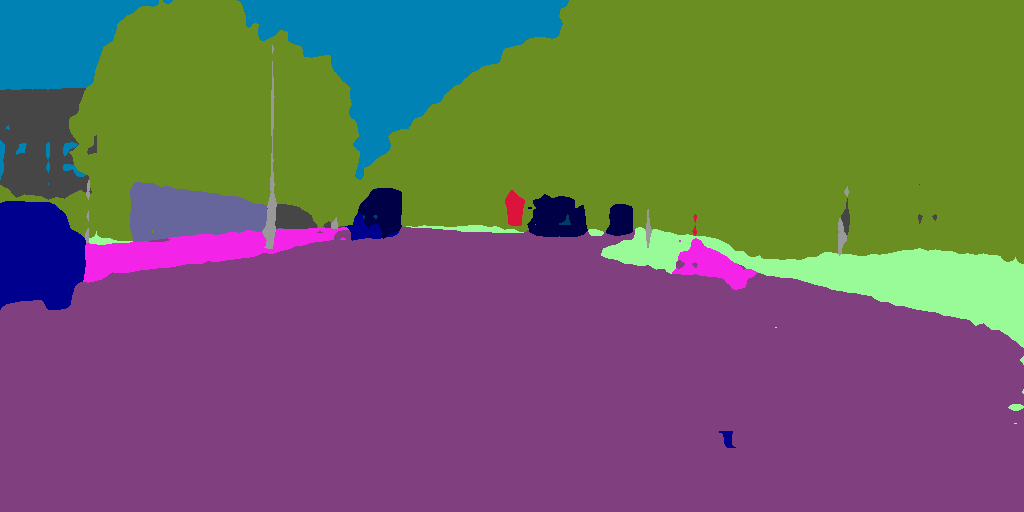}
\end{subfigure}%
\begin{subfigure}{\imgWidth}
\includegraphics[width=\textwidth]{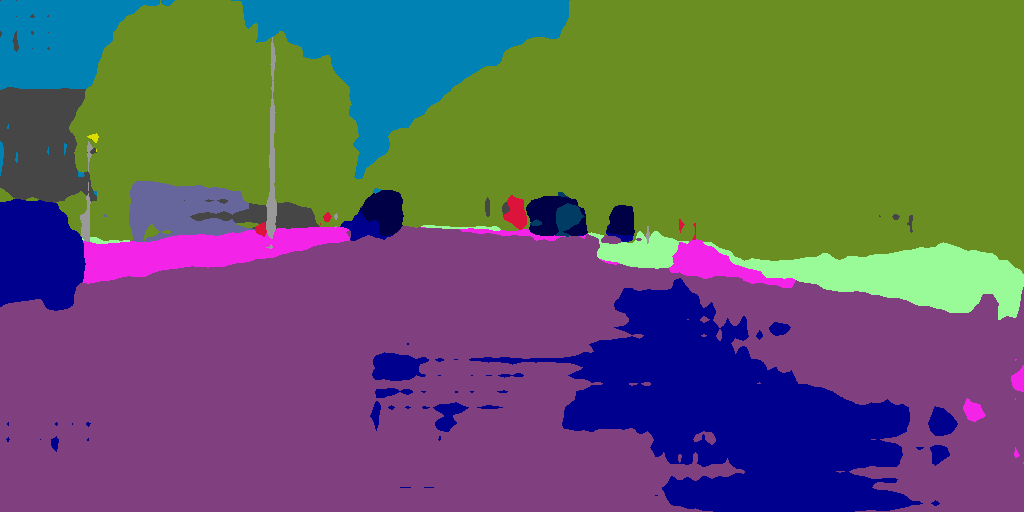}
\end{subfigure}%
\begin{subfigure}{\imgWidth}
\includegraphics[width=\textwidth]{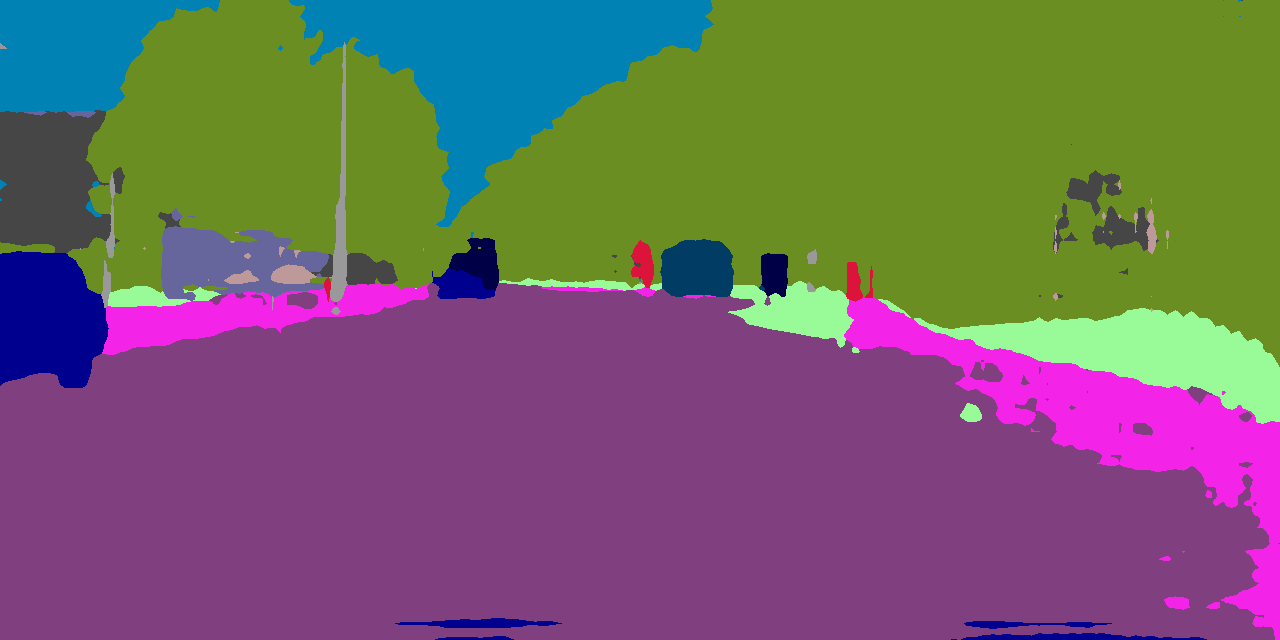}
\end{subfigure}

\hspace*{.2em}%
\begin{subfigure}{\imgWidth}
\includegraphics[width=\textwidth]{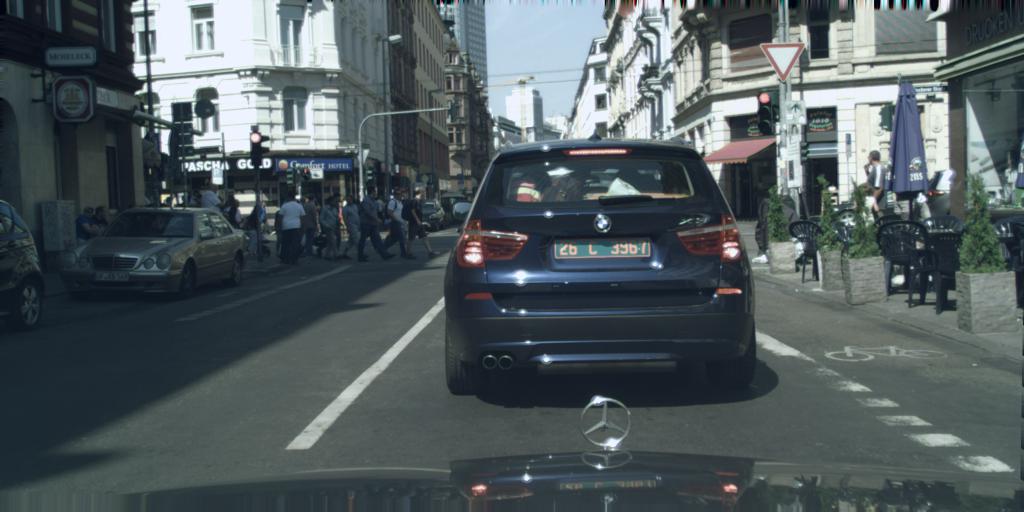}
\end{subfigure}%
\begin{subfigure}{\imgWidth}
\includegraphics[width=\textwidth]{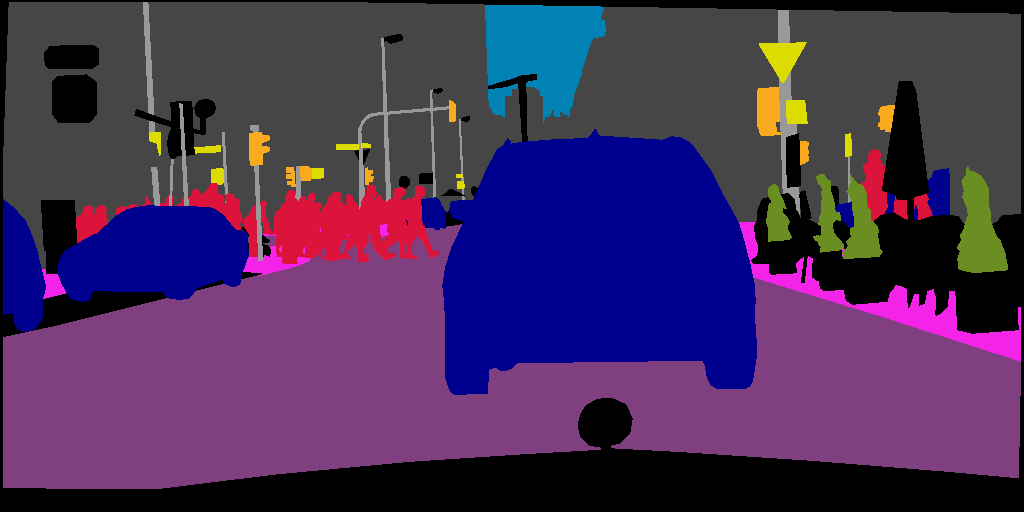}
\end{subfigure}%
\begin{subfigure}{\imgWidth}
\includegraphics[width=\textwidth]{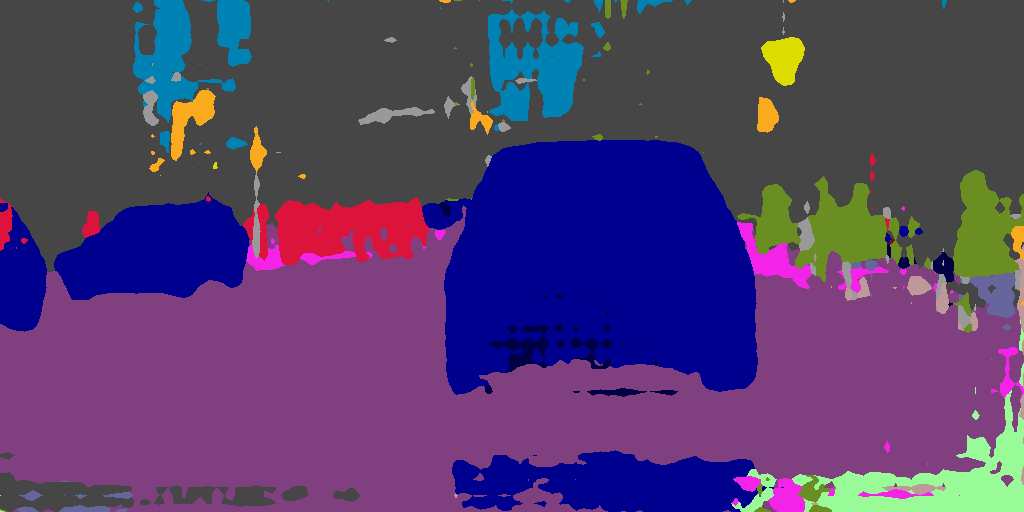}
\end{subfigure}%
\begin{subfigure}{\imgWidth}
\includegraphics[width=\textwidth]{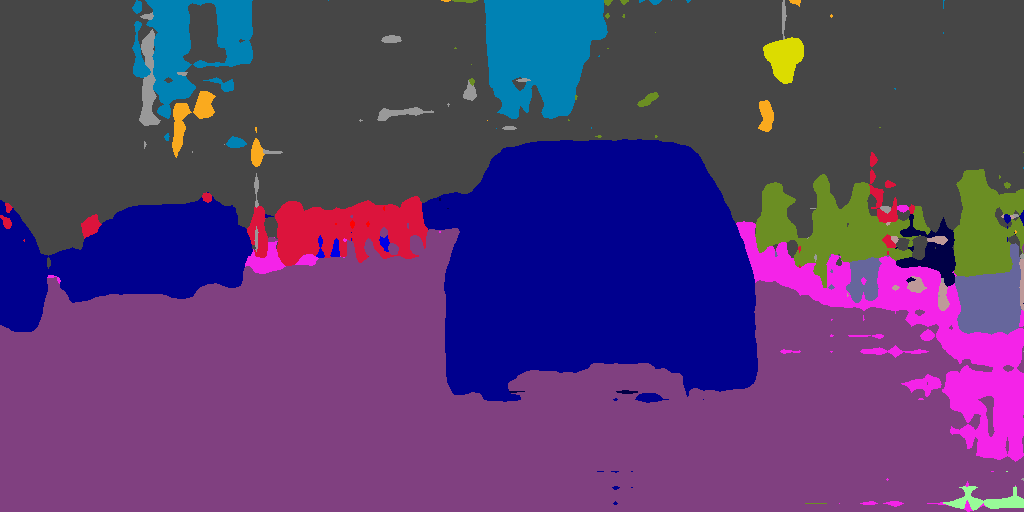}
\end{subfigure}%
\begin{subfigure}{\imgWidth}
\includegraphics[width=\textwidth]{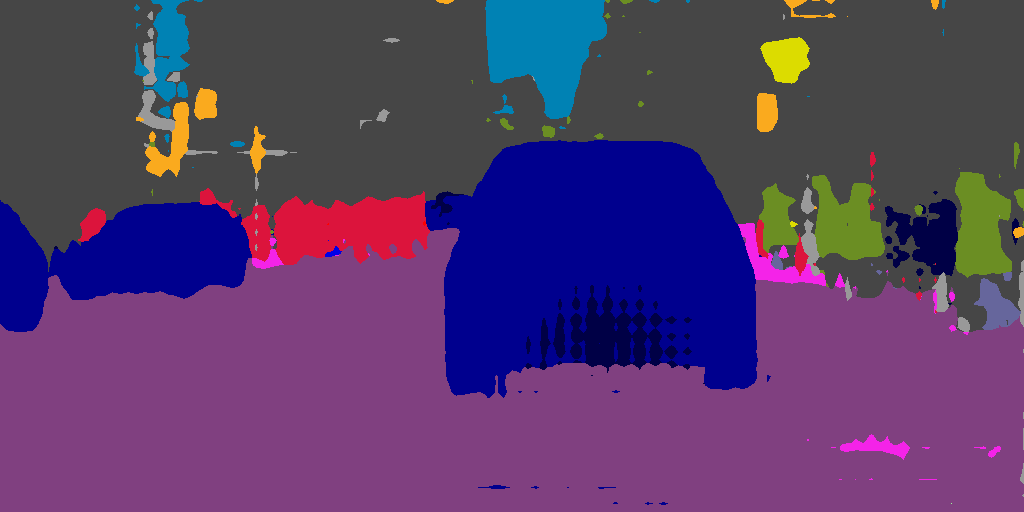}
\end{subfigure}%
\begin{subfigure}{\imgWidth}
\includegraphics[width=\textwidth]{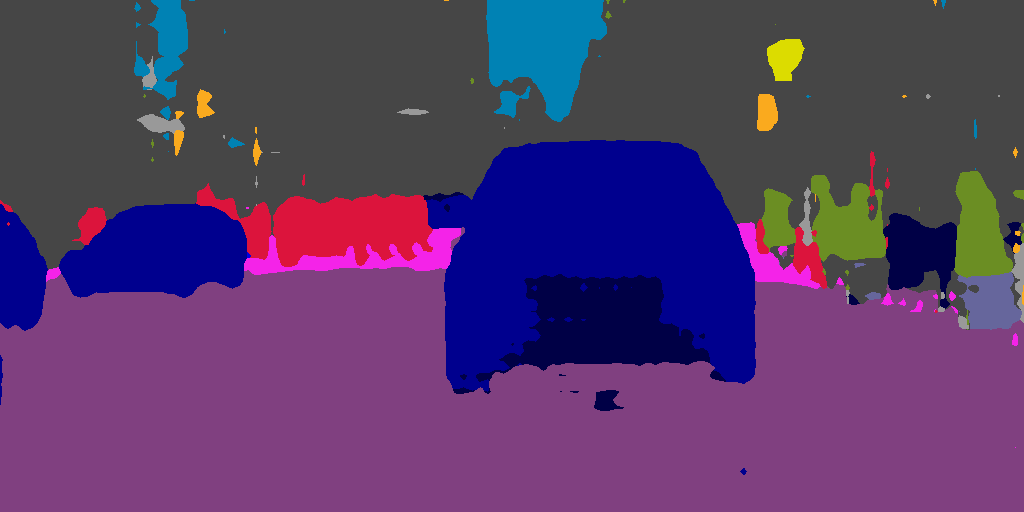}
\end{subfigure}%
\begin{subfigure}{\imgWidth}
\includegraphics[width=\textwidth]{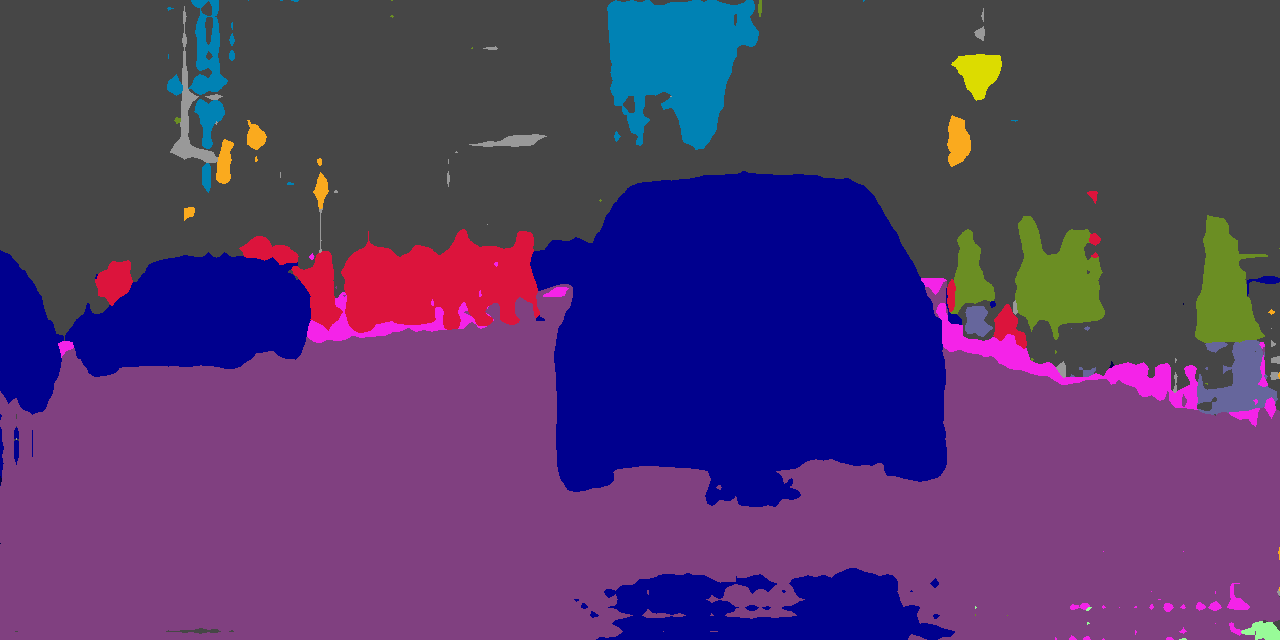}
\end{subfigure}
\end{subfigure}

\begin{subfigure}{\textwidth}
\small
\begin{tabularx}{\textwidth}{YYYYYYYYYY}
\cellcolor{road} \textcolor{white}{Road} & \cellcolor{sidewalk} Sidewalk & \cellcolor{building} \textcolor{white}{Building} & \cellcolor{wall} \textcolor{white}{Wall} & \cellcolor{fence} Fence & \cellcolor{pole} Pole & \cellcolor{tlight} T. Light & \cellcolor{tsign} T. Sign & \cellcolor{vegetation} \textcolor{white}{\footnotesize Vegetation} & \cellcolor{terrain} Terrain \\
\cellcolor{sky} Sky & \cellcolor{person} \textcolor{white}{Person} & \cellcolor{rider} \textcolor{white}{Rider} & \cellcolor{car} \textcolor{white}{Car} & \cellcolor{truck} \textcolor{white}{Truck} & \cellcolor{bus} \textcolor{white}{Bus} &  \cellcolor{train} \textcolor{white}{Train} & \cellcolor{motorbike} \footnotesize \textcolor{white}{Motorbike} & \cellcolor{bicycle} \textcolor{white}{Bicycle} & \cellcolor{unlabelled} \footnotesize \textcolor{white}{Unlabeled}
\end{tabularx}
\end{subfigure}
\begin{subfigure}{.6em}
\scriptsize\rotatebox{90}{~~~~~\textit{SYNTHIA}$\rightarrow$\textit{Cityscapes}}
\end{subfigure}%
\begin{subfigure}{\textwidth-1em}
\hspace*{.2em}%
\begin{subfigure}{\imgWidth}
\includegraphics[width=\textwidth]{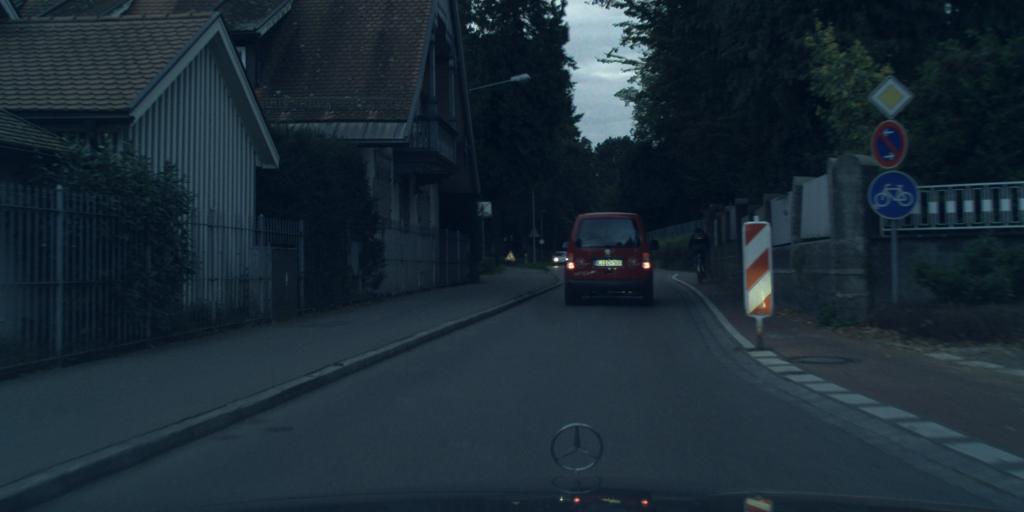}
\end{subfigure}%
\begin{subfigure}{\imgWidth}
\includegraphics[width=\textwidth]{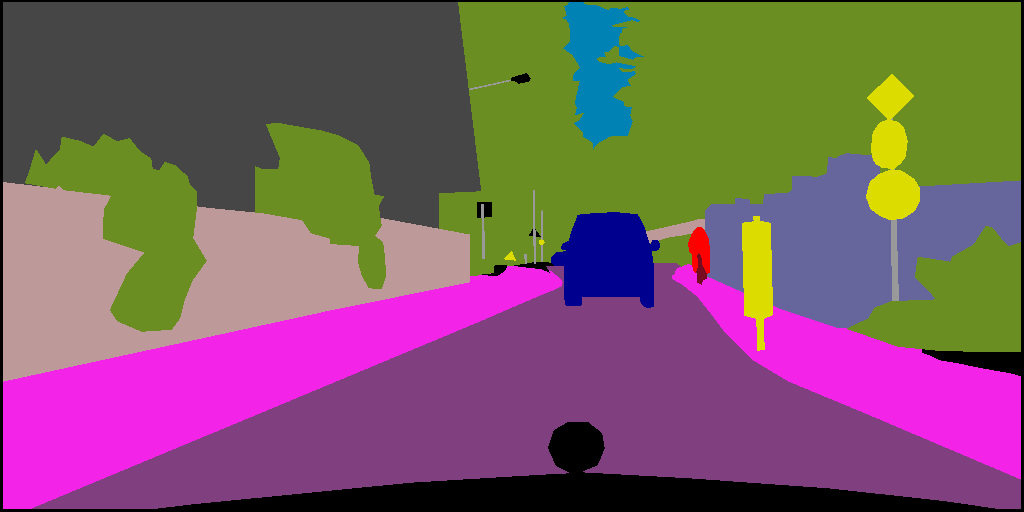}
\end{subfigure}%
\begin{subfigure}{\imgWidth}
\includegraphics[width=\textwidth]{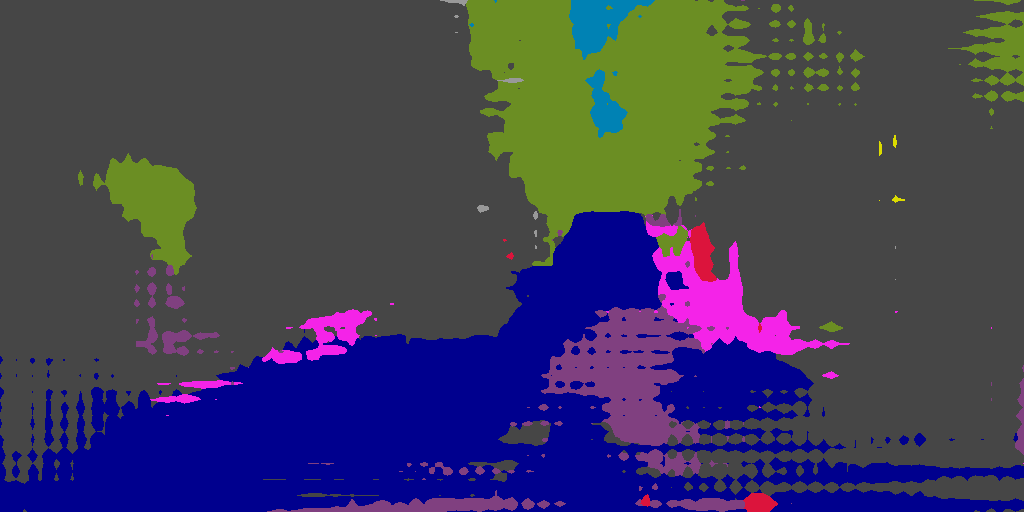}
\end{subfigure}%
\begin{subfigure}{\imgWidth}
\includegraphics[width=\textwidth]{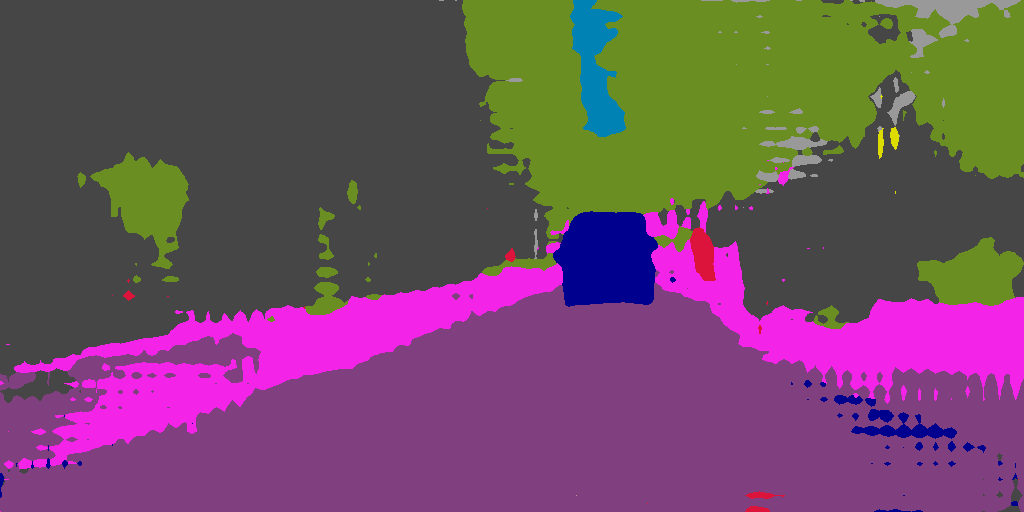}
\end{subfigure}%
\begin{subfigure}{\imgWidth}
\includegraphics[width=\textwidth]{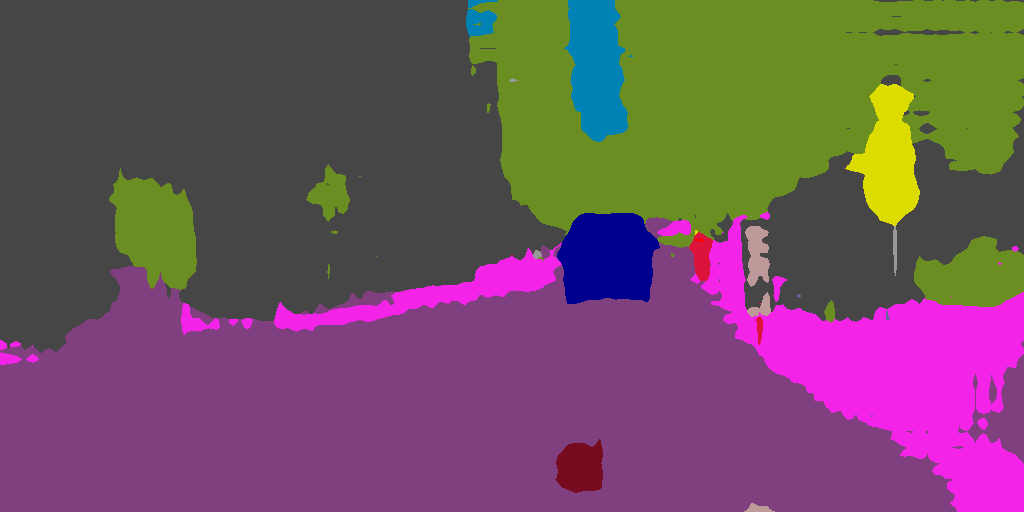}
\end{subfigure}%
\begin{subfigure}{\imgWidth}
\includegraphics[width=\textwidth]{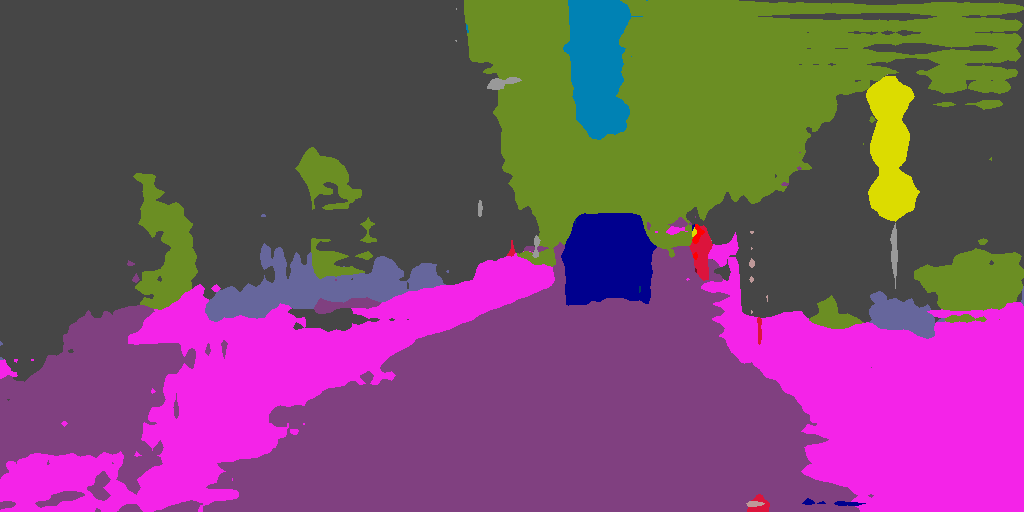}
\end{subfigure}%
\begin{subfigure}{\imgWidth}
\includegraphics[width=\textwidth]{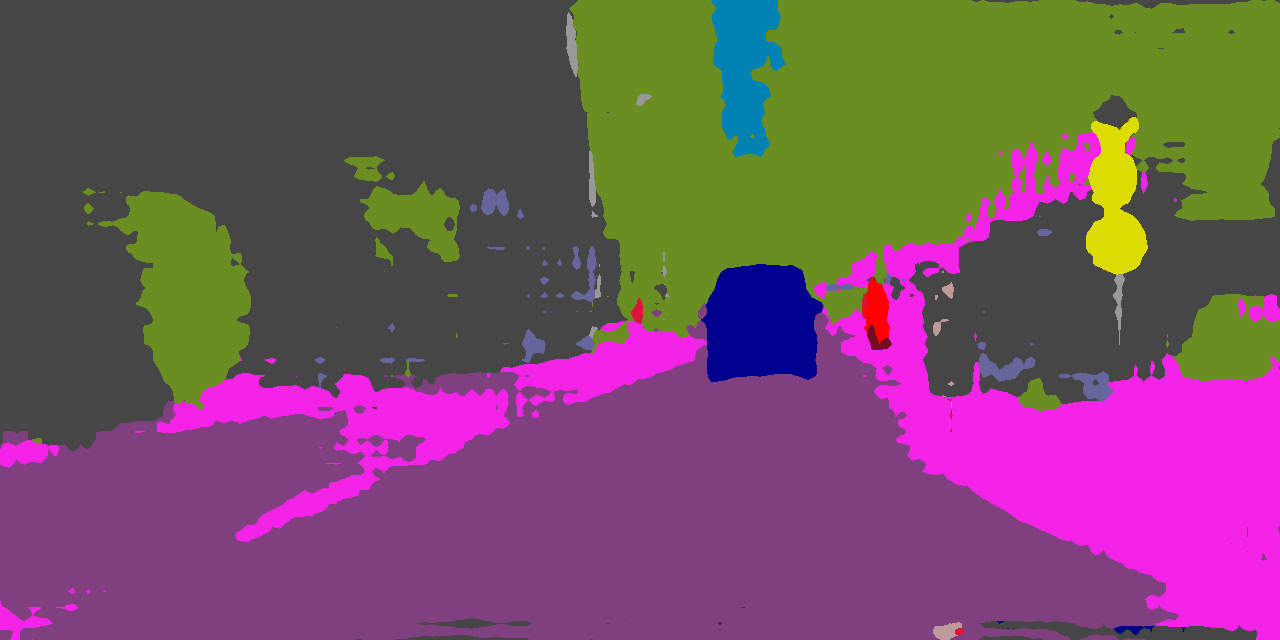}
\end{subfigure}

\hspace*{.2em}%
\begin{subfigure}{\imgWidth}
\includegraphics[width=\textwidth]{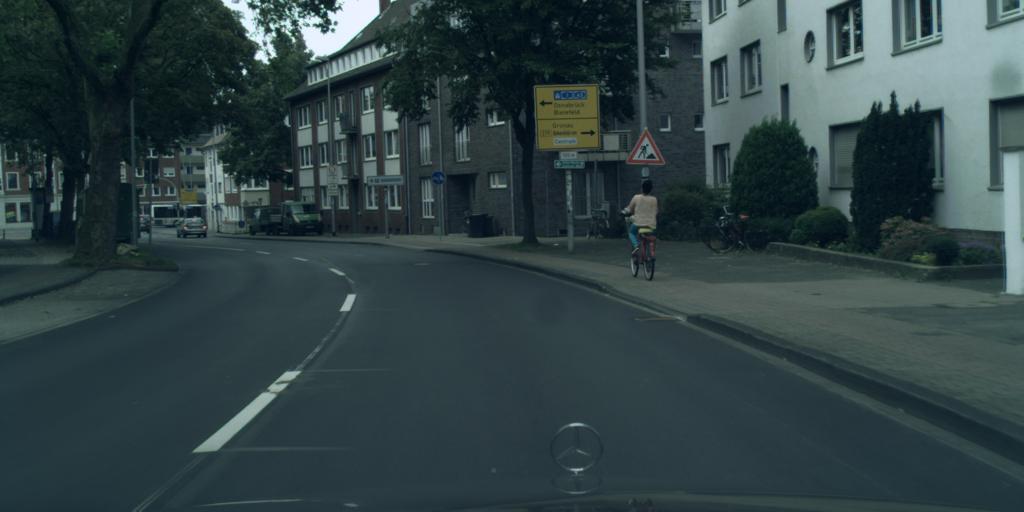}
\caption{Color image}
\end{subfigure}%
\begin{subfigure}{\imgWidth}
\includegraphics[width=\textwidth]{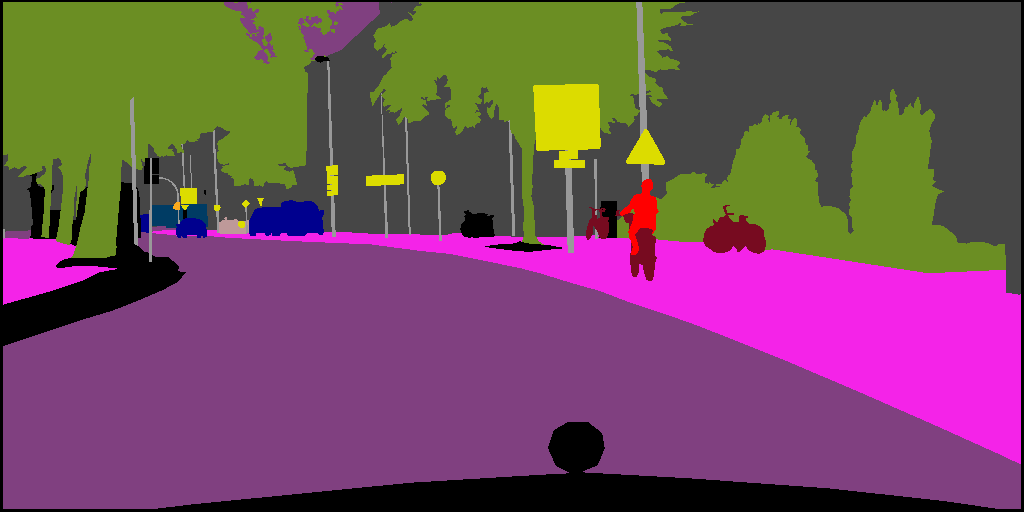}
\caption{Ground truth}
\end{subfigure}%
\begin{subfigure}{\imgWidth}
\includegraphics[width=\textwidth]{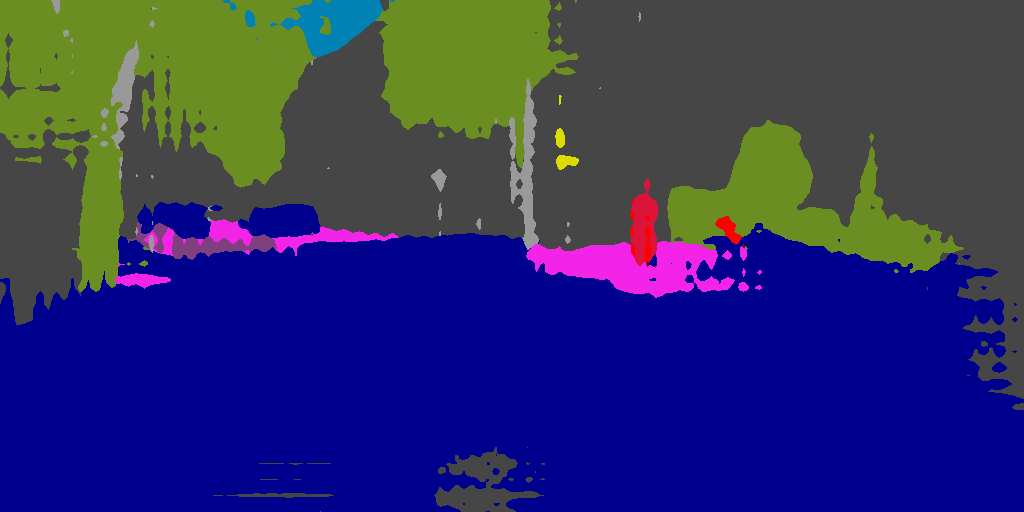}
\caption{Source only}
\end{subfigure}%
\begin{subfigure}{\imgWidth}
\includegraphics[width=\textwidth]{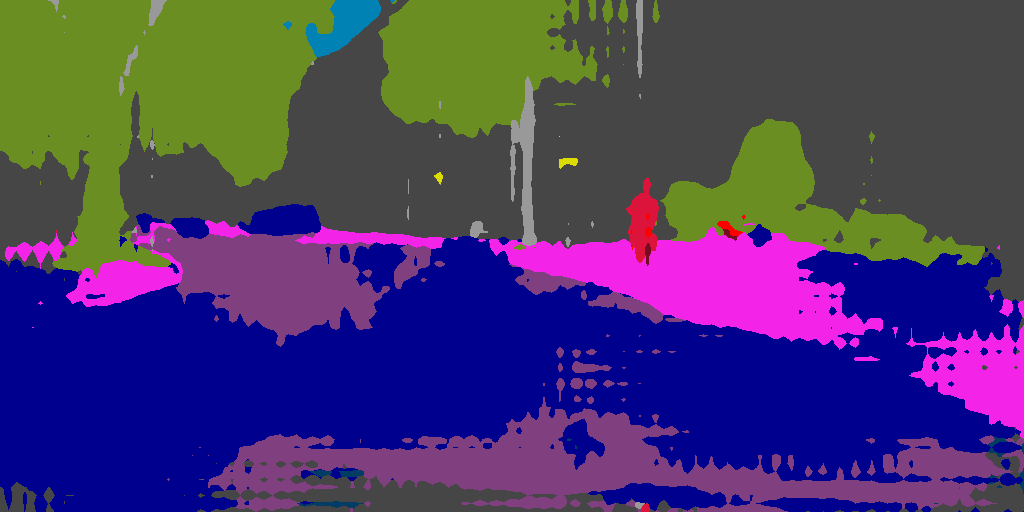}
\caption{MSIW~\cite{Chen2019}}
\end{subfigure}%
\begin{subfigure}{\imgWidth}
\includegraphics[width=\textwidth]{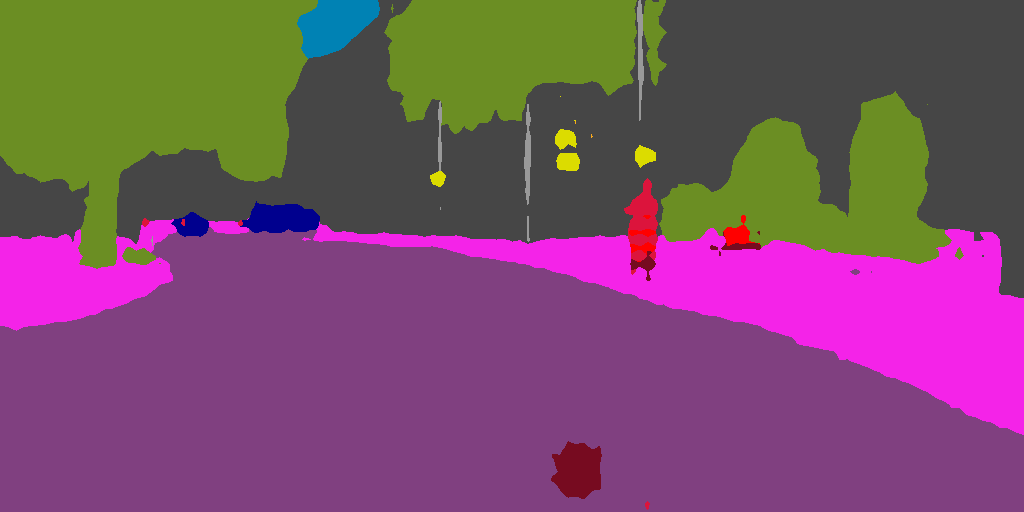}
\caption{UDA OCE~\cite{toldo2020clustering}}
\end{subfigure}%
\begin{subfigure}{\imgWidth}
\includegraphics[width=\textwidth]{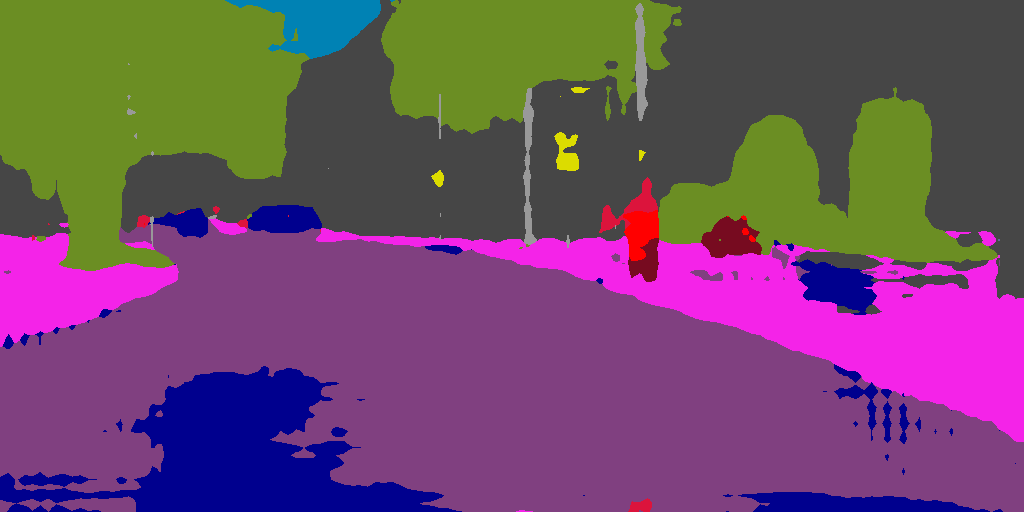}
\caption{LSR~\cite{barbato2021latent}}
\end{subfigure}%
\begin{subfigure}{\imgWidth}
\includegraphics[width=\textwidth]{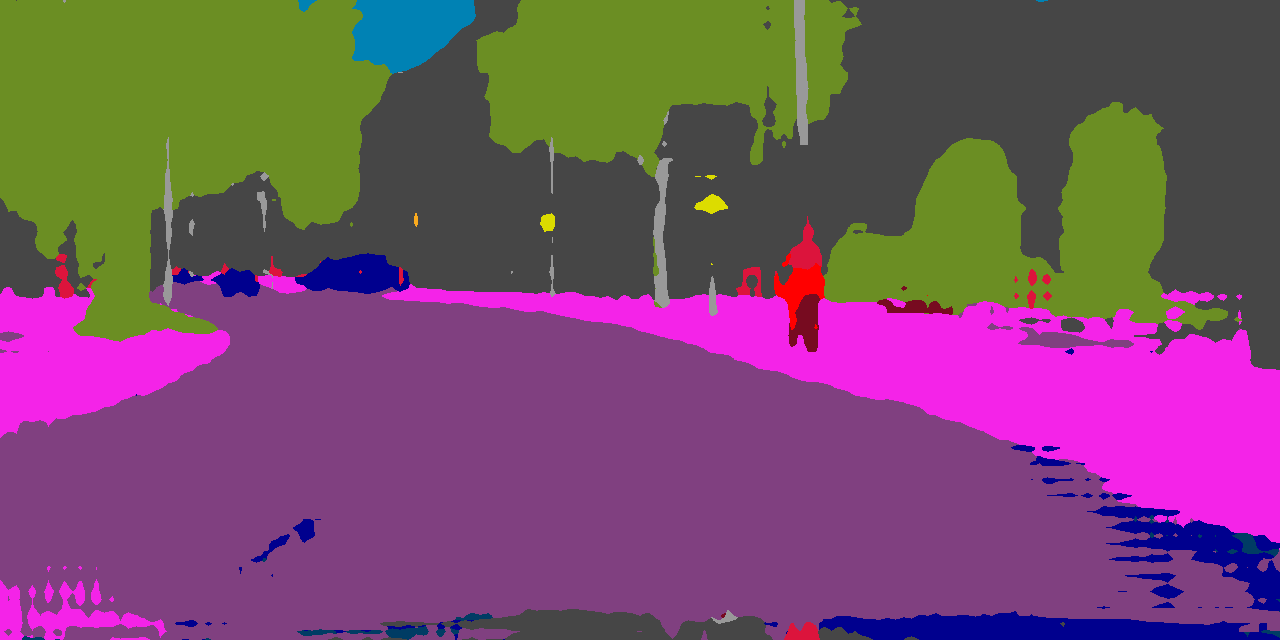}
\caption{LSR$^{+}$ (ours)}
\end{subfigure}
\end{subfigure}
\caption{Qualitative results sampled from the \textit{Cityscapes} validation split.}
\label{fig:results}
\end{figure*}

\subsection{Adaptation from Synthetic Data to Cityscapes}
\label{subsec:cityscapes}

When adapting source knowledge from the \textit{GTA5} dataset to the \textit{Cityscapes} one, our approach (LSR$^+$) achieves a mIoU of $46.9\%$, with a gain of $10\%$ compared to the baseline and of $0.9\%$ compared to the conference version (LSR) \cite{barbato2021latent}, thanks to the refined space-shaping objectives. %
Moreover, it outperforms all other strategies and the only techniques able to get close to its performance are the recent works by \cite{toldo2020clustering} and \cite{Chen2019}, while the other competitors see a significant score drop. %
The performance improvement is quite stable across %
per-class IoUs%
, and is particularly noticeable in challenging classes, like \textit{terrain} and \textit{t.\ light}
 where our strategy shows very high percentage gains, and on \textit{train} where we significantly outperform the competitors by doubling the score of the second-best strategy.

Some qualitative results are reported in the top half of Fig.~\ref{fig:results}. %
From visual inspection, %
we can verify the increased precision of edges in the \textit{t.\ sign}, \textit{t.\ light}, \textit{pole} and \textit{person} classes in both images. %
Furthermore, our approach is the only one to correctly classify the \textit{bus} on the right of the first image, which is confused as \textit{truck} by the other strategies. Importantly, we can also see the effects of our two-pass labeling (see Sec.\ \ref{sec:setup}-C) on the left of the top image (where part of the \textit{fence} is correctly classified by our strategy, while being missed by all competitors) and of the second image (where LSR$^{+}$ significantly reduces the confusion between \textit{sky} and the white building).

In the \textit{SYNTHIA} to \textit{Cityscapes} setup, LSR$^+$ surpasses its conference version (LSR) by about $1\%$ of mIoU in the 16-classes setup and by $0.6\%$ in the 13-classes one, achieving a final score of $42.6\%$ and $48.7\%$, respectively. It also outperforms all the other competitors, with a slight margin of $1\%$ on average with respect to \cite{toldo2020clustering} and a larger one (more than $3\%$) compared to all the other approaches.

Qualitative results are reported in the bottom half of Fig.~\ref{fig:results}, where the overall increase in segmentation accuracy for many classes such as \textit{car}, \textit{road} and \textit{sidewalk} is evident.
In the first image (third row of Fig. \ref{fig:results}) we can see how LSR$^+$ is the only strategy to correctly classify both \textit{rider} and \textit{bike}, whereas other strategies even miss the \textit{t. sign} in the foreground. %
Similarly, in the second image we note improvements on the prediction on such classes and, fundamentally, of the \textit{road} in foreground (confused for \textit{car} and \textit{bycicle} by the competitors).

\subsection{Adaptation from Cityscapes to Cross-City}
\label{subsec:crosscity}

Besides using synthetic data, another key requirement is the capability of adapting networks trained on road scenes coming from certain geographical areas to other regions.
However, the great variability of road scenes across the world limits a wide application of locally-trained models on a global scale. To investigate the capability of our approach to cope with this problem, we evaluate the performance on the \textit{Cross-City} real-to-real benchmark in
Table~\ref{table:crosscity}. This benchmark is comprised of $4$ cities with a completely different type of urban setting: Rome, Rio, Tokyo and Taipei. When evaluated on those setups, our strategy reaches an mIoU score of $56.2\%$, $52.3\%$, $50.0\%$ and $50.0\%$ surpassing the source only model by $5.2\%$, $3.4\%$, $2.2\%$ and $3.7\%$, respectively. 
Importantly, our approach achieves consistent results across the setups (LSR$^+$ is the top scorer in $3$ out of $4$ setups, and second in the remaining one) surpassing the average best competitor score by $0.5\%$ mIoU ($52.1\%$ versus $51.6\%$). We remark that the best competitor changes depending on the setup, being \cite{toldo2020clustering}, \cite{toldo2020clustering}, \cite{tsai2018} and \cite{tsai2018} for Rome, Rio, Tokyo and Taipei, respectively, underlining the unstable performances of many approaches usually associated with this benchmark.\\
Looking at the per-class IoU scores, we can see how our strategy significantly outperforms the competitors in \textit{t.\ light} and \textit{rider} in the \textit{Cityscapes}$\rightarrow$\textit{Rome} setup (increase of $6\%$ of IoU),  in \textit{person} and \textit{rider} in the \textit{Cityscapes}$\rightarrow$\textit{Rio} setup (increase of $4\%$ of IoU) and in \textit{motorbike} in the \textit{Cityscapes}$\rightarrow$\textit{Taipei} setup (increase of $9.3\%$ of IoU).
Qualitative results on this benchmark are presented in the Supplementary Material.

\subsection{Results with Different Backbones}
\label{subsec:backbones}

 Table~\ref{table:backbones} shows the performance of our strategy on \textit{GTAV}$\rightarrow$\textit{Cityscapes} using multiple encoder-decoder backbones in order to evaluate the generalization properties of the approach to different network architectures.
Here we can see how LSR$^+$ outperforms the source-only models (\ie, without adaptation) by $13.3\%$, $12.7\%$ and $7.8\%$ using ResNet50, VGG-16 and VGG-13, respectively. Even more importantly, we can see how the performance improvement is consistent across all backbones, in opposition to what happens to  competing strategies.
Finally, we remark the stability of the mASR score of our strategy, hovering around a mean of $57.0\%$ with a very tight standard deviation of $1.4\%$ (the mean values of the other strategies are $48.5\%$ and $42.2\%$, and the standard deviations are $2.5\%$ and $31.3\%$, respectively). Per-class IoUs are reported in the Supplementary Material.

\begin{table}
\caption{Additional quantitative results with multiple backbones, \textit{GTAV}$\rightarrow$\textit{Cityscapes} setup. (r) indicates that the strategy was re-trained, starting from the official code. }
\label{table:backbones}
\centering
\begin{tabularx}{.45\textwidth}{P{\lenA}YP{\lenC}P{\lenC}}
\toprule[1pt]
\rotatebox{45}{Backbone} & Configuration & \makebox{mIoU} & \makebox{mASR}\\
\noalign{\smallskip} 
\toprule[1pt]
\multirow{5}{*}{\rotatebox{90}{ResNet 50}} & Target only & 65.2 & 100 \\
\cdashline{2-4}
\noalign{\smallskip}
& Source only & 27.6 & 39.1 \\
& MaxSquareIW~\cite{Chen2019} (r) & 36.8 & 52.0 \\
& UDA OCE~\cite{toldo2020clustering} (r) & 36.6 & 51.7 \\
& LSR$^{+}$ (ours) & \textbf{40.9} & \textbf{58.6} \\
\toprule[1pt]
\multirow{5}{*}{\rotatebox{90}{VGG 16}} & Target only & 59.6 & 100 \\
\cdashline{2-4}
\noalign{\smallskip}
& Source only & 25.5 & 42.4 \\
& MaxSquareIW~\cite{Chen2019} (r) & 31.7 & 46.9 \\
& UDA OCE~\cite{toldo2020clustering} (r) & 34.2 & 51.5 \\
& LSR$^{+}$ (ours) & \textbf{37.2} & \textbf{57.2} \\
\toprule[1pt]
\multirow{5}{*}{\rotatebox{90}{VGG 13}} & Target only & 59.5 & 100 \\
\cdashline{2-4}
\noalign{\smallskip}
& Source only & 28.5 & 42.6 \\
& MaxSquareIW~\cite{Chen2019} (r) & 31.6 & 46.7 \\
& UDA OCE~\cite{toldo2020clustering} (r) & 16.8 & 23.3\\
& LSR$^{+}$ (ours) & \textbf{36.3} & \textbf{55.5} \\
\toprule[1pt]
\noalign{\smallskip}
\end{tabularx}
\end{table}

%% file: sections/ablation.tex
\section{Ablation Studies}
\label{sec:ablation}

In this section, we evaluate the impact of each component of the approach on the final accuracy. 
Quantitative results are reported in Table~\ref{tab:ablation}, where we evaluate our strategy by removing each constraint independently and evaluating the impact on the final accuracy. In particular, we show how the absence of each of our losses reduces the final performance by a minimum of $0.8\%$ mIoU and an average of $1\%$ mIoU. Each module brings a significant improvement in terms of accuracy and all the components are needed for the best results. Furthermore, the comparison with \cite{barbato2021latent} highlights how the improvements are distributed over all the constraints and how the novel implementation of the space shaping constraints has less overlap with respect to the entropy minimization, resulting in a much higher performance when they are employed in conjunction.

\begin{table}
\caption{Ablation studies, mIoU and mASR scores comparison when removing any of the losses. Implementations of losses  from~\cite{barbato2021latent} are compared with the new ones in this work. %
}
\label{tab:ablation}
\centering
\setlength{\tabcolsep}{.3em}
\begin{tabularx}{.37\textwidth}{cc:cc:cc:c|P{\lenB}P{\lenB}}
\multicolumn{2}{c:}{$\mathcal{L}_C$} & \multicolumn{2}{c:}{$\mathcal{L}_P$} & \multicolumn{2}{c:}{$\mathcal{L}_N$} & $\mathcal{L}_{EM}$ & \multirow{2}{*}{mIoU} & \multirow{2}{*}{mASR} \\
\cite{barbato2021latent} & ours & \cite{barbato2021latent} & ours & \cite{barbato2021latent} & ours & & &\\
\toprule
& & & & & & & 42.8 & 64.4\\
\cdashline{1-9}
& & \checkmark & & \checkmark & & \checkmark & 44.8 & - \\
& & & \checkmark & & \checkmark & \checkmark & 44.9 & 66.3\\
\cdashline{1-9}
\checkmark & & & & \checkmark & & \checkmark & 44.9 & - \\
& \checkmark & & & & \checkmark & \checkmark & 45.3 & 66.7\\
\cdashline{1-9}
\checkmark & & \checkmark & & & & \checkmark & 45.2 & - \\
& \checkmark & & \checkmark & & & \checkmark & 46.0 & 68.3\\
\cdashline{1-9}
\checkmark & &\checkmark & &\checkmark & & & 44.2 & - \\
& \checkmark & & \checkmark & & \checkmark & & 44.5 & 66.1\\
\cdashline{1-9}
\checkmark & & \checkmark & & \checkmark & & \checkmark & 46.0 & 67.7\\
& \checkmark & & \checkmark & & \checkmark & \checkmark & \textbf{46.9} & \textbf{69.5}\\
\end{tabularx}
\end{table}

\subsection{Analysis of the Latent Space Regularization}

For visualization purposes and for a fair comparative analysis across the classes, the plots of this section are computed on a balanced subset of feature vectors ($250$ vectors per class) extracted from the \textit{Cityscapes} validation set.

\begin{figure}[ht]
\centering
\includegraphics[width=0.45\textwidth]{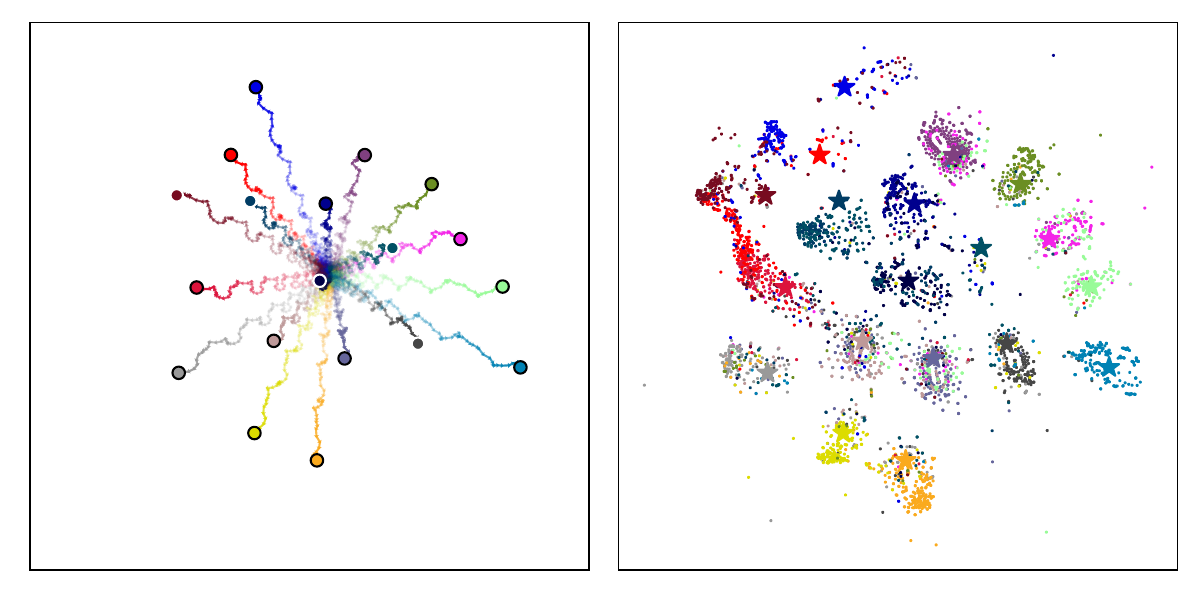}
\caption{t-SNE embedding of the target feature vectors. On the left: trajectories of prototypes sampled over $200$ training steps. On the right: features produced by the final model embedded according to the shared t-SNE projection.}
\label{fig:traj_feats}
\end{figure}
\begin{figure}[ht]
\hspace{-.7em}%
\begin{subfigure}{.17\textwidth}
\centering
\includegraphics[width=\textwidth]{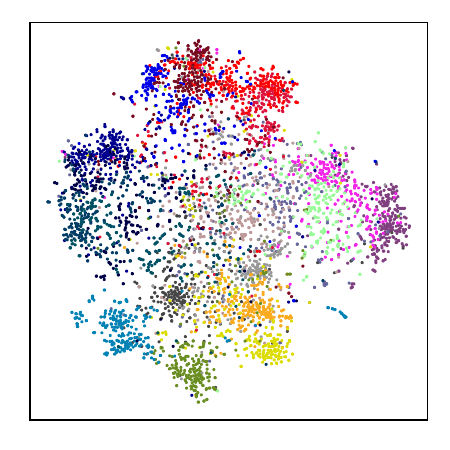}
\caption{Baseline}
\label{fig:tsne_gtab}
\end{subfigure}%
\begin{subfigure}{.17\textwidth}
\centering
\includegraphics[width=\textwidth]{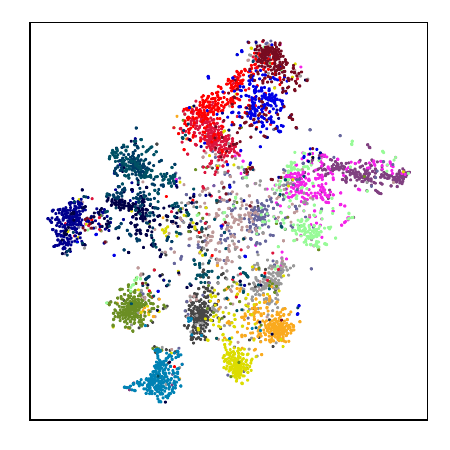}
\caption{LSR}
\label{fig:tsne_lsr}
\end{subfigure}%
\begin{subfigure}{.17\textwidth}
\centering
\includegraphics[width=\textwidth]{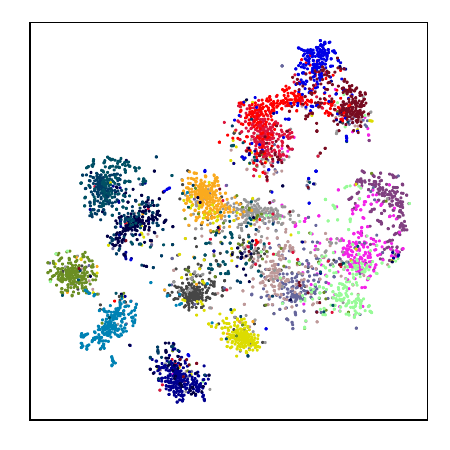}
\caption{LSR$^+$}
\label{fig:tsne_gall}
\end{subfigure}
\caption{t-SNE embeddings of the normalized feature vectors.}
\label{fig:tsne}
\end{figure}

\textbf{Two-pass prototypes and clustering.}
To investigate the semantic feature representation learning produced by our approach we computed a shared t-SNE~\cite{maaten2008visualizing} embedding of the  prototypes sampled during the training procedure and of the target features produced by the final model. We remind the reader that, in order to more effectively shift target features closer to the source ones, we resort to a two-stage label assignment procedure which recovers target awareness (by averaging target-extracted features) from prototypes computed on the source domain (by centroid computation) as reported in Sec.~\ref{sec:setup}-C.
In the left plot of Fig.~\ref{fig:traj_feats} we report the learned prototype trajectory embeddings, and on the right the respective feature vectors. Here we can appreciate how prototypes get farther apart while training goes on and how features extracted from the target domain lie in a neighborhood of the prototype, which we recall is computed exclusively via source-supervision. This underlines the effectiveness of our clustering strategy, which is able to shift the target feature distribution closer to the source one.  %

Finally, to further analyze our clustering objective we produce additional t-SNE embeddings starting from the normalized features (to remove the norm information, focusing on the angular one), which is reported in Fig.~\ref{fig:tsne}.
Our strategy significantly improves the cluster separation in the embedded space and %
increases the spacing between clusters belonging to different classes, promoting features' disentanglement. This cross-talk reduction is also reflected in the decreased probability of confusing visually similar classes (\eg, the \textit{truck} class with the \textit{bus} and \textit{train} ones). 

Finally, PCA embeddings are reported in the Supplementary Material to evaluate the effect of latent-spacing techniques when projected to a lower dimension via a linear function.

\begin{figure}[t]
\hspace*{.2em}%
\begin{subfigure}{.16\textwidth}
\centering
\includegraphics[width=\textwidth]{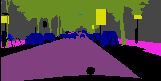}
\caption{\footnotesize Nearest}
\end{subfigure}%
\begin{subfigure}{.16\textwidth}
\centering
\includegraphics[width=\textwidth]{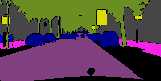}
\caption{\footnotesize Histogram}
\end{subfigure}%
\begin{subfigure}{.16\textwidth}
\centering
\includegraphics[width=\textwidth]{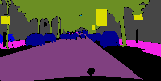}
\caption{\footnotesize Weighted Histogram}
\end{subfigure}
\caption{Sample image downsampled nearest (left), frequency-aware \cite{barbato2021latent} (middle) or weighted frequency-aware (LSR$^+$).}
\label{fig:down}
\end{figure}

\begin{figure}[t]
\includegraphics[width=.45\textwidth]{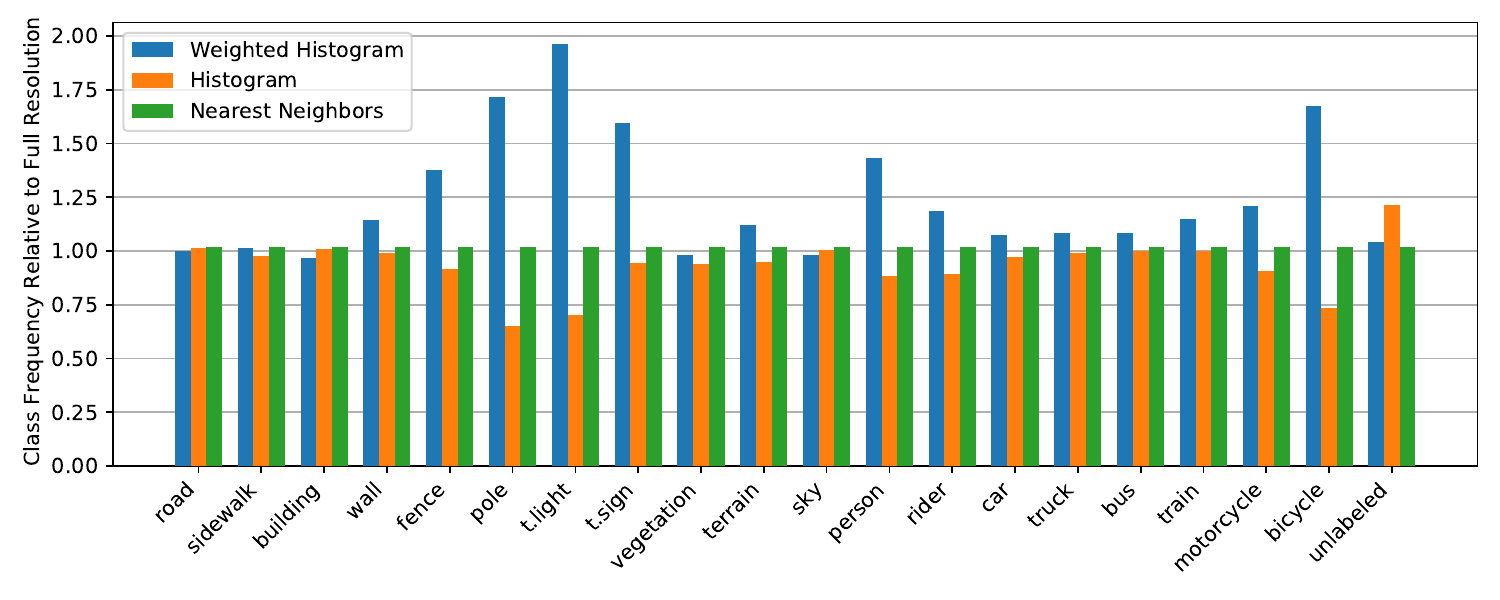}
\caption{Class frequency of the downsampled feature-level segmentation maps.}
\label{fig:freq}
\end{figure}

\textbf{Weighted histogram-aware downsampling.} In this work, we extended the scheme proposed in \cite{barbato2021latent} by adding class weights inversely proportional to the class-frequency in the training dataset (see Sec.~\ref{sec:setup}). 
Our goal is to provide labeling only to spatial locations in feature maps where a clear class association can be performed, by relying on a frequency-aware scheme. 
By doing so, we seek for the disentanglement of activations belonging to different classes, even when their feature vectors are neighbors in a given label map.
This effect can be noted in Fig.~\ref{fig:down}, where our downsampling algorithms enhanced with frequency-awareness are able to identify some feature locations close to class edges as \textit{unlabeled} in the downsampled label map (middle and right),
keeping only faithful features. As expected, class-weighting (right plot of Fig.~\ref{fig:down}) promotes rarer classes at the feature level compared to the version without it \cite{barbato2021latent} (middle plot of Fig.~\ref{fig:down}): for instance, compare the traffic sign (in yellow).
Further evidence of this can be found in the class distribution of segmentations maps (computed after their downsampling to the latent space spatial resolution), which we
reported in Fig.~\ref{fig:freq} for our weighted histogram-aware scheme, the previous un-weighted histogram-aware scheme of the conference version \cite{barbato2021latent} and the standard nearest neighbor. In particular, the schemes based on histogram-awareness generally seldom preserve small object classes, promoting \textit{unlabeled} classification when discrimination between classes is uncertain. Our weighted histogram-aware scheme improves uniformity across rarer or smaller semantic categories, which were over-penalized by the previous approach \cite{barbato2021latent}, where all classes were treated equally, regardless of their occurrence.

\begin{figure}[ht]
\centering
\includegraphics[width=0.45\textwidth]{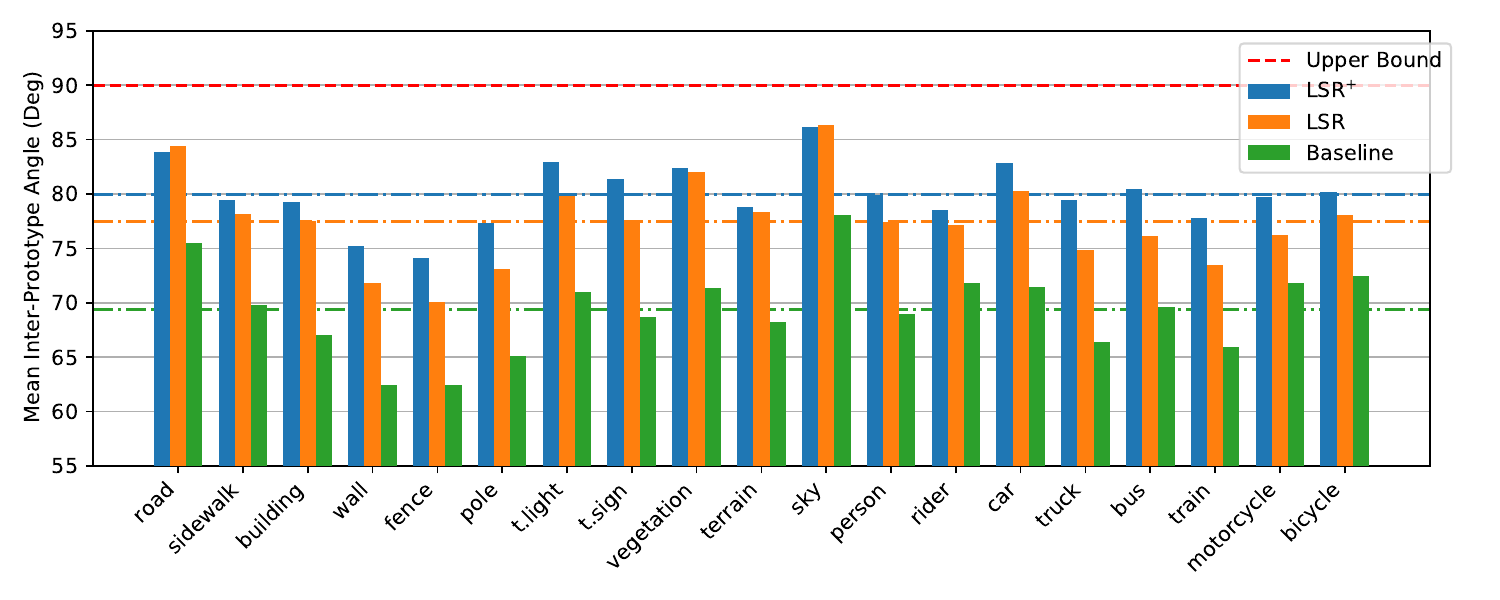}
\caption{Mean inter-prototype angle.}
\label{fig:perp}
\end{figure}
\begin{figure}[ht]
\centering
\includegraphics[width=0.45\textwidth]{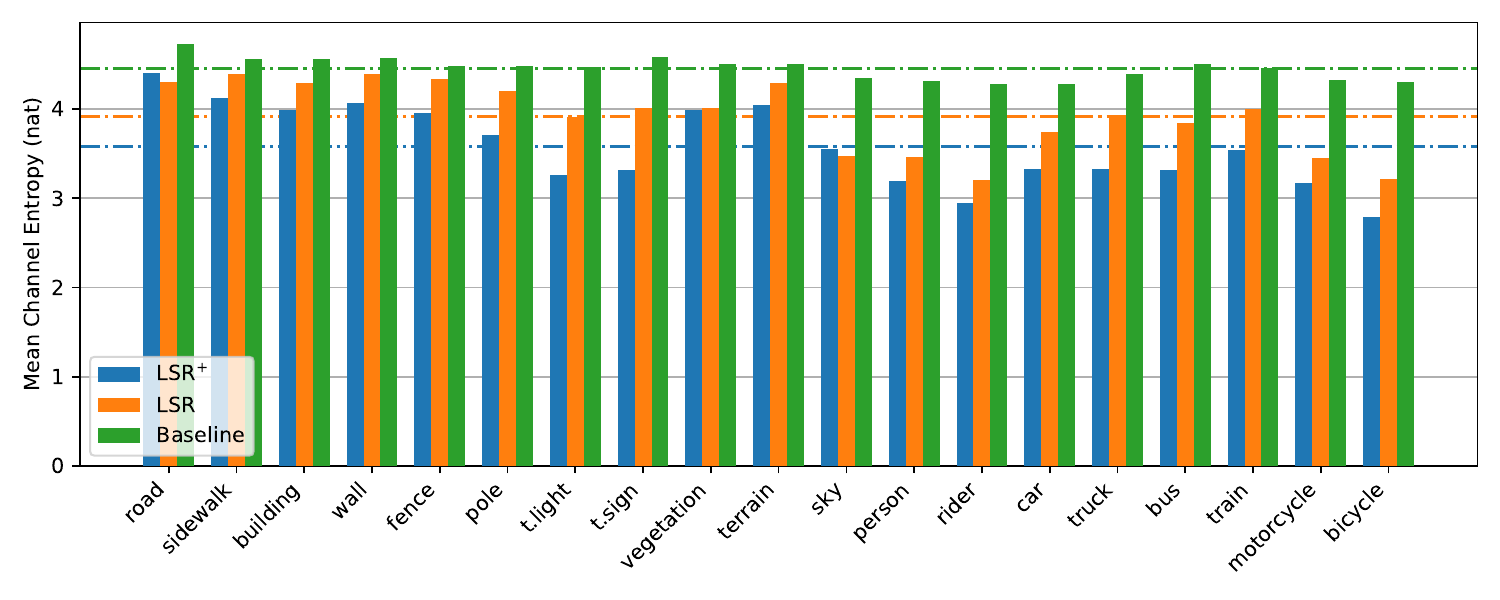}
\caption{Average channel entropy.}
\label{fig:entropy}
\end{figure}

\textbf{Perpendicularity} is analyzed in Fig.~\ref{fig:perp} 
where we display the average angular distance between each prototypes and all the remaining ones.
Our goal is to achieve prototype perpendicularity, such that we minimize the the overlap (\ie, cross-talk) among distinct semantic categories over feature activations.
By the red dashed line we highlight the upper bound to the angular distance, which is set to $90$ degrees since we assume feature vectors to have non-negative entries.
From the figure, it emerges clearly that LSR-based approaches increase the inter-prototypical angle and that LSR$^+$ makes prototypes even more orthogonal with an improvement of more than $2$ degrees on average.

\textbf{Norm Alignment} is analyzed in Fig.~\ref{fig:entropy}, where we show the mean channel entropy for each class. %
We observe that the entropy corresponding to feature vectors produced by LSR$^{+}$ is significantly reduced, meaning that features are characterized by more relevant peaks and fewer poorly-activated channels.

%% file: sections/conclusions.tex
\section{Conclusions}
\label{subsec:conclusions}
In this work, we tackled the generalization of road scene segmentation models by introducing a set of latent-space regularization strategies for unsupervised domain adaptation. 
We improved domain invariance using different latent space-shaping constraints (\ie, class clustering, class perpendicularity and norm alignment), to space apart features belonging to different classes while clustering together features of the same class in a consistent way on both the source and target domain. To support their computation, we introduced a novel target pseudo-labeling scheme and a weighted label decimation strategy. %
Results have been evaluated using both the standard mIoU and a novel metric (mASR), which captures the relative performance between an adapted model and its target supervised counterpart. We outperformed state-of-the-art methods in feature-level adaptation on two widely used synthetic-to-real road scenes benchmarks and in  real-to-real setups, paving the way to a new set of feature-level adaptation strategies capable to  improve the discrimination ability of road scene understanding approaches. 

Future work will focus on designing novel feature-level techniques and on evaluating their capability of generalizing  to various tasks in driving scenarios. The adaptation from multiple source domains to multiple target ones will also be considered together with the application also to multimodal data (\textit{e.g.}, LIDARs or depth cameras) mounted on cars. 

%% file: sections/appendixA.tex
This document contains some supplementary results supporting our work from both a quantitative and qualitative perspective.

\section{Additional Quantitative Results}
\label{sec:quantitative}

In Table~\ref{table:backbones} we report the extended version of Table III of the main paper, where we analyzed the stability of our strategy by varying the backbone network (we used the ResNet50 \cite{he2016deep}, ResNet101 \cite{he2016deep}, VGG16 \cite{simonyan2015very} and VGG13 \cite{simonyan2015very} models).
From this table we can see how the consistent performance of LSR$^+$ is preserved even in the class-wise IoU scores, particularly in the \textit{train} class, where we significantly outperform the competition, gaining an average of $7\%$ IoU across the four backbones with respect to the second best strategy.

\section{Additional Qualitative Results}
\label{sec:qualitative}

\begin{figure*}
\centering
\includegraphics[width=\textwidth]{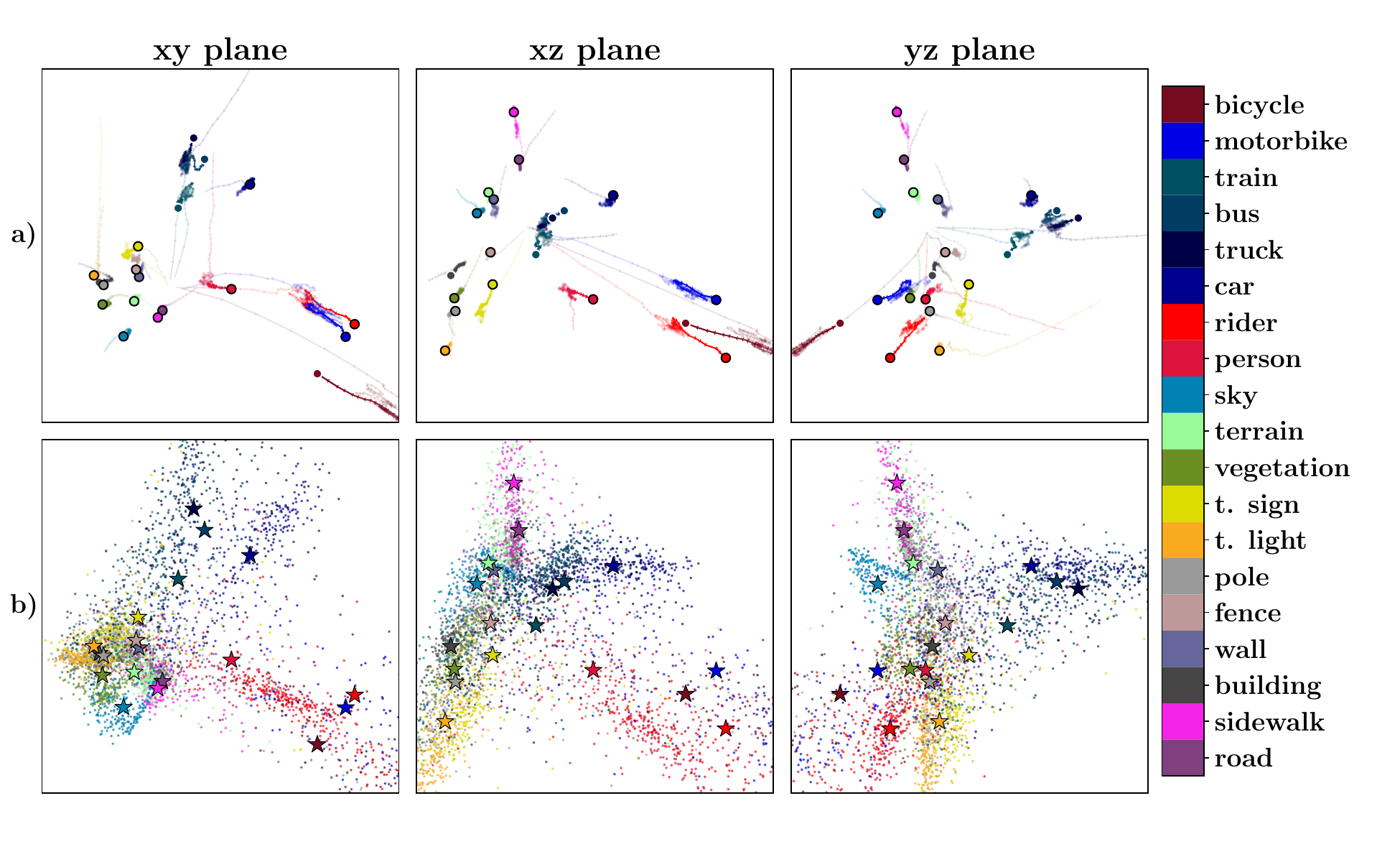}
\caption{Prototypes trajectories and target feature vectors projected via PCA. Projection is 3-dimensional; here we report the three $xy$, $xz$ and $yz$ planes.}
\label{fig:traj_feats}
\end{figure*}

In Fig.~\ref{fig:traj_feats} we report the PCA counterpart of Fig.~$5$ of the main paper. Here we show how the distancing of the prototypes and source-target alignment is preserved even when projected using a linear function, as opposed to the non-linear t-SNE. In particular, Fig.~\ref{fig:traj_feats}a) reports the prototypes trajectories, while Fig.~\ref{fig:traj_feats}b) reports the target vectors embedding. 

In Fig.~\ref{fig:crosscity} we report some qualitative results on the Cross-City benchmark. Here we present two images for each city (Rome, Rio, Tokyo, Taipei) and compare our strategy with three other strategies (Source only, MaxSquareIW~\cite{Chen2019} and UDA OCE~\cite{toldo2020clustering}).\\
From a visual inspection of the images we can see an overall increase in the discrimination of the object borders, particularly for  classes such as \textit{car}, \textit{road}, \textit{building}, \textit{vegetation} and \textit{person}.
In Rome we see how LSR$^+$ is the only strategy that correctly identifies the \textit{rider} behind the cars in the second image.
In Rio, our architecture significantly reduces the amount of confusion regarding the \textit{building} on the left of the second image.
Again, in Tokyo, we note how LSR$^+$ is the only technique able to recognize the \textit{traffic sign} on the right of the first image.
Finally, in Taipei, we see how our approach is the only to correctly identify the \textit{person} and \textit{motorcycle} in the second image.

In Fig.~\ref{fig:cityscapes} we report some qualitative results on the synthetic-to-real GTAV$\rightarrow$Cityscapes and SYNTHIA$\rightarrow$Cityscapes benchmarks. Here we present three images per setup and compare our strategy with three other approaches (Source only, MaxSquareIW~\cite{Chen2019}, UDA OCE~\cite{toldo2020clustering} and our conference work \cite{barbato2021latent}).\\
In the GTAV$\rightarrow$Cityscapes setup, we can see an overall improvement in the precision, particularly along the borders of objects.
In particular, we can see how LSR$^+$ significantly improves the prediction of the \textit{road} class in the second image, and is one of the only two that correctly identifies the \textit{wall} on the left of the third image, doing so much closer to the ground truth than the competitor.
In SYNTHIA$\rightarrow$Cityscapes we observe the same overall improvement in precision as seen in all other benchmarks. More in detail, we see how LSR$^+$ is the only strategy that correctly identifies the \textit{pole} on the right of the first image and the \textit{bicycles} on the left of the second image. Finally, on the third image, we can see how our strategy is the only one that recognizes the \textit{traffic sign} on the far right.

\newcolumntype{P}[1]{>{\fontsize{8}{8}\selectfont\centering\arraybackslash}p{#1}}
\newcolumntype{Q}[1]{>{\fontsize{8}{8}\selectfont}p{#1}}
\newcolumntype{Y}[1]{>{\fontsize{8}{8}\selectfont\centering\arraybackslash}X{#1}}
\renewcommand{\lenA}{.45em}
\renewcommand{\lenB}{.8em}
\renewcommand{\lenC}{1.5em}
\begin{table*}[ht]
\caption{Per-class IoU, mIoU and mASR with multiple backbones.}
\label{table:backbones}
\linespread{0.7}\selectfont\centering
\begin{tabularx}{\textwidth}{P{\lenA}YQ{\lenA}Q{\lenA}Q{\lenA}Q{\lenA}Q{\lenA}Q{\lenA}Q{\lenA}Q{\lenA}Q{\lenA}Q{\lenA}Q{\lenA}Q{\lenA}Q{\lenA}Q{\lenA}Q{\lenA}Q{\lenA}Q{\lenA}Q{\lenA}Q{\lenB}P{\lenC}P{\lenC}}
\toprule[1pt]
\rotatebox{45}{Backbone} & Configuration & \rotatebox{45}{Road} & \rotatebox{45}{Sidewalk} & \rotatebox{45}{Building} & \rotatebox{45}{Wall\textsuperscript{1}} & \rotatebox{45}{Fence\textsuperscript{1}} & \rotatebox{45}{Pole\textsuperscript{1}} & \rotatebox{45}{Traffic Light} & \rotatebox{45}{Traffic Sign} & \rotatebox{45}{Vegetation} & \rotatebox{45}{Terrain} & \rotatebox{45}{Sky} & \rotatebox{45}{Person} & \rotatebox{45}{Rider} & \rotatebox{45}{Car} & \rotatebox{45}{Truck} & \rotatebox{45}{Bus} & \rotatebox{45}{Train} & \rotatebox{45}{Motorbike} & \rotatebox{45}{Bicycle} & \makebox{mIoU} & \makebox{mASR}\\
\noalign{\smallskip}
\toprule[1pt]

\multirow{5}{*}{\rotatebox{90}{ResNet 101}} & Target Only & 96.5 & 73.8 & 88.4 & 42.2 & 43.7 & 40.7 & 46.1 & 58.6 & 88.5 & 54.9 & 91.9 & 68.7 & 46.2 & 90.7 & 68.8 & 69.9 & 48.8 & 47.6 & 64.5 & 64.8 & 100 \\
\cdashline{2-23}
\noalign{\smallskip}
\footnotesize
& Source only~\cite{toldo2020clustering} & 71.4 & 15.3 & 74.0 & 21.1 & 14.4 & 22.8 & 33.9 & 18.6 & 80.7 & 20.9 & 68.5 & 56.6 & 27.1 & 67.4 & 32.8 & \phantomDigit{5.6} & \phantomDigit{7.7} & 28.4 & 33.8 & 36.9 & 54.0\\
& MaxSquareIW~\cite{Chen2019} & 87.7 & 25.2 & 82.9 & 30.9 & 24.0 & 29.0 & 35.4 & 24.2 & 84.2 & 38.2 & 79.2 & 59.0 & 27.7 & 79.5 & 34.6 & 44.2 & \phantomDigit{7.5} & 31.1 & 40.3 & 45.5 & 62.2 \\
& UDA OCE~\cite{toldo2020clustering} & 89.4 & 30.7 & 82.1 & 23.0 & 22.0 & 29.2 & 37.6 & 31.7 & 83.9 & 37.9 & 78.3 & 60.7 & 27.4 & 84.6 & 37.6 & 44.7 & \phantomDigit{7.3} & 26.0 & 38.9 & 45.9 & 67.3 \\
& LSR$^{+}$ (ours) & 88.9 & 26.6 & 82.0 & 21.0 & 24.4 & 30.1 & 41.1 & 27.0 & 84.7 & 42.7 & 80.1 & 63.0 & 26.4 & 83.1 & 30.4 & 44.3 & 16.8 & 35.8 & 42.4 & \textbf{46.9} & \textbf{69.5} \\
\toprule[1pt]

\multirow{5}{*}{\rotatebox{90}{ResNet 50}} & Target Only & 97.0 & 76.6 & 88.6 & 48.0 & 43.5 & 45.0 & 47.3 & 61.4 & 89.2 & 52.8 & 92.0 & 69.0 & 46.9 & 90.8 & 58.2 & 69.6 & 49.2 & 48.9 & 65.0 & 65.2 & 100 \\
\cdashline{2-23}
\noalign{\smallskip}
\footnotesize
& Source only & 57.4 & \phantomDigit{8.9} & 74.7 & 13.8 & 21.7 & 23.0 & 24.2 & 11.6 & 73.5 & 10.3 & 66.8 & 48.6 & 13.1 & 31.6 & 16.6 & \phantomDigit{8.5} & \phantomDigit{0.0} & 15.8 & \phantomDigit{5.0} & 27.6 & 39.1\\
& MaxSquareIW~\cite{Chen2019} & 83.6 & 17.3 & 79.4 & 20.8 & 17.9 & 24.4 & 27.2 & 15.6 & 80.1 & 31.5 & 79.9 & 58.6 & 21.3 & 67.5 & 25.1 & 12.8 & \phantomDigit{1.8} & 19.7 & 14.3 & 36.8 & 52.0 \\
& UDA OCE~\cite{toldo2020clustering} & 83.9 & 14.4 & 80.0 & 22.8 & 20.8 & 24.7 & 26.4 & 16.5 & 80.3 & 31.7 & 80.1 & 57.9 & 21.6 & 72 & 26.7 & 14.5 & \phantomDigit{3.6} & 13.2 & \phantomDigit{3.6} & 36.6 & 51.7 \\
& LSR$^{+}$ (ours) & 88.6 & 31.1 & 79.5 & 23.9 & 24.1 & 26.8 & 30.8 & 15.9 & 84.5 & 33.0 & 75.7 & 55.0 & 17.6 & 80.0 & 23.4 & 34.6 & 20.3 & 17.5 & 15.4 & \textbf{40.9} & \textbf{58.6} \\
\toprule[1pt]
\multirow{5}{*}{\rotatebox{90}{VGG 16}} & Target Only & 96.5 & 73.7 & 86.8 & 39.3 & 41.2 & 35.2 & 40.5 & 51.4 & 87.4 & 49.4 & 89.1 & 63.8 & 40.5 & 88.8 & 46.2 & 63.5 & 37.2 & 41.4 & 60.6 & 59.6 & 100\\
\cdashline{2-23}
\noalign{\smallskip}
\footnotesize
& Source only & 26.5 & 13.3 & 45.1 & \phantomDigit{6.0} & 15.2 & 16.5 & 21.3 & \phantomDigit{8.5} & 78.0 & \phantomDigit{8.3} & 59.7 & 45.0 & 10.5 & 69.1 & 22.8 & 17.9 & \phantomDigit{0.0} & 16.4 & \phantomDigit{2.7} & 25.4 & 38.7 \\
& MaxSquareIW~\cite{Chen2019} & 81.4 & 20.0 & 75.4 & 19.4 & 19.1 & 16.1 & 24.4 & \phantomDigit{7.9} & 78.8 & 22.9 & 65.9 & 45.0 & 12.3 & 74.6 & 16.1 & 10.3 & \phantomDigit{0.2} & 11.3 & \phantomDigit{1.0} & 31.7 & 46.9 \\
& UDA OCE~\cite{toldo2020clustering} & 86.0 & 13.5 & 79.4 & 20.4 & 18.5 & 21.5 & 27.6 & 15.2 & 80.8 & 21.9 & 72.6 & 46.3 & 18.1 & 80.0 & 16.9 & 13.1 & \phantomDigit{1.0} & 14.6 & \phantomDigit{2.0} & 34.2 & 51.5 \\
& LSR$^{+}$ (ours) & 88.9 & 32.2 & 79.3 & 25.9 & 23.6 & 26.9 & 30.6 & 10.7 & 83.3 & 33 & 73.1 & 47.6 & 20.5 & 81.3 & 18.2 & 10.7 & \phantomDigit{0.1} & 14.7 & \phantomDigit{5.9} & \textbf{37.2} & \textbf{57.2} \\
\toprule[1pt]
\multirow{5}{*}{\rotatebox{90}{VGG 13}} & Target Only & 96.5 & 74.3 & 86.5 & 34.2 & 41.1 & 35.9 & 39.5 & 51.7 & 87.4 & 52.8 & 89.2 & 63.1 & 39.1 & 88.3 & 44.2 & 61.9 & 43.6 & 40.1 & 61.5 & 59.5 & 100\\
\cdashline{2-23}
\noalign{\smallskip}
\footnotesize
& Source only & 62.3 & 15.1 & 67.8 & 12.1 & 29.8 & 16.6 & 19.1 & \phantomDigit{6.5} & 75.8 & 12.8 & 75.5 & 48.5 & \phantomDigit{4.6} & 60.8 & 16.2 & \phantomDigit{3.6} & \phantomDigit{0.0} & 12.8 & \phantomDigit{1.8} & 28.5 & 42.4\\
& MaxSquareIW~\cite{Chen2019} & 82.3 & \phantomDigit{3.8} & 78.8 & 16.2 & 31.4 & 12.1 & 18.5 & \phantomDigit{4.7} & 79.9 & 28.6 & 74.9 & 42.6 & \phantomDigit{2.8} & 79.4 & 23.2 & 10.7 & \phantomDigit{0.0} & 10.9 & \phantomDigit{0.3} & 31.6 & 46.7\\
& UDA OCE~\cite{toldo2020clustering} & \phantomDigit{0.7} & \phantomDigit{0.3} & 68.0 & \phantomDigit{0.1} & 11.9 & \phantomDigit{6.0} & 18.4 & \phantomDigit{4.7} & 76.1 & 10.9 & 75.2 & 17.5 & \phantomDigit{2.2} & 22.0 & \phantomDigit{0.4} & \phantomDigit{0.6} & \phantomDigit{0.0} & \phantomDigit{3.9} & \phantomDigit{0.0} & 16.8 & 23.3\\
& LSR$^{+}$ (ours) & 86.4 & 33.1 & 79.8 & 26.2 & 17.6 & 23.6 & 23.5 & \phantomDigit{7.6} & 80.4 & 27.9 & 70.8 & 50.5 & 12.0 & 82.0 & 24.9 & 26.3 & \phantomDigit{2.8} & 12.0 & \phantomDigit{3.0} & \textbf{36.3} & \textbf{55.5} \\
\toprule[1pt]
\noalign{\smallskip}
\end{tabularx}
\end{table*}

\renewcommand{\imgWidth}{0.165\textwidth+.09em}
\begin{figure*}
\newcolumntype{Y}{>{\centering\arraybackslash}X}
\centering

\begin{subfigure}{\textwidth}
\small
\begin{tabularx}{\textwidth}{YYYYYYY}
\cellcolor{road} \textcolor{white}{Road} & \cellcolor{sidewalk} Sidewalk & \cellcolor{building} \textcolor{white}{Building} & \cellcolor{tlight} T. Light & \cellcolor{tsign} T. Sign & \cellcolor{vegetation} \textcolor{white}{Vegetation} & \cellcolor{sky} Sky \\ \cellcolor{person} \textcolor{white}{Person} & \cellcolor{rider} \textcolor{white}{Rider} & \cellcolor{car} \textcolor{white}{Car} & \cellcolor{bus} \textcolor{white}{Bus} & \cellcolor{motorbike} \footnotesize \textcolor{white}{Motorbike} & \cellcolor{bicycle} \textcolor{white}{Bicycle} & \cellcolor{unlabelled} \textcolor{white}{Unlabeled}
\end{tabularx}
\end{subfigure}

\begin{subfigure}{.6em}
\scriptsize\rotatebox{90}{Rome}
\end{subfigure}%
\begin{subfigure}{\textwidth-1em}
\hspace*{.2em}%
\begin{subfigure}{\imgWidth}
\includegraphics[width=\textwidth]{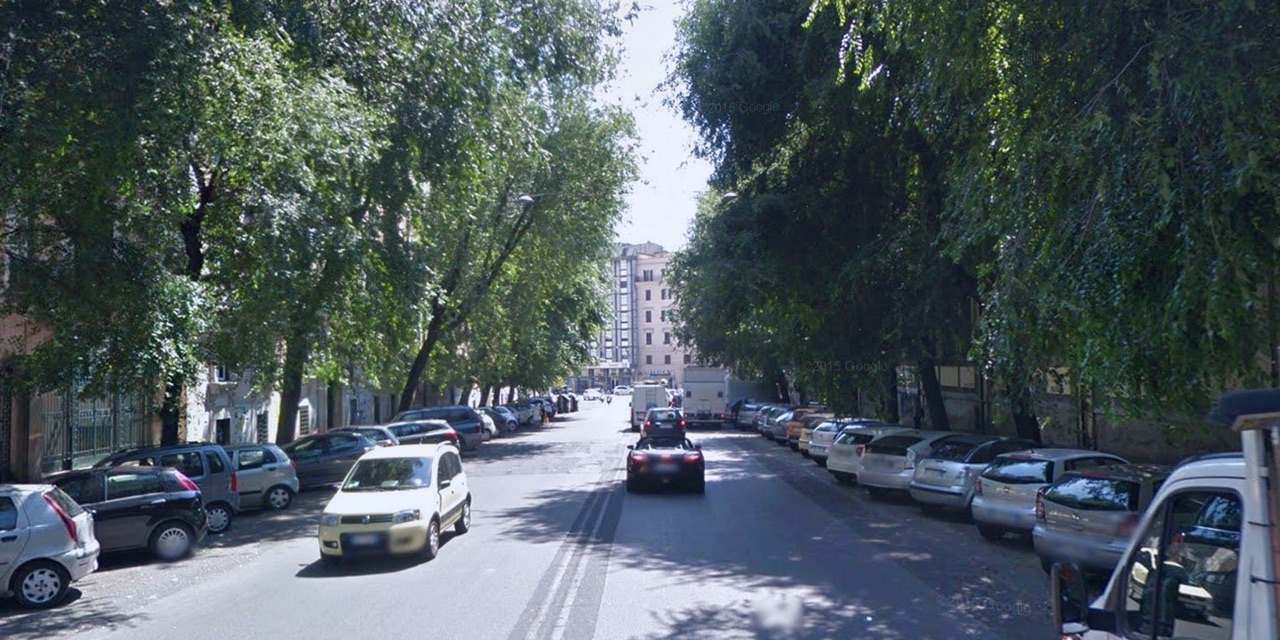}
\end{subfigure}%
\begin{subfigure}{\imgWidth}
\includegraphics[width=\textwidth]{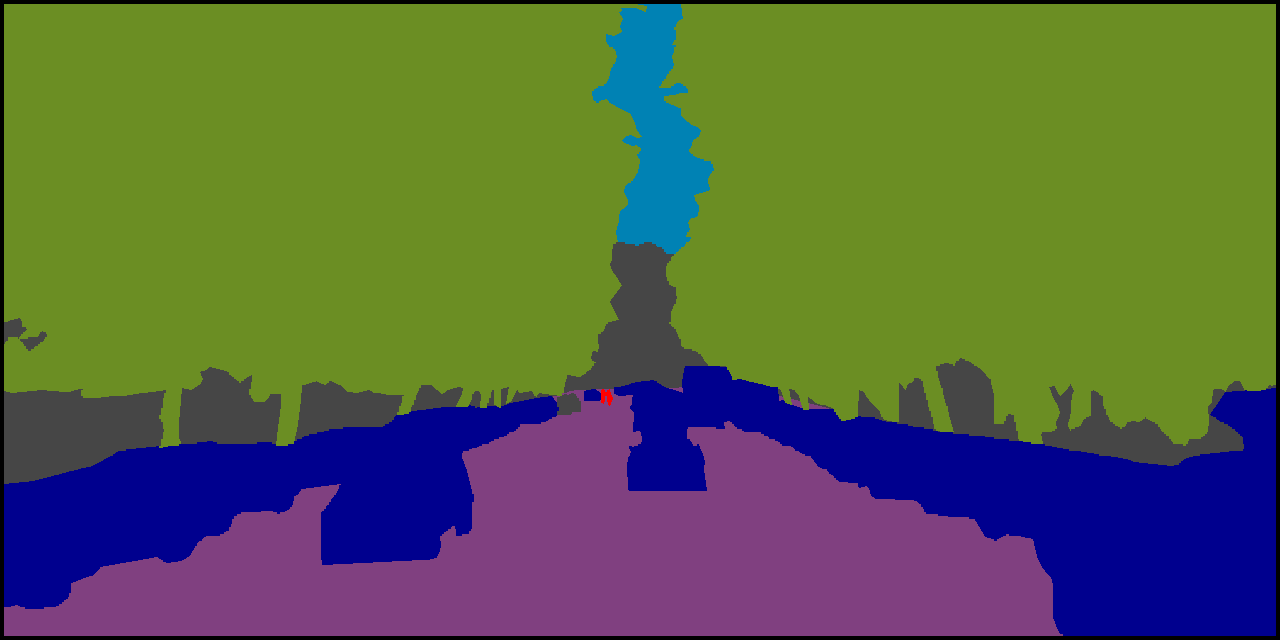}
\end{subfigure}%
\begin{subfigure}{\imgWidth}
\includegraphics[width=\textwidth]{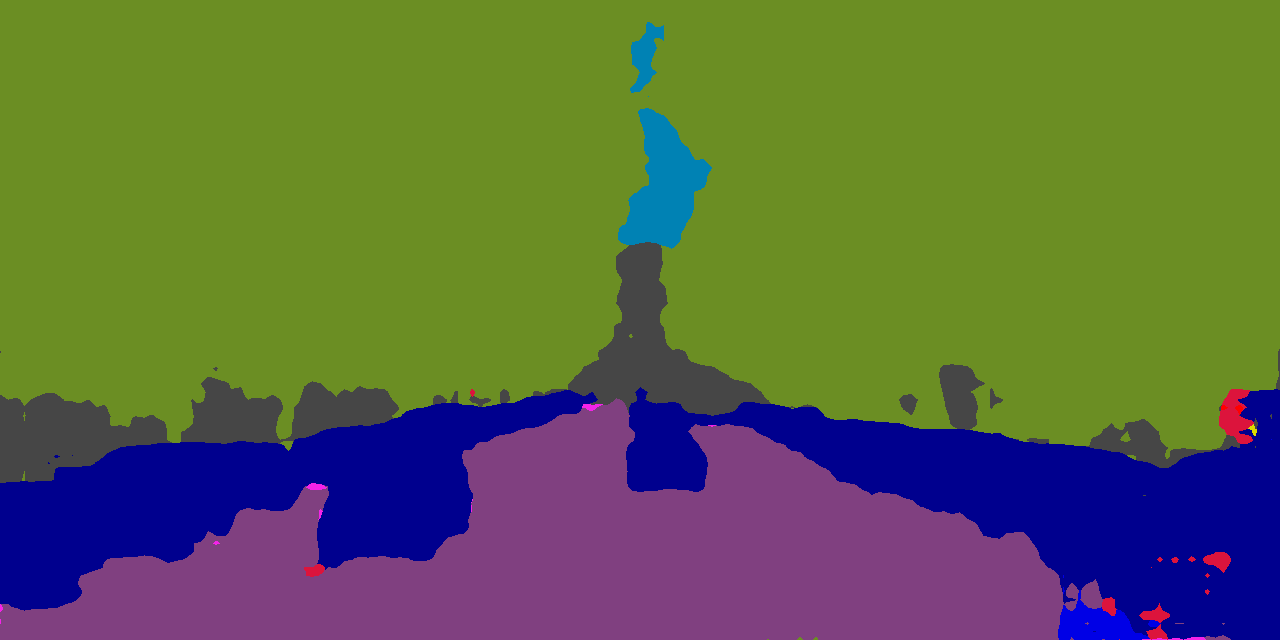}
\end{subfigure}%
\begin{subfigure}{\imgWidth}
\includegraphics[width=\textwidth]{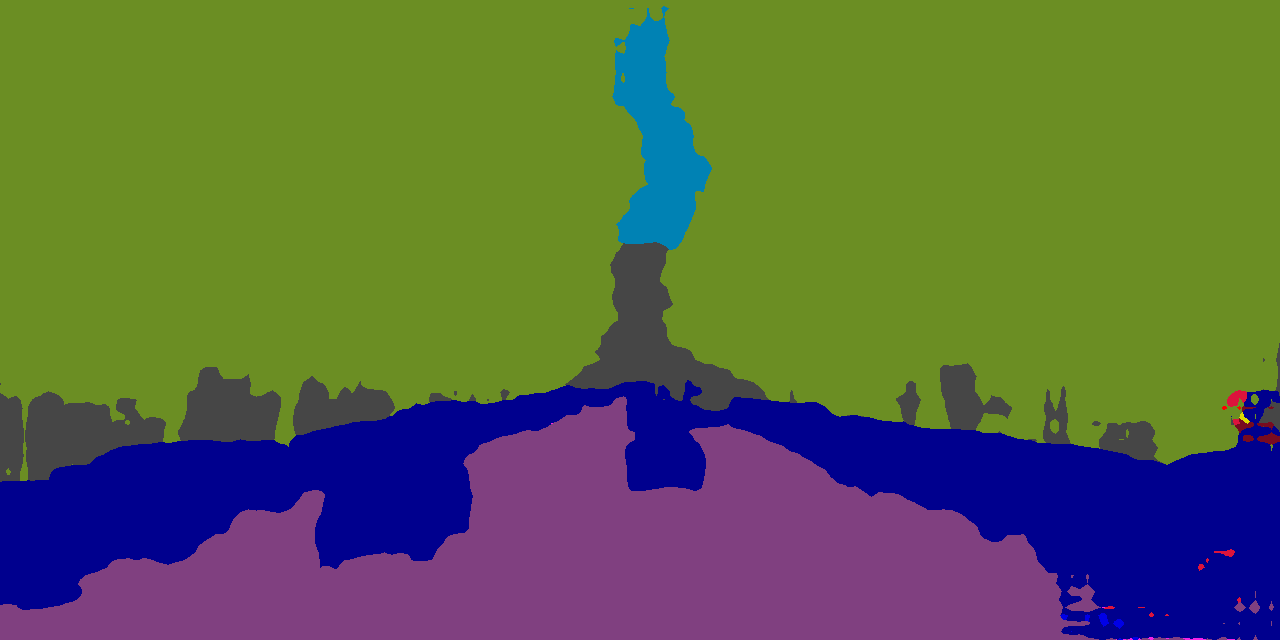}
\end{subfigure}%
\begin{subfigure}{\imgWidth}
\includegraphics[width=\textwidth]{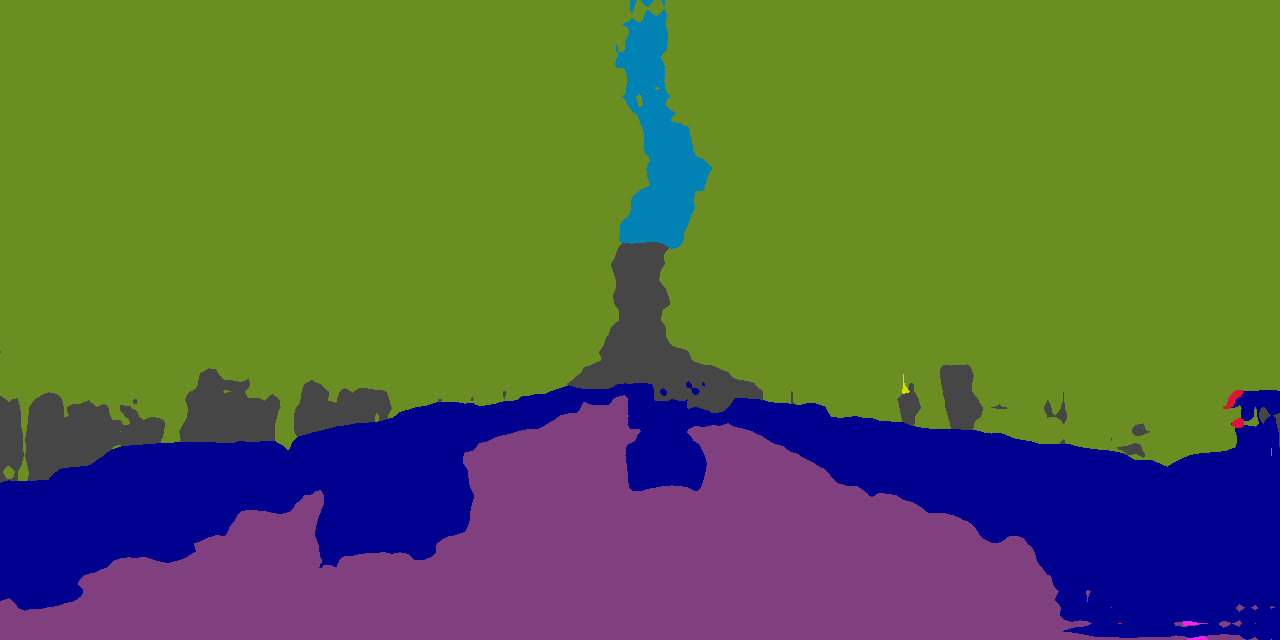}
\end{subfigure}%
\begin{subfigure}{\imgWidth}
\includegraphics[width=\textwidth]{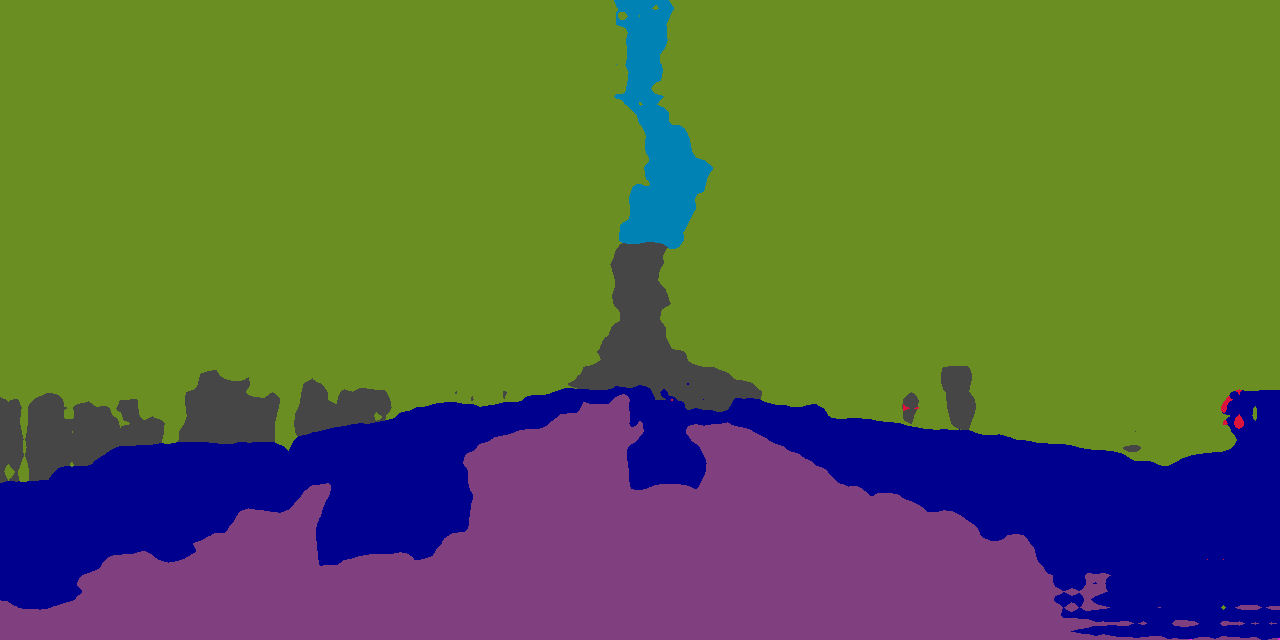}
\end{subfigure}
\hspace*{.2em}%
\begin{subfigure}{\imgWidth}
\includegraphics[width=\textwidth]{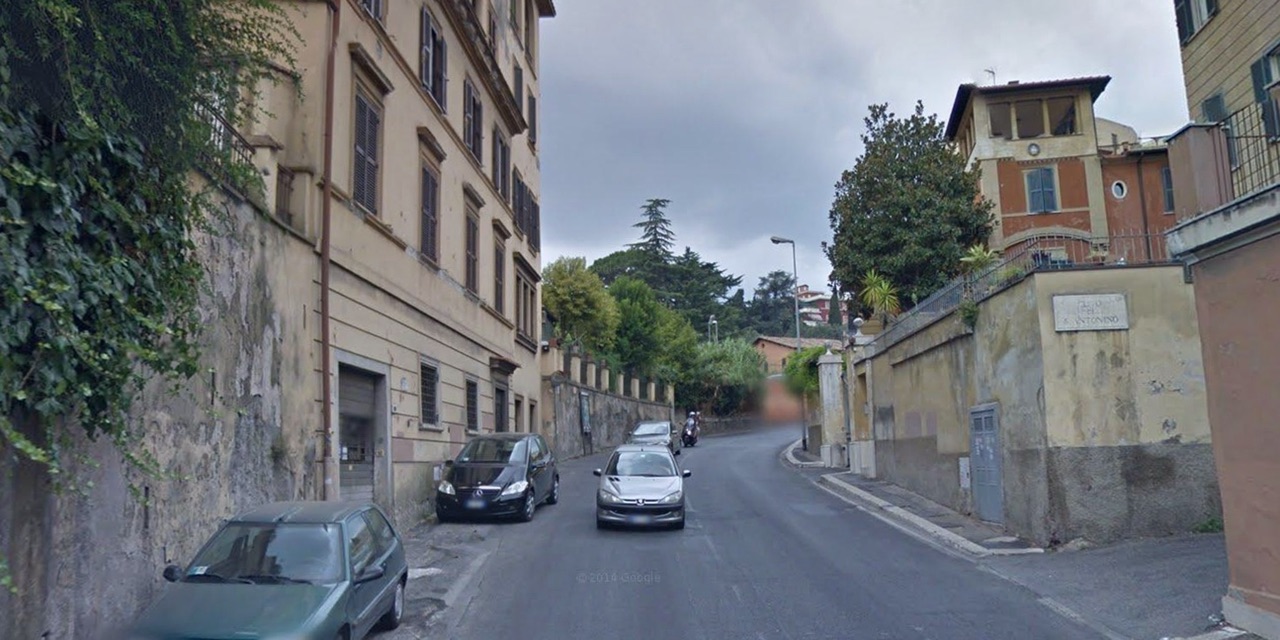}
\end{subfigure}%
\begin{subfigure}{\imgWidth}
\includegraphics[width=\textwidth]{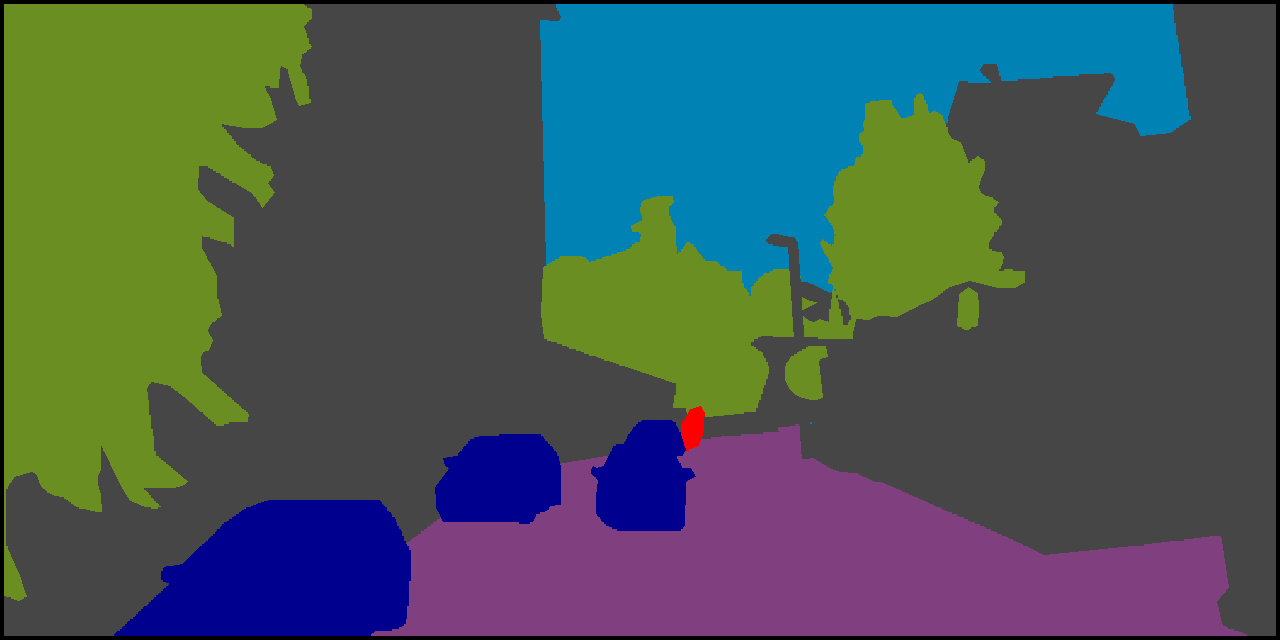}
\end{subfigure}%
\begin{subfigure}{\imgWidth}
\includegraphics[width=\textwidth]{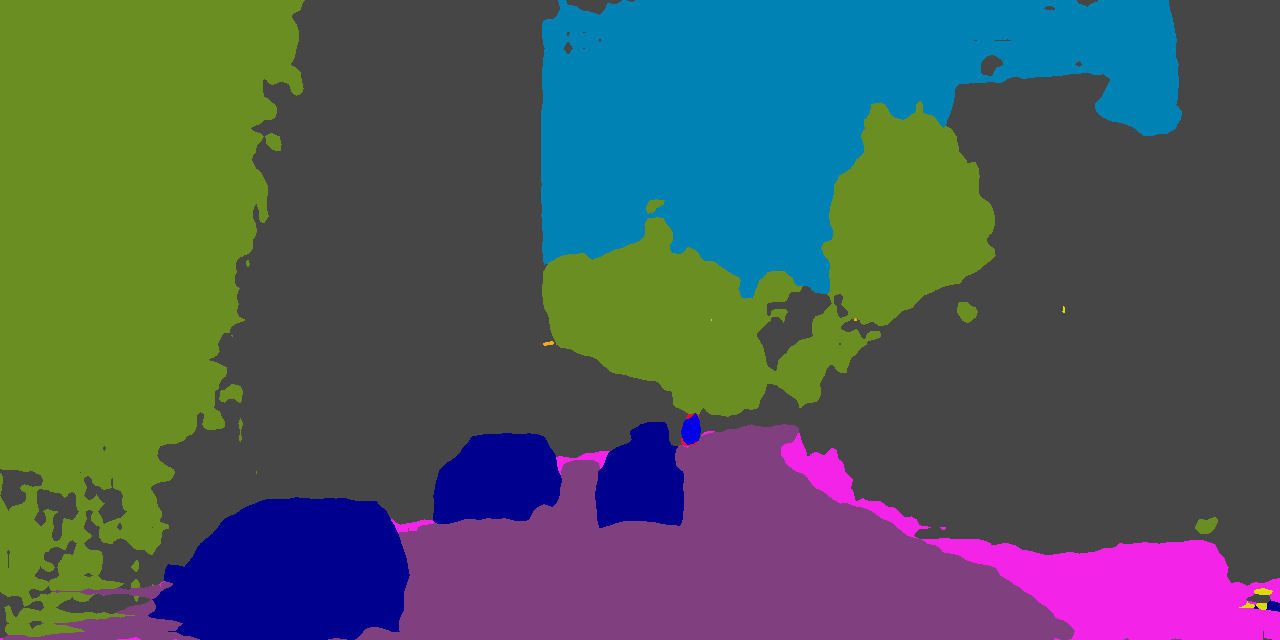}
\end{subfigure}%
\begin{subfigure}{\imgWidth}
\includegraphics[width=\textwidth]{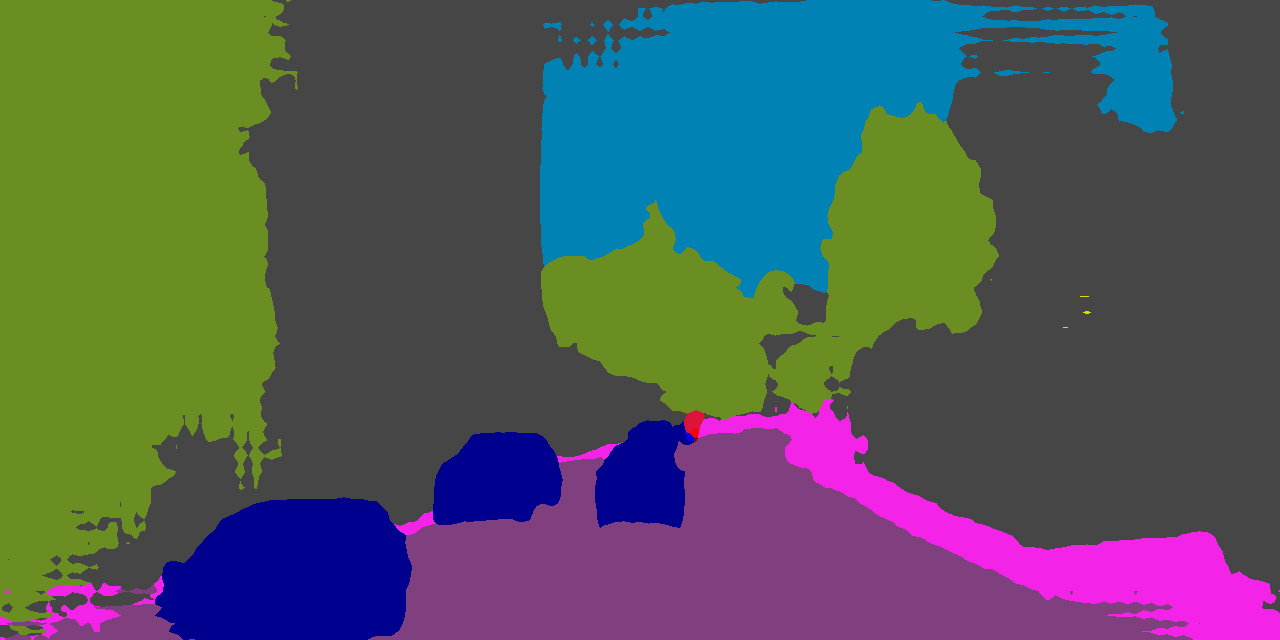}
\end{subfigure}%
\begin{subfigure}{\imgWidth}
\includegraphics[width=\textwidth]{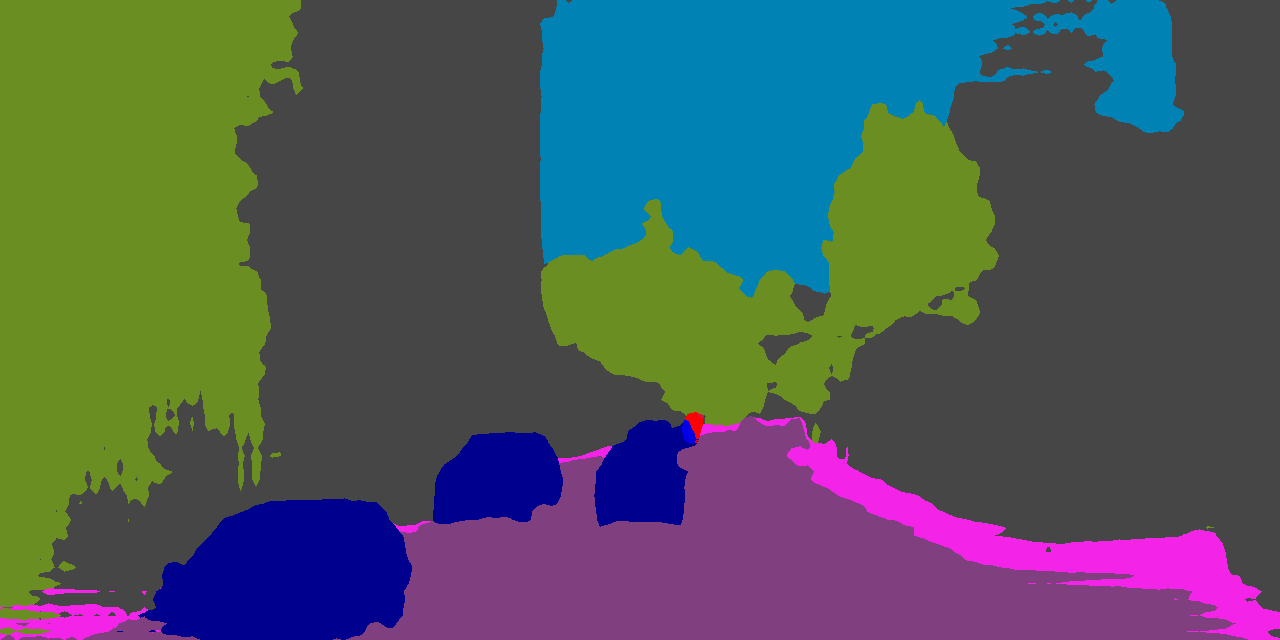}
\end{subfigure}%
\begin{subfigure}{\imgWidth}
\includegraphics[width=\textwidth]{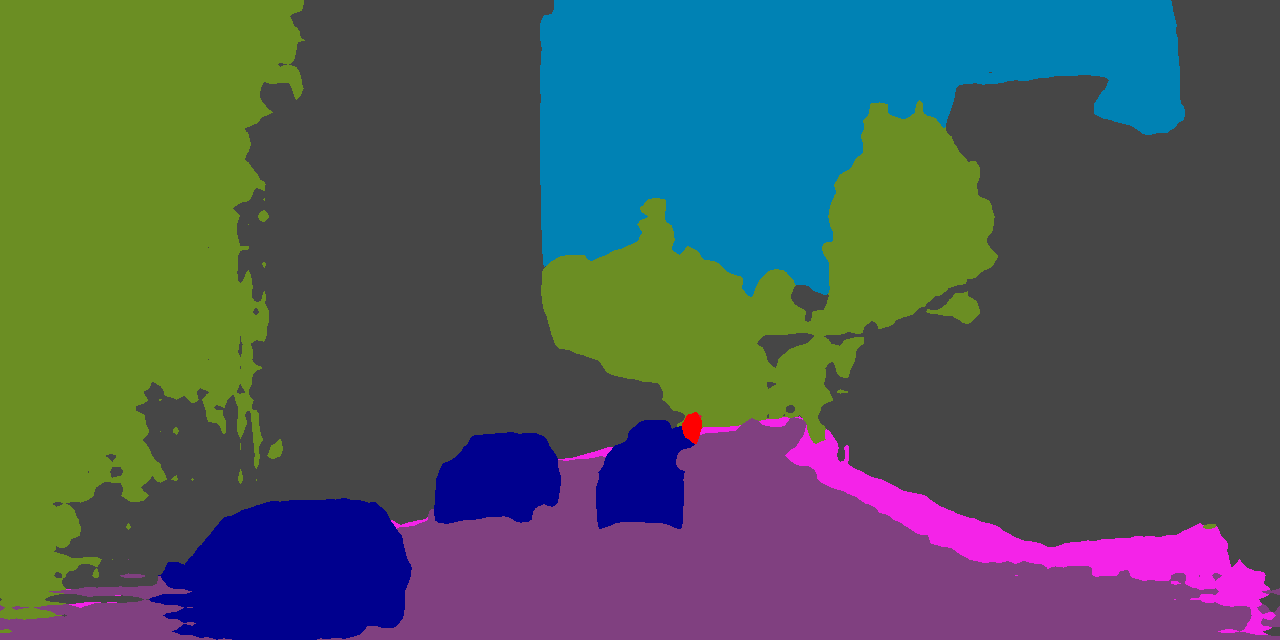}
\end{subfigure}
\end{subfigure}

\vspace{.2em}

\begin{subfigure}{.6em}
\scriptsize\rotatebox{90}{Rio}
\end{subfigure}%
\begin{subfigure}{\textwidth-1em}
\hspace*{.2em}%
\begin{subfigure}{\imgWidth}
\includegraphics[width=\textwidth]{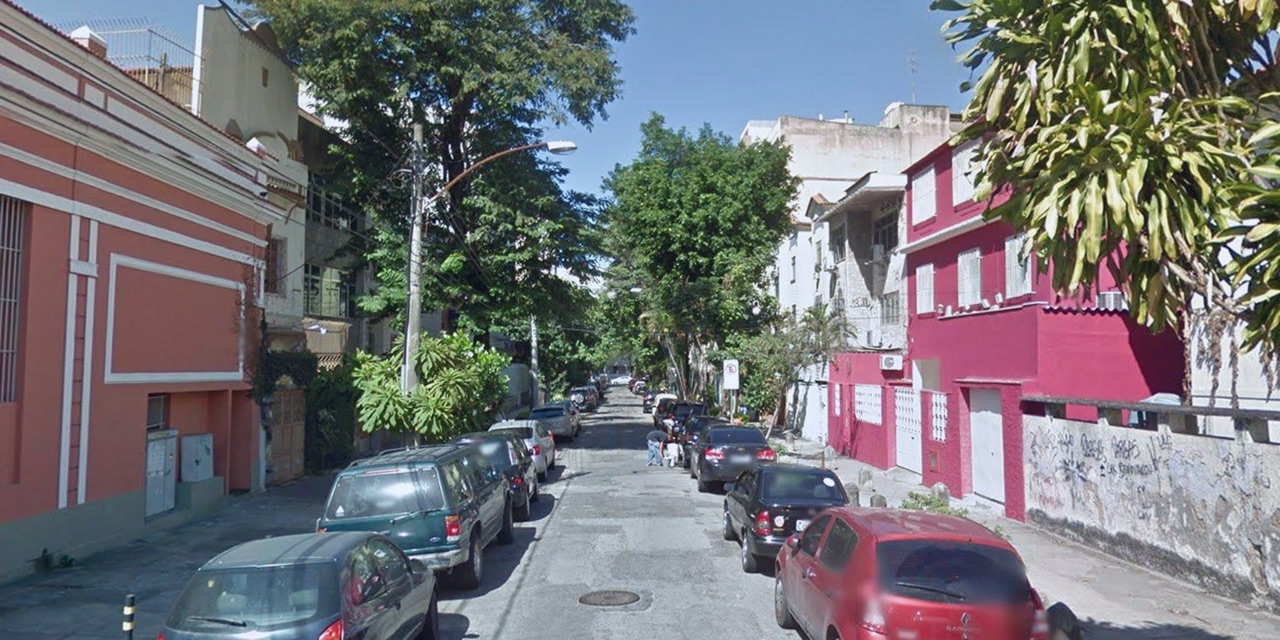}
\end{subfigure}%
\begin{subfigure}{\imgWidth}
\includegraphics[width=\textwidth]{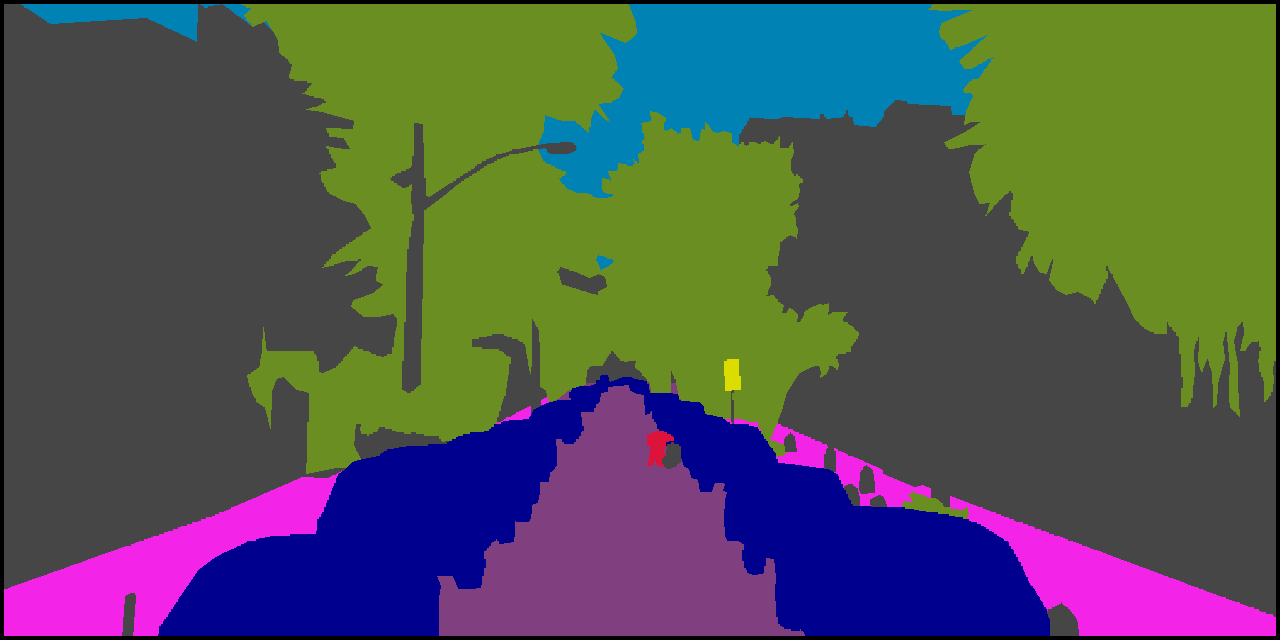}
\end{subfigure}%
\begin{subfigure}{\imgWidth}
\includegraphics[width=\textwidth]{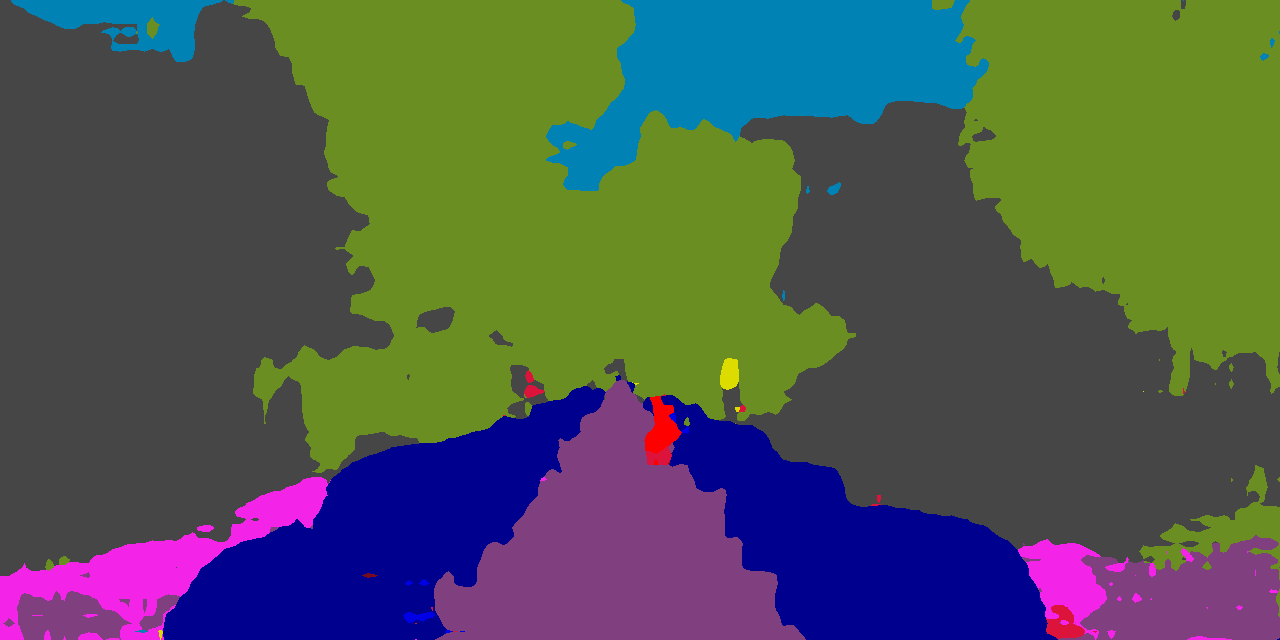}
\end{subfigure}%
\begin{subfigure}{\imgWidth}
\includegraphics[width=\textwidth]{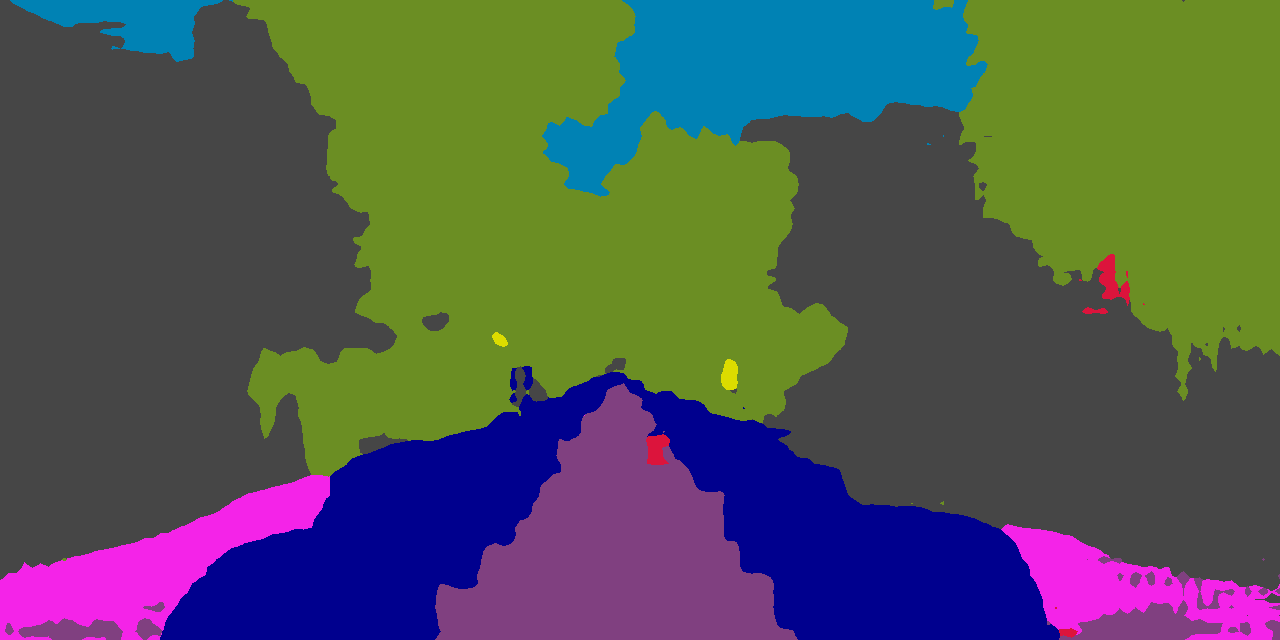}
\end{subfigure}%
\begin{subfigure}{\imgWidth}
\includegraphics[width=\textwidth]{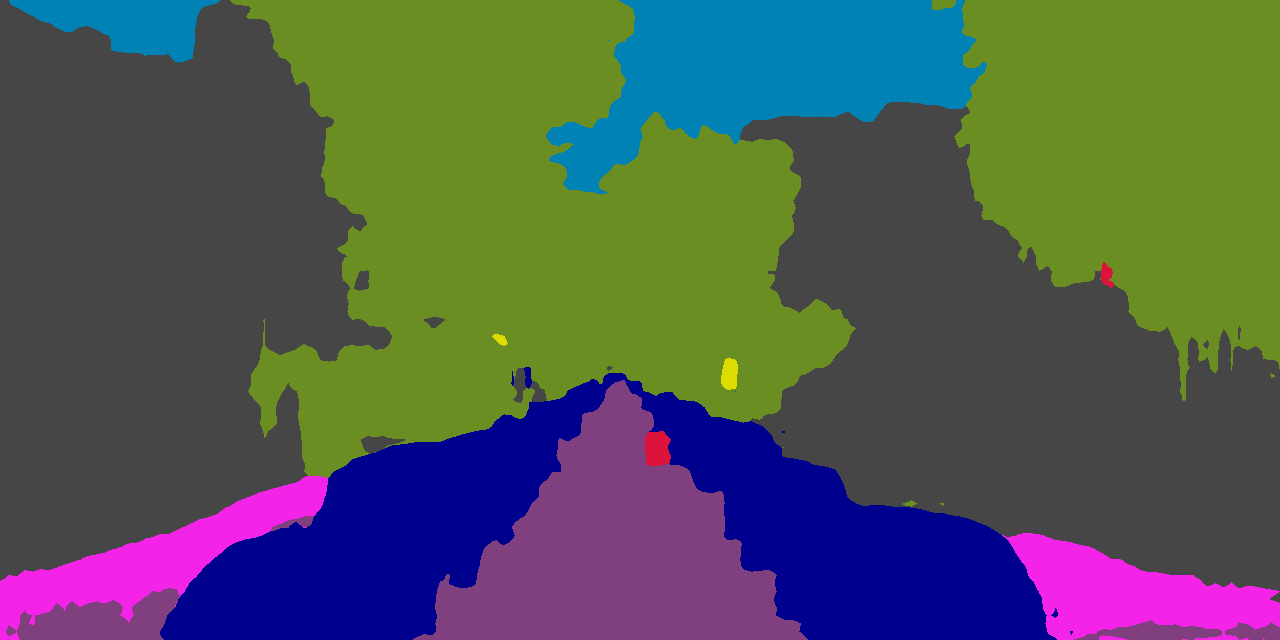}
\end{subfigure}%
\begin{subfigure}{\imgWidth}
\includegraphics[width=\textwidth]{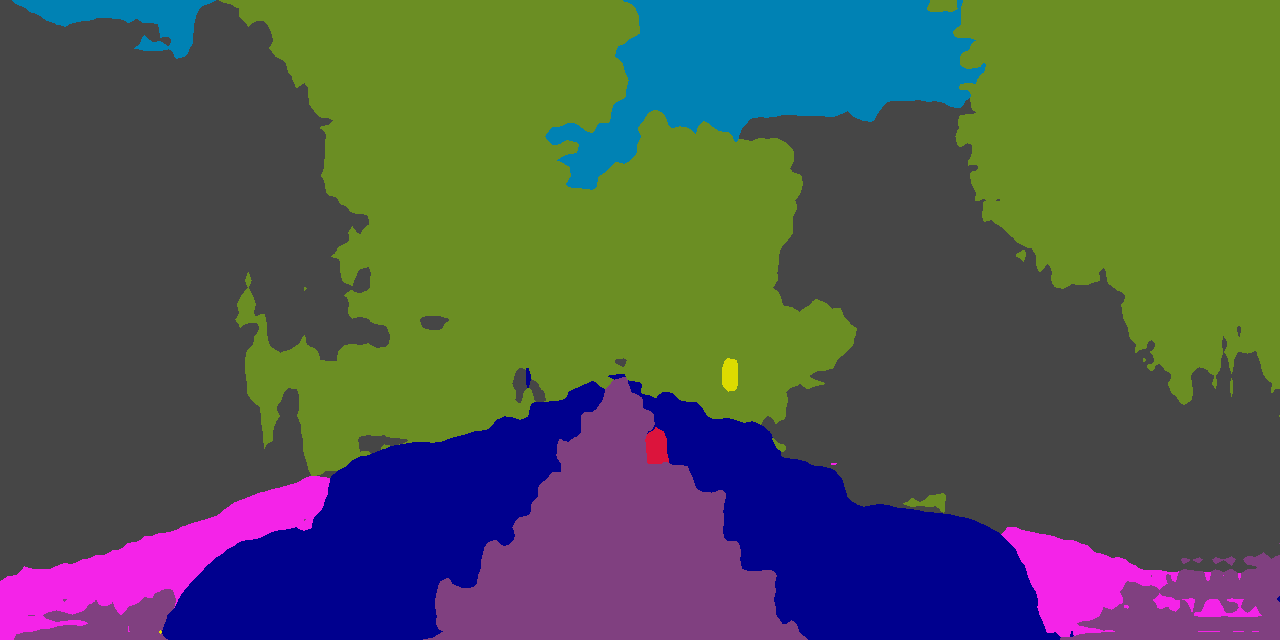}
\end{subfigure}
\hspace*{.2em}%
\begin{subfigure}{\imgWidth}
\includegraphics[width=\textwidth]{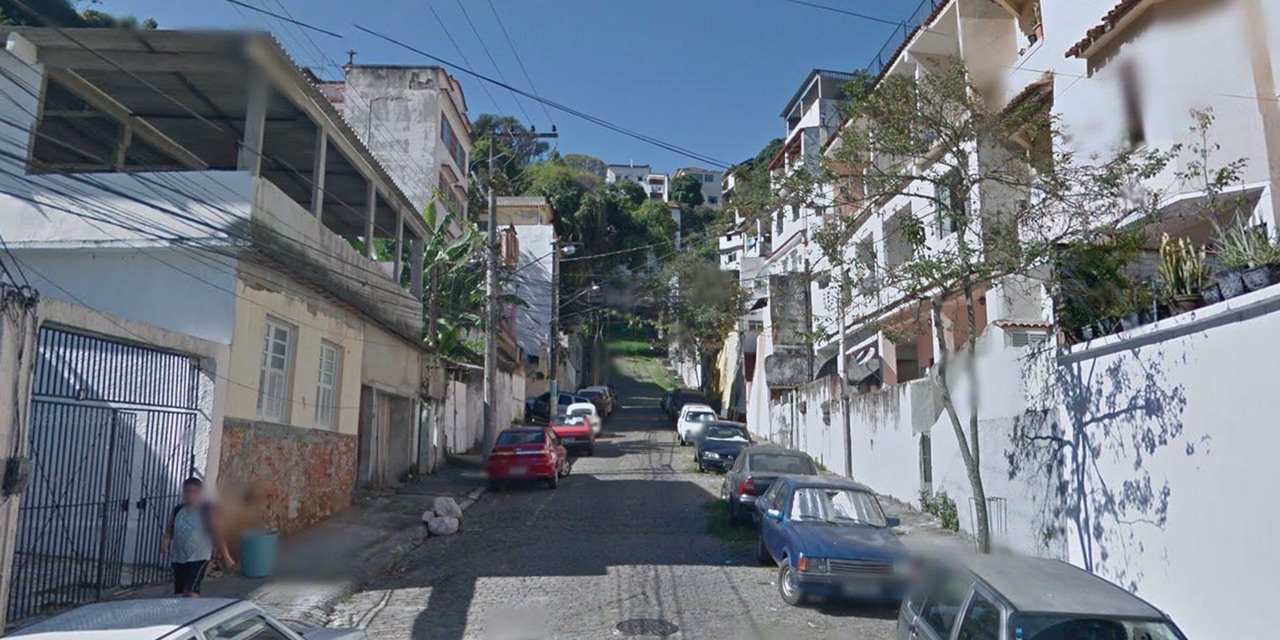}
\end{subfigure}%
\begin{subfigure}{\imgWidth}
\includegraphics[width=\textwidth]{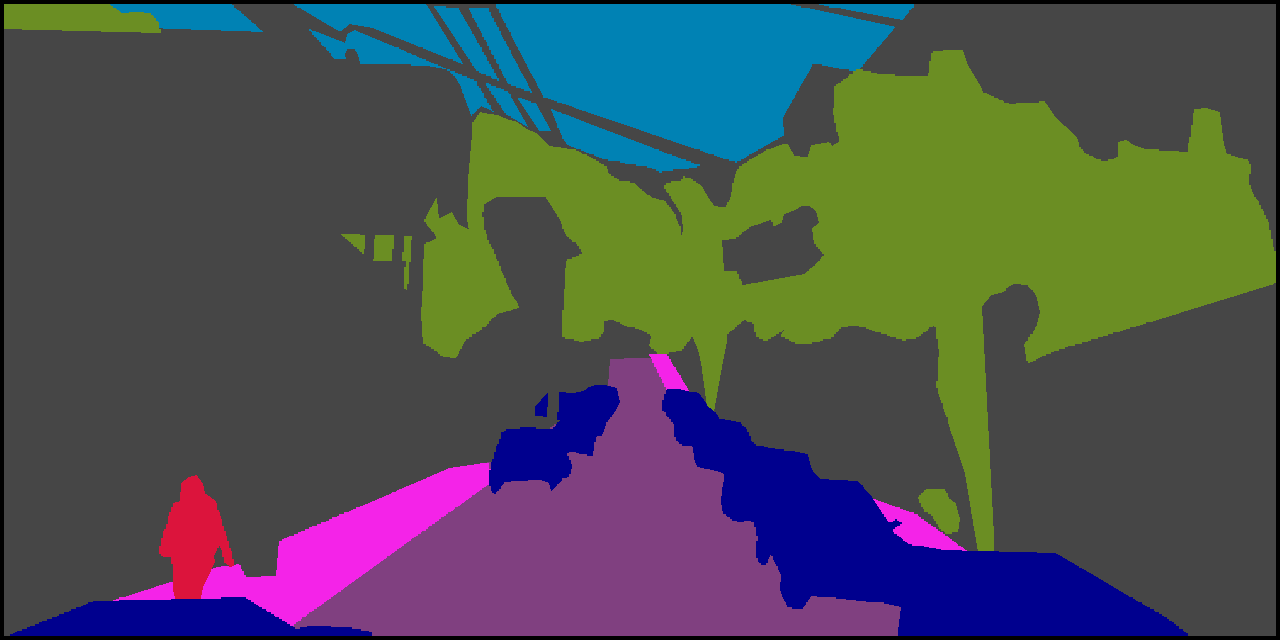}
\end{subfigure}%
\begin{subfigure}{\imgWidth}
\includegraphics[width=\textwidth]{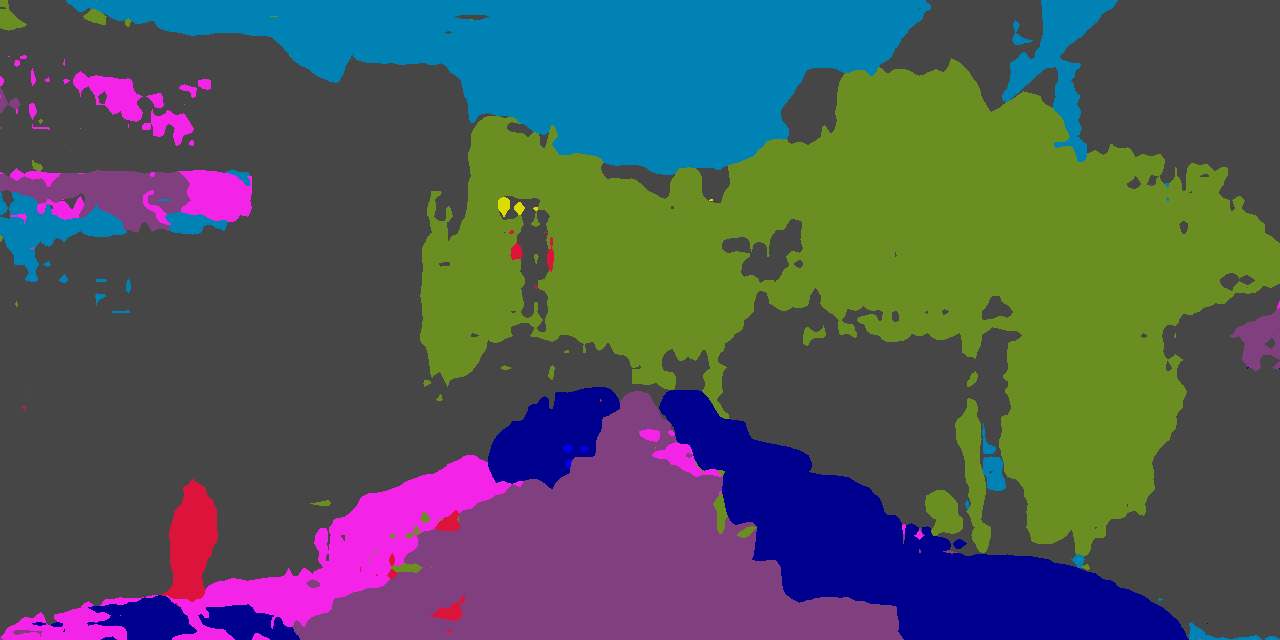}
\end{subfigure}%
\begin{subfigure}{\imgWidth}
\includegraphics[width=\textwidth]{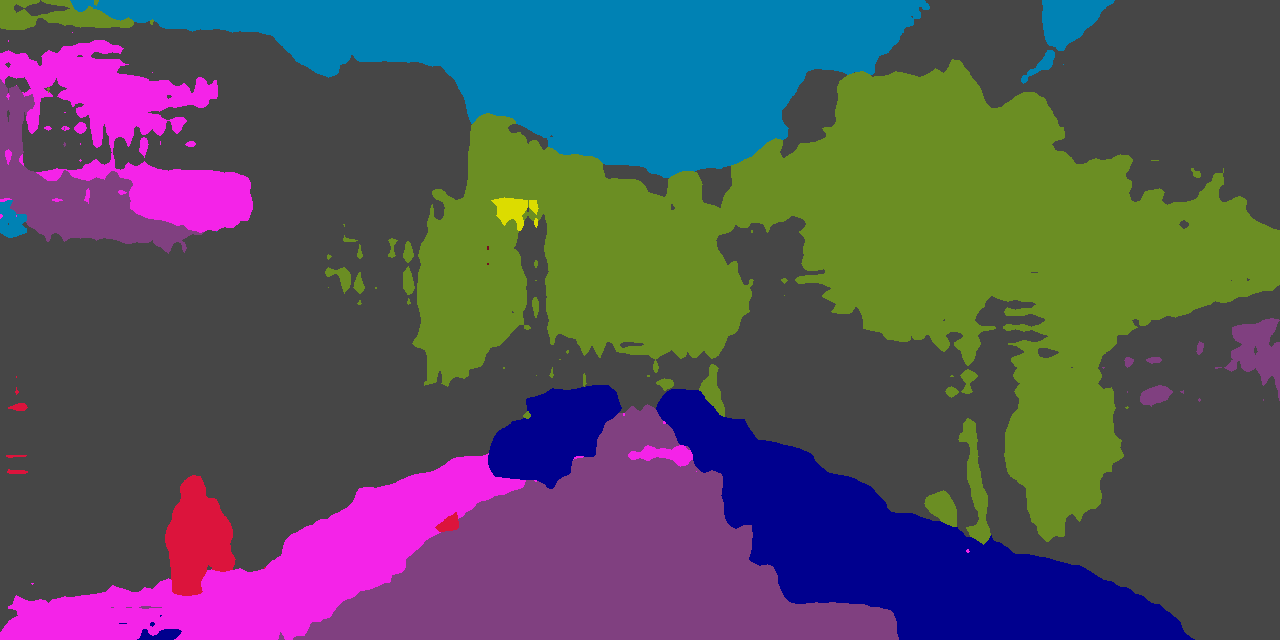}
\end{subfigure}%
\begin{subfigure}{\imgWidth}
\includegraphics[width=\textwidth]{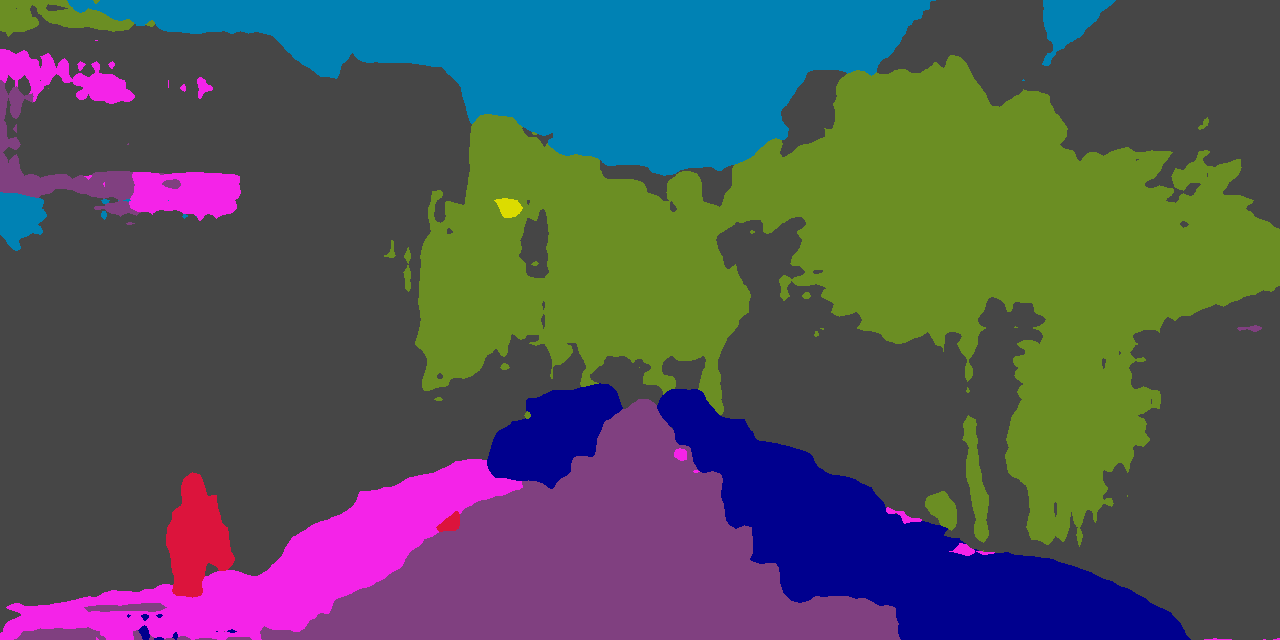}
\end{subfigure}%
\begin{subfigure}{\imgWidth}
\includegraphics[width=\textwidth]{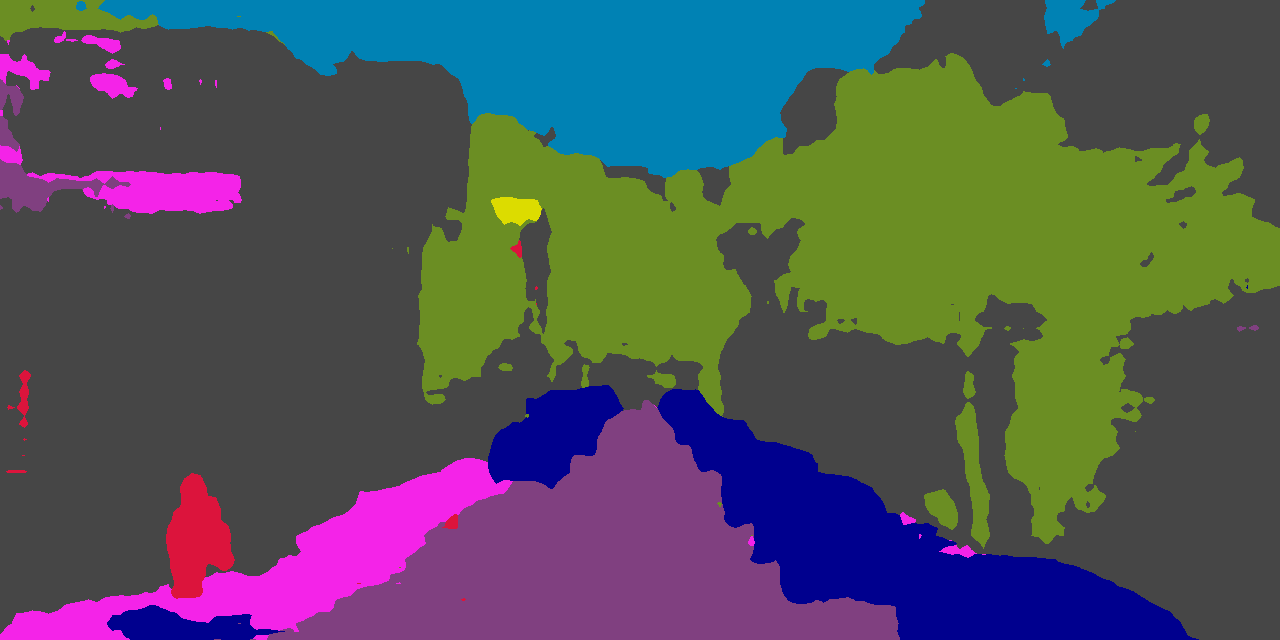}
\end{subfigure}
\end{subfigure}

\vspace{.2em}

\begin{subfigure}{.6em}
\scriptsize\rotatebox{90}{Tokyo}
\end{subfigure}%
\begin{subfigure}{\textwidth-1em}
\hspace*{.2em}%
\begin{subfigure}{\imgWidth}
\includegraphics[width=\textwidth]{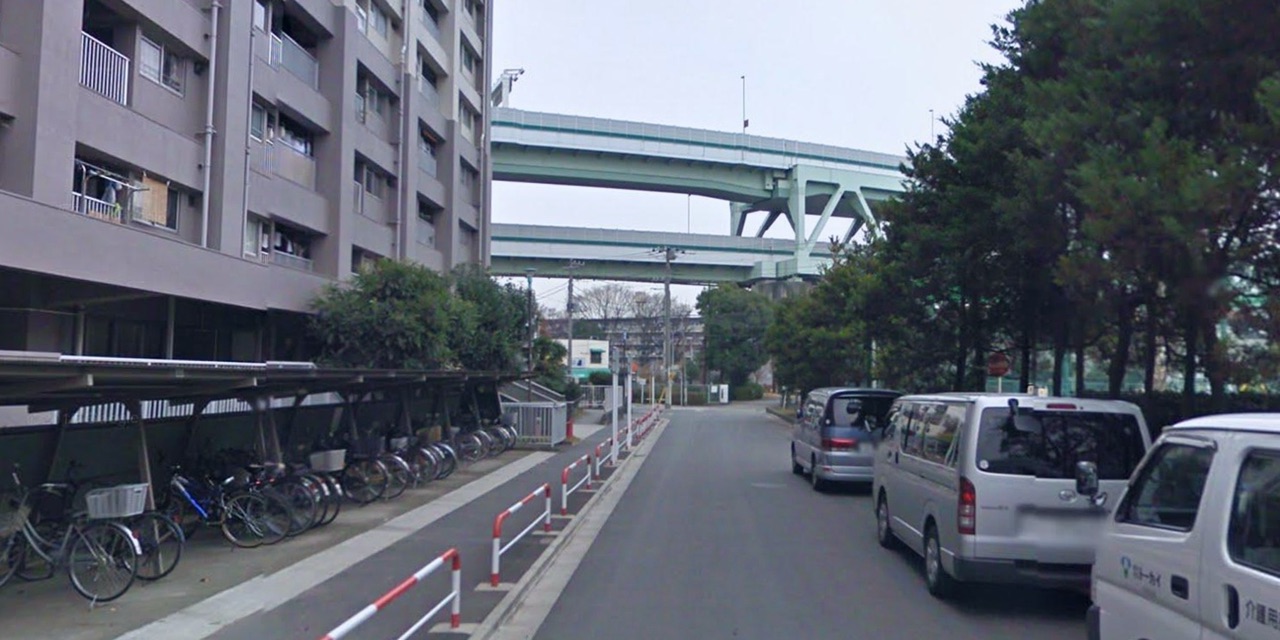}
\end{subfigure}%
\begin{subfigure}{\imgWidth}
\includegraphics[width=\textwidth]{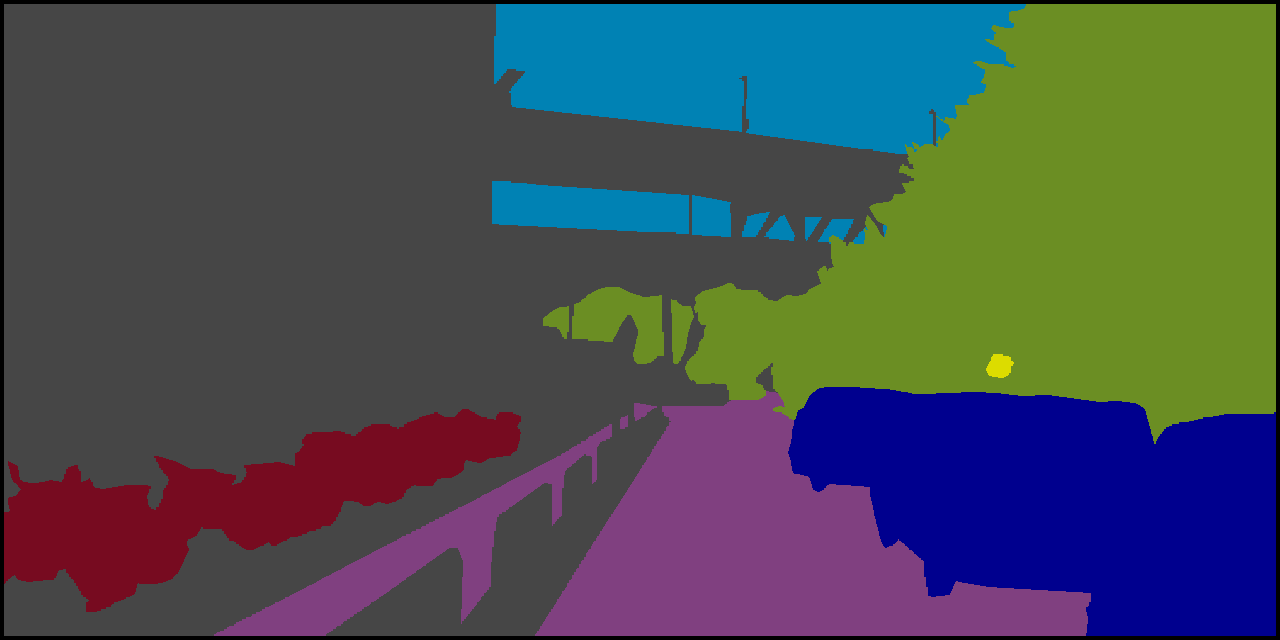}
\end{subfigure}%
\begin{subfigure}{\imgWidth}
\includegraphics[width=\textwidth]{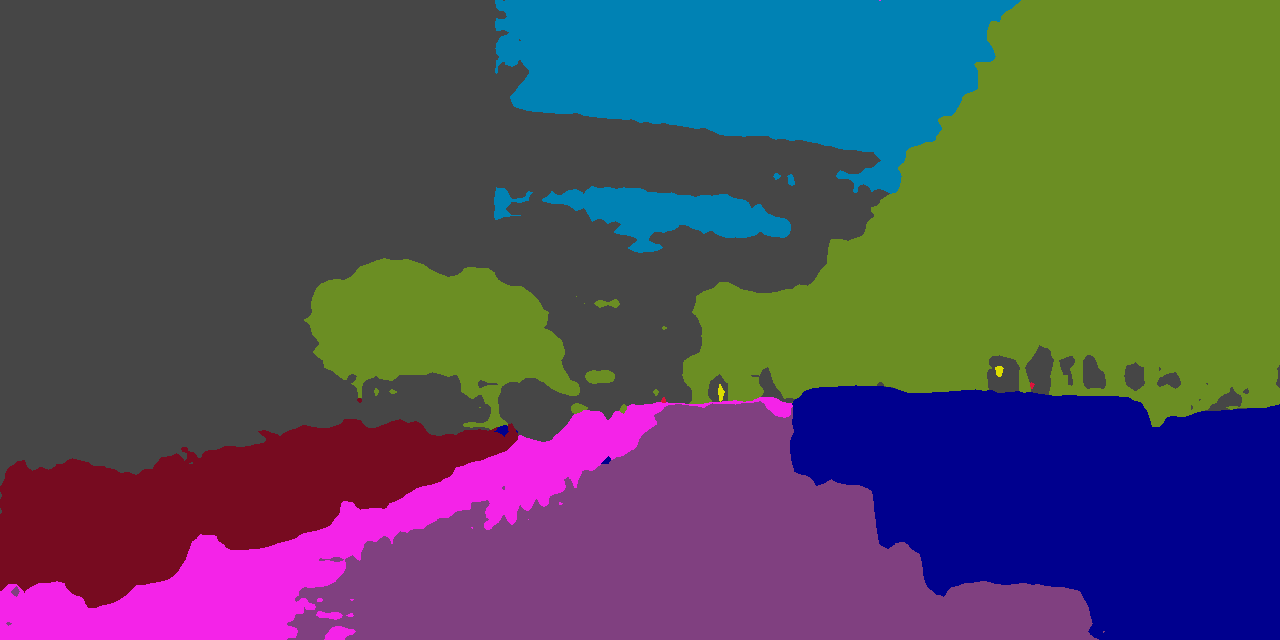}
\end{subfigure}%
\begin{subfigure}{\imgWidth}
\includegraphics[width=\textwidth]{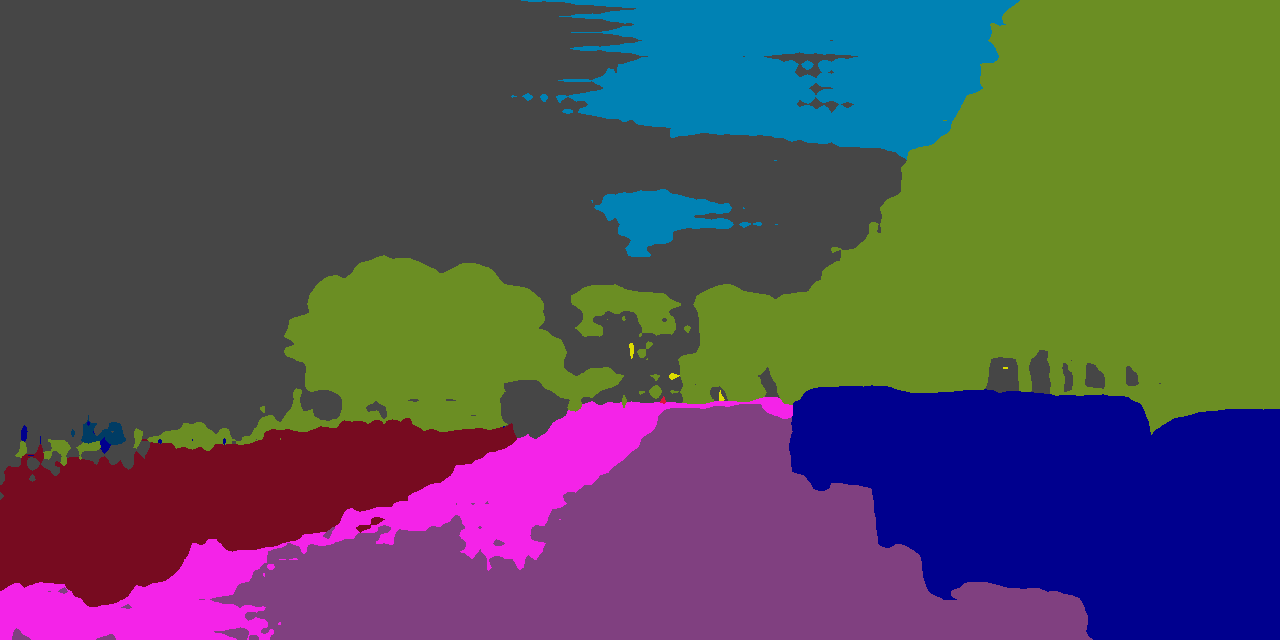}
\end{subfigure}%
\begin{subfigure}{\imgWidth}
\includegraphics[width=\textwidth]{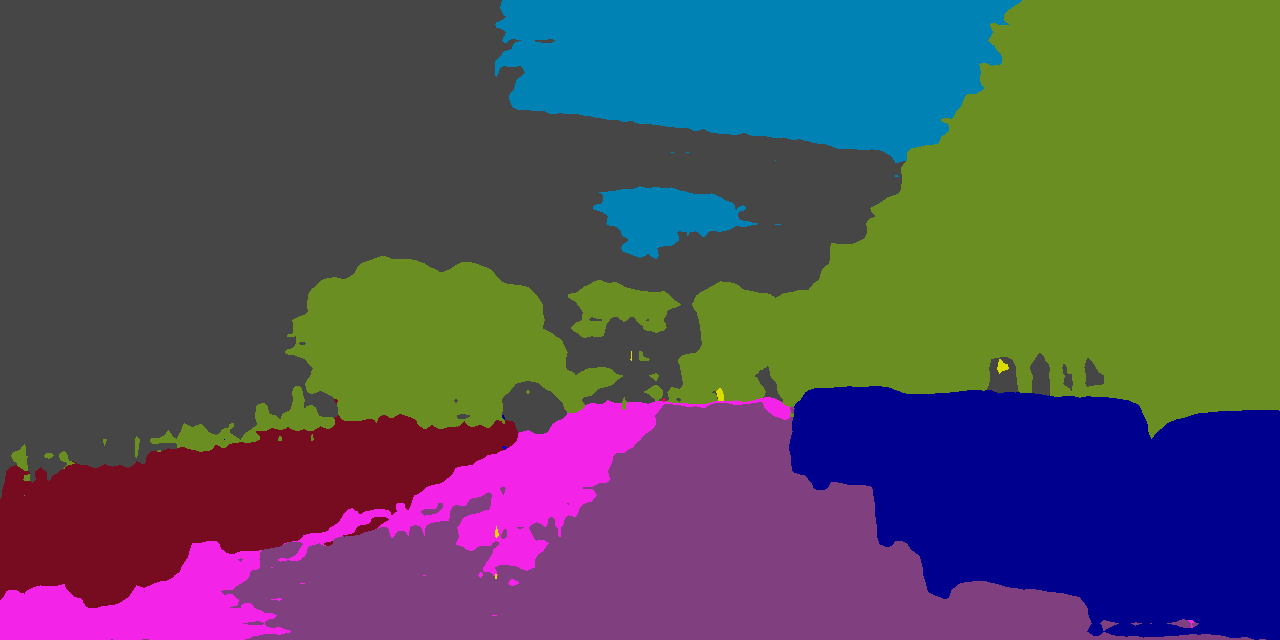}
\end{subfigure}%
\begin{subfigure}{\imgWidth}
\includegraphics[width=\textwidth]{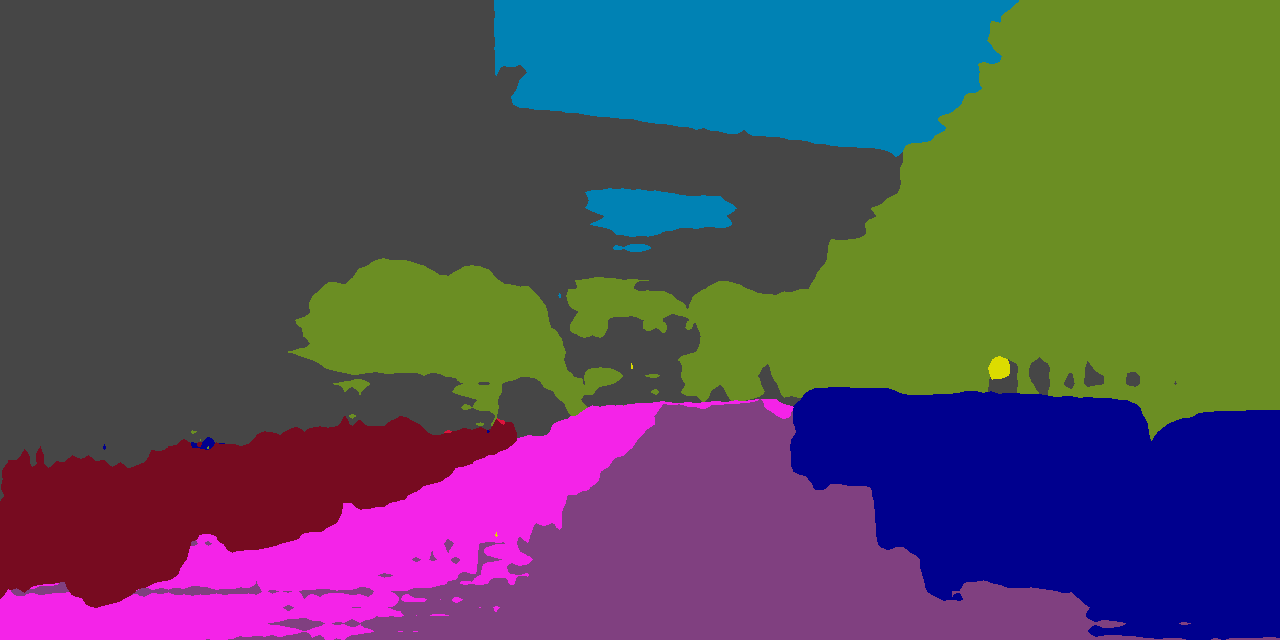}
\end{subfigure}
\hspace*{.2em}%
\begin{subfigure}{\imgWidth}
\includegraphics[width=\textwidth]{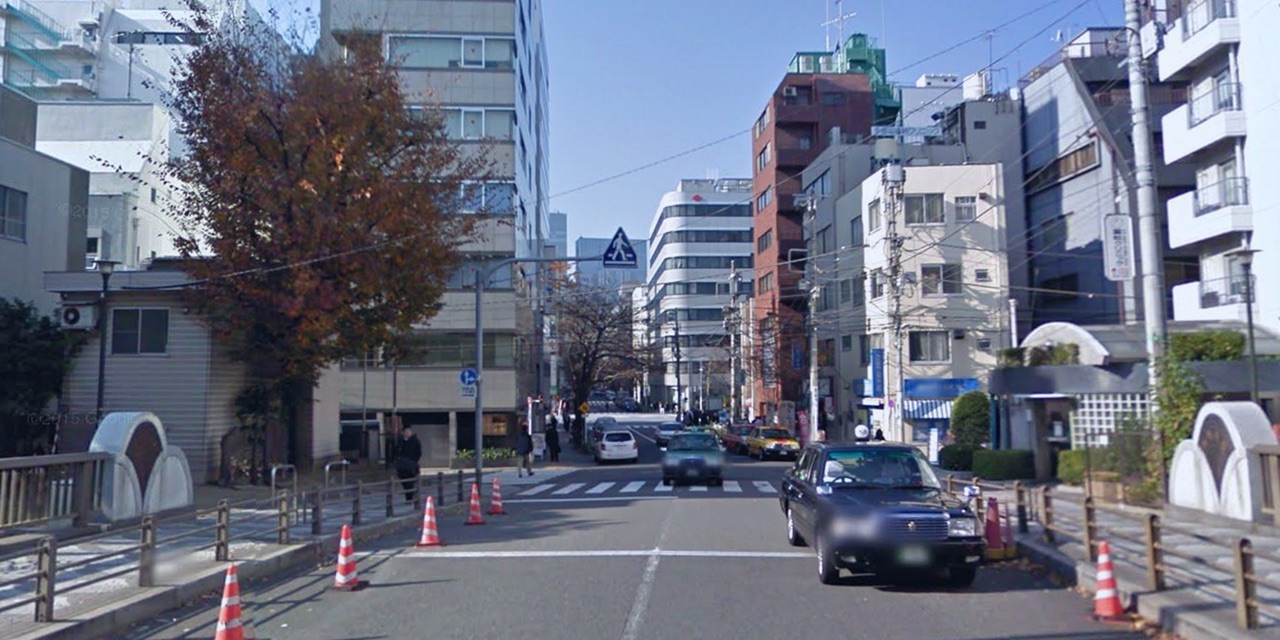}
\end{subfigure}%
\begin{subfigure}{\imgWidth}
\includegraphics[width=\textwidth]{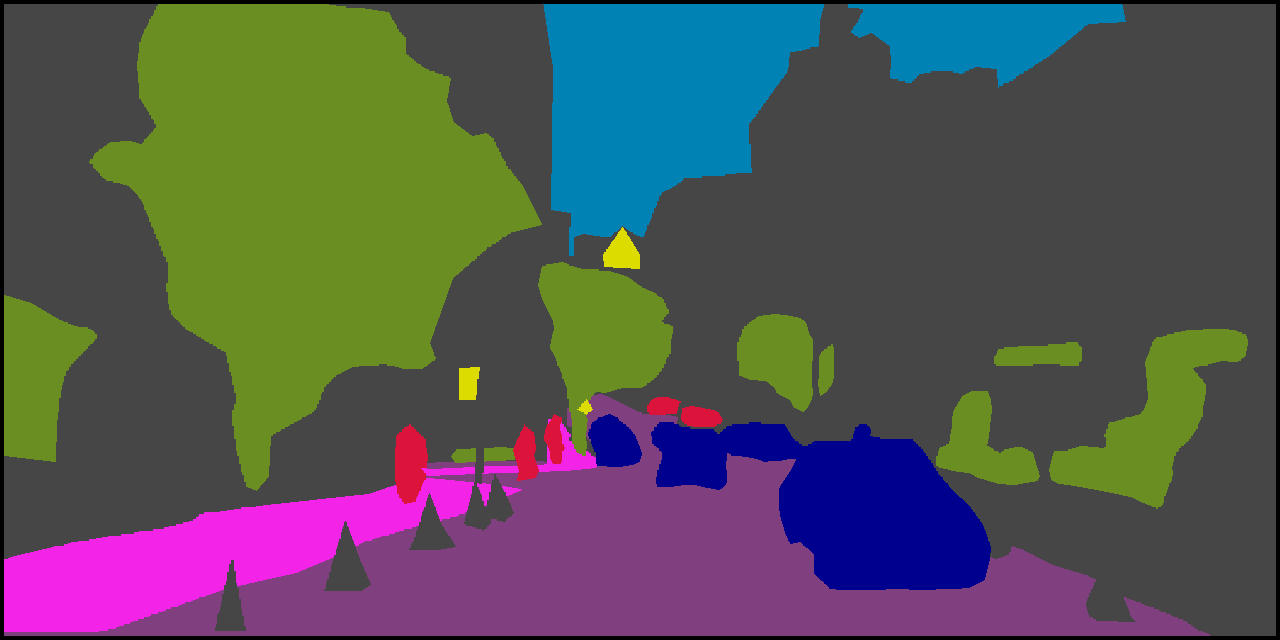}
\end{subfigure}%
\begin{subfigure}{\imgWidth}
\includegraphics[width=\textwidth]{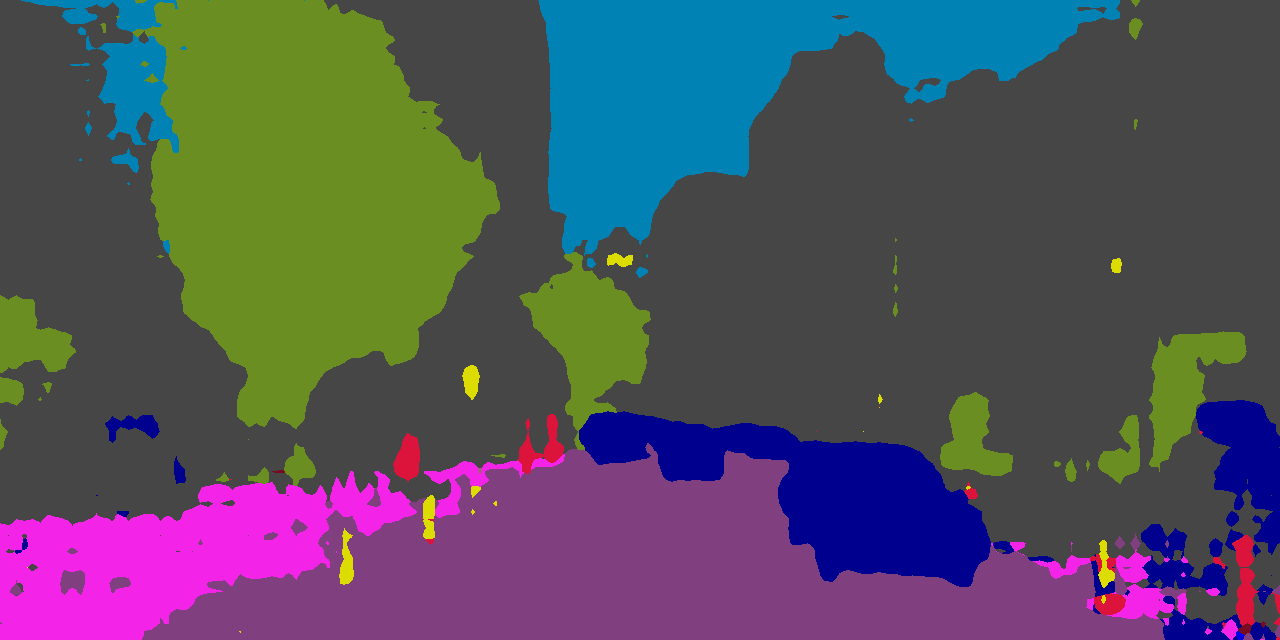}
\end{subfigure}%
\begin{subfigure}{\imgWidth}
\includegraphics[width=\textwidth]{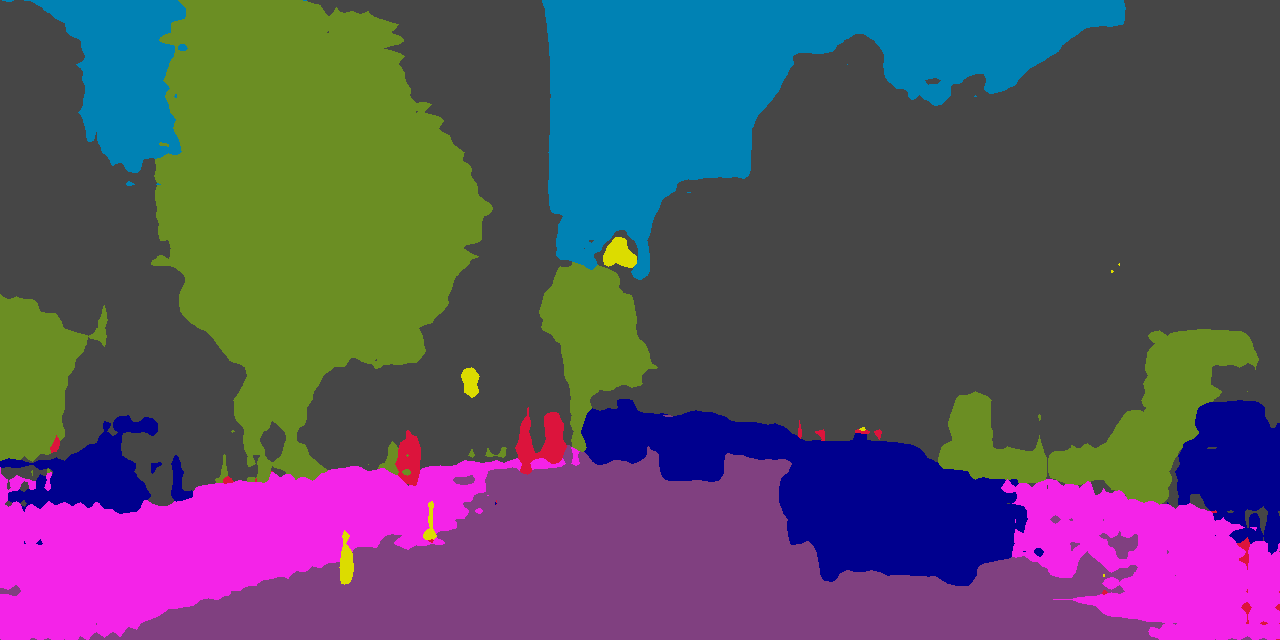}
\end{subfigure}%
\begin{subfigure}{\imgWidth}
\includegraphics[width=\textwidth]{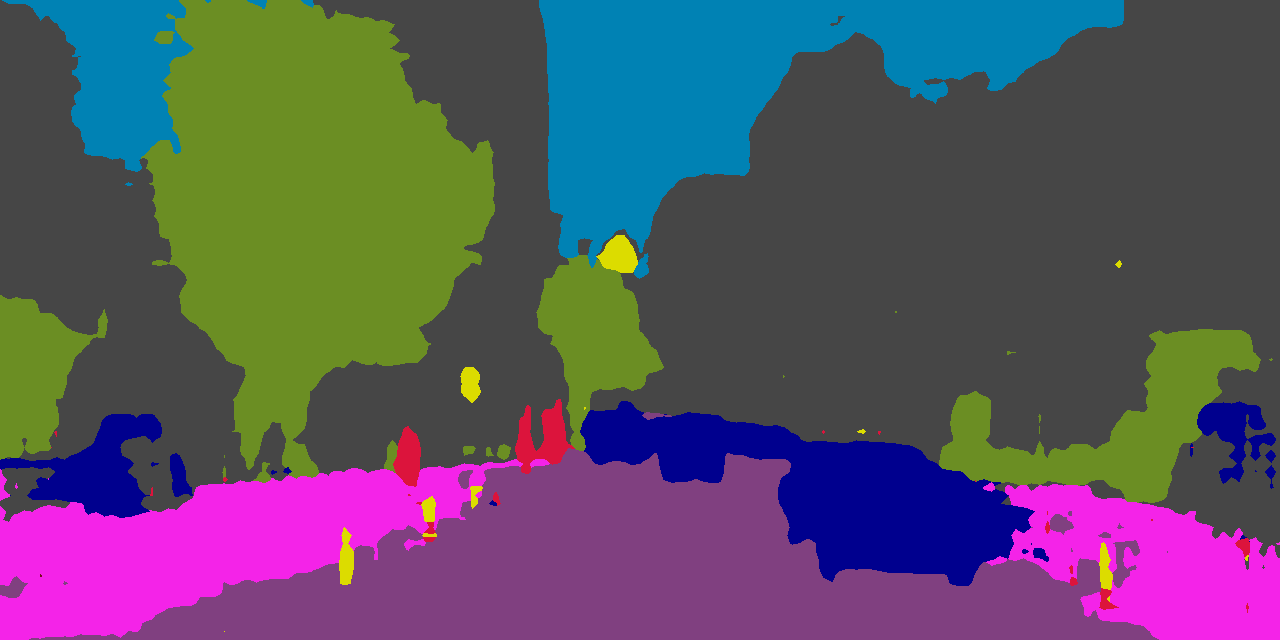}
\end{subfigure}%
\begin{subfigure}{\imgWidth}
\includegraphics[width=\textwidth]{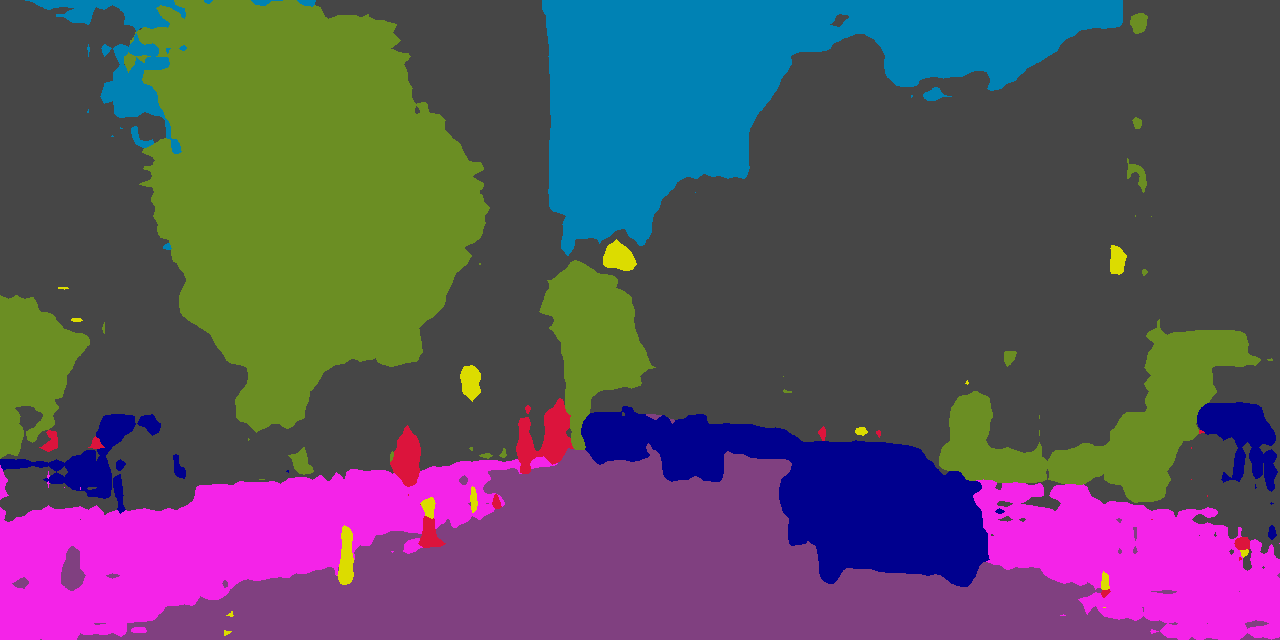}
\end{subfigure}
\end{subfigure}

\vspace{.2em}

\begin{subfigure}{.6em}
\scriptsize\rotatebox{90}{Taipei}
\end{subfigure}%
\begin{subfigure}{\textwidth-1em}
\hspace*{.2em}%
\begin{subfigure}{\imgWidth}
\includegraphics[width=\textwidth]{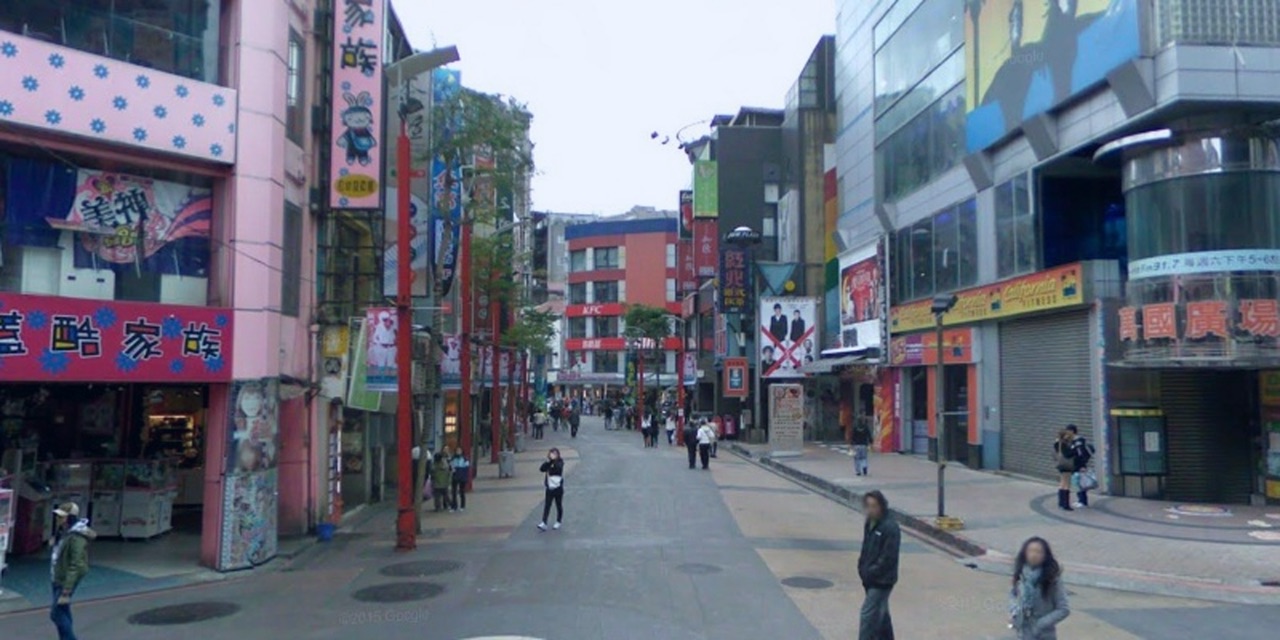}
\end{subfigure}%
\begin{subfigure}{\imgWidth}
\includegraphics[width=\textwidth]{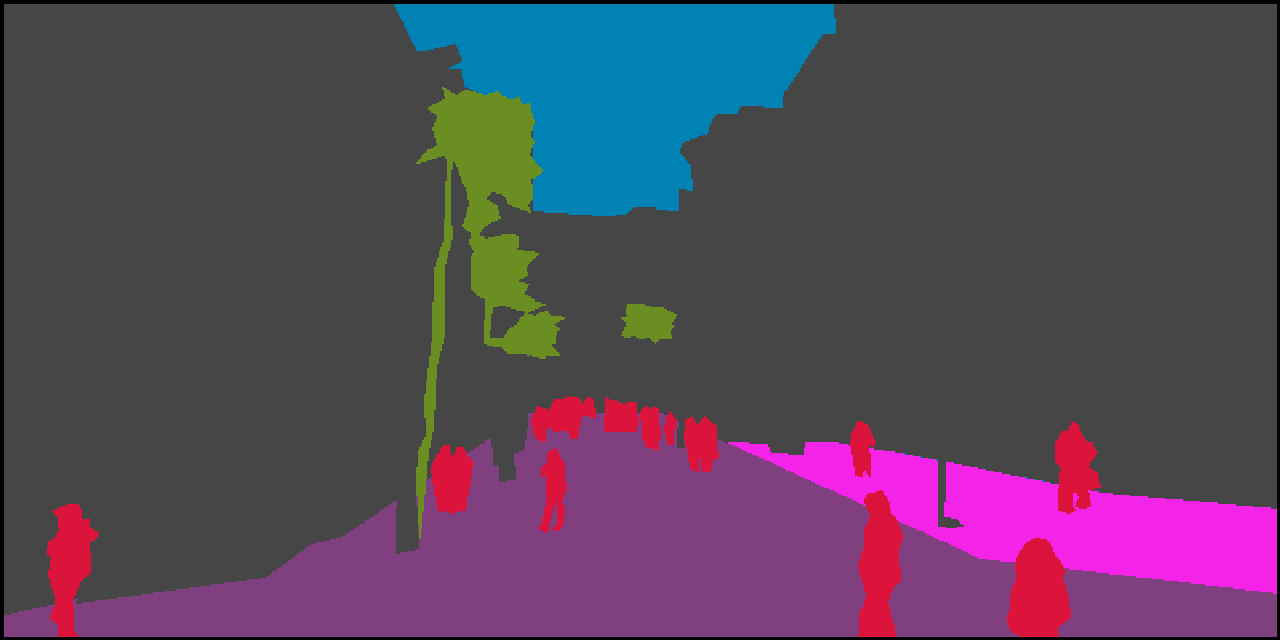}
\end{subfigure}%
\begin{subfigure}{\imgWidth}
\includegraphics[width=\textwidth]{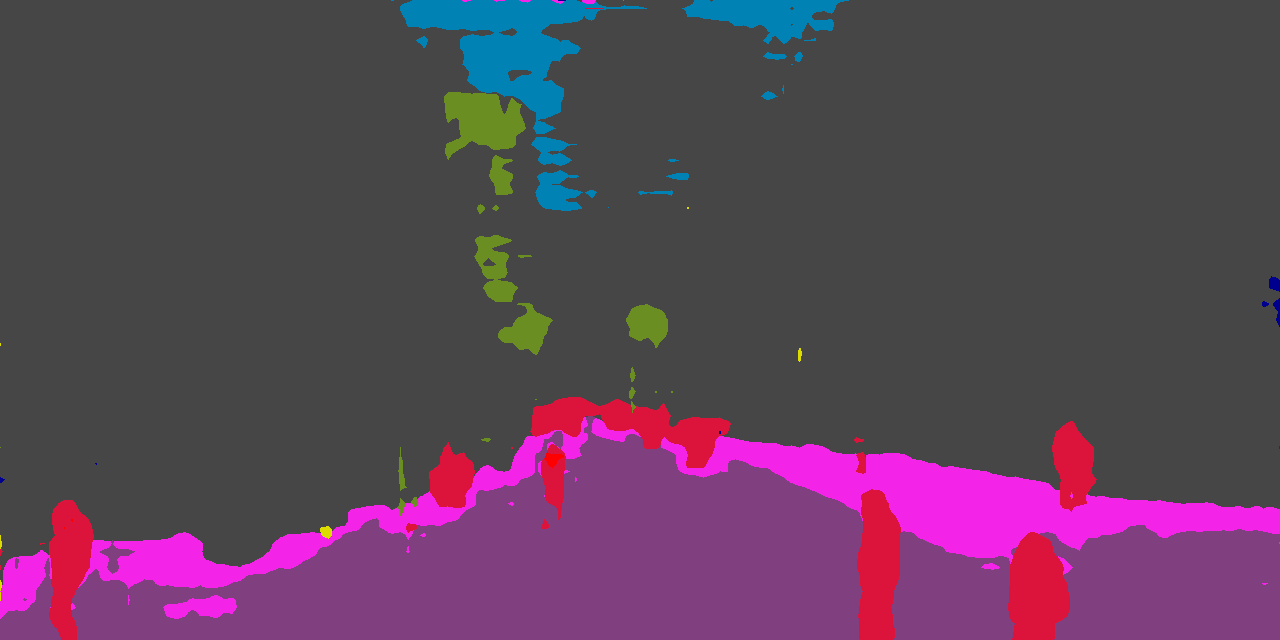}
\end{subfigure}%
\begin{subfigure}{\imgWidth}
\includegraphics[width=\textwidth]{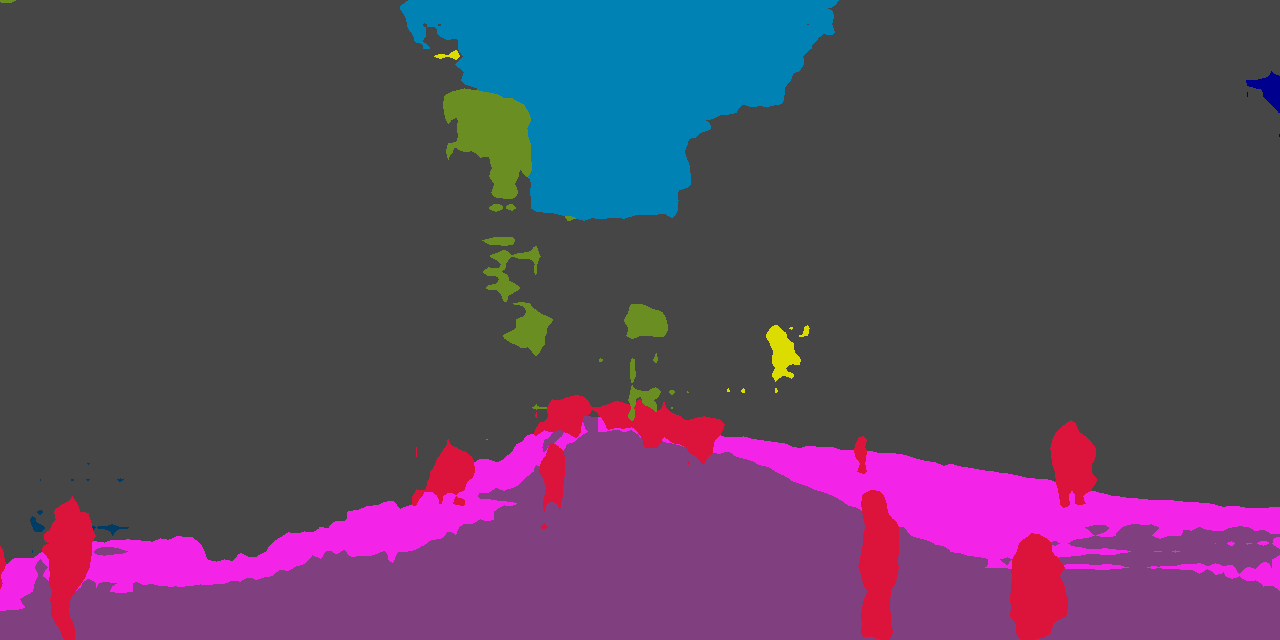}
\end{subfigure}%
\begin{subfigure}{\imgWidth}
\includegraphics[width=\textwidth]{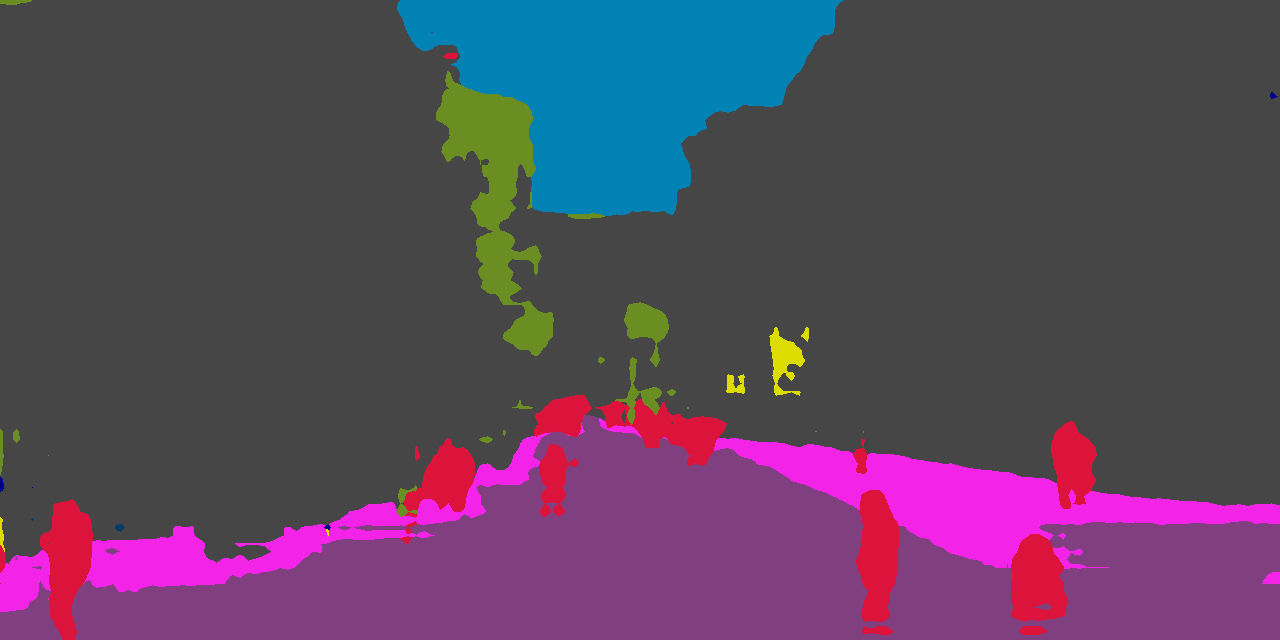}
\end{subfigure}%
\begin{subfigure}{\imgWidth}
\includegraphics[width=\textwidth]{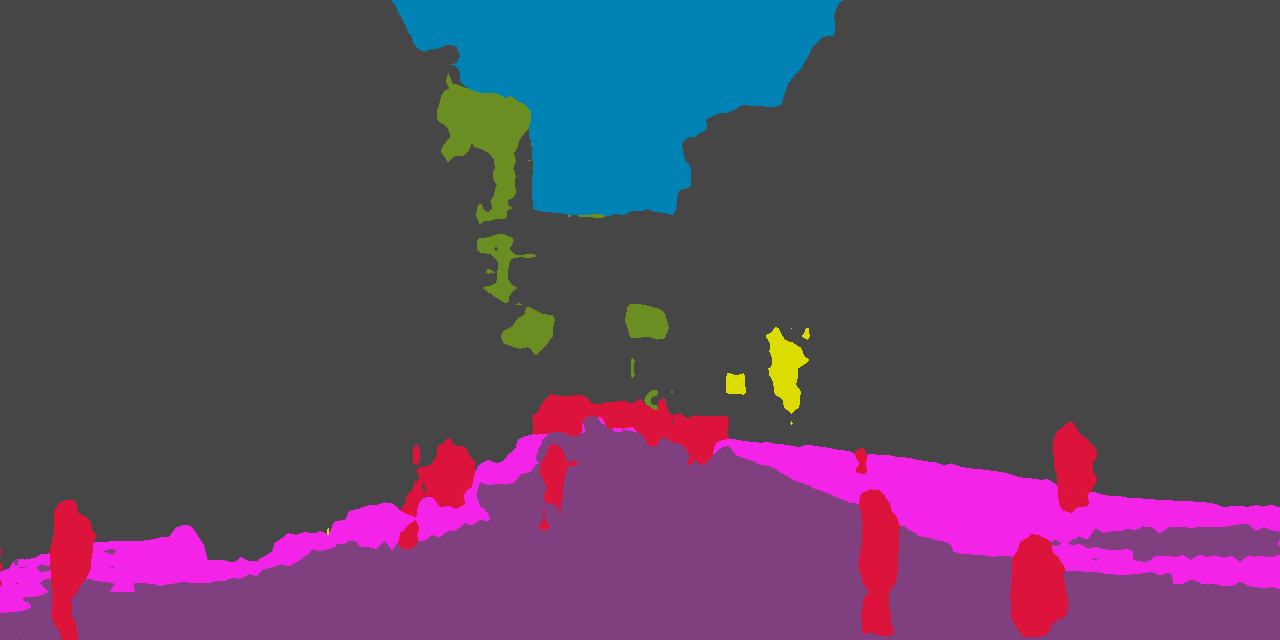}
\end{subfigure}
\hspace*{.2em}%
\begin{subfigure}{\imgWidth}
\includegraphics[width=\textwidth]{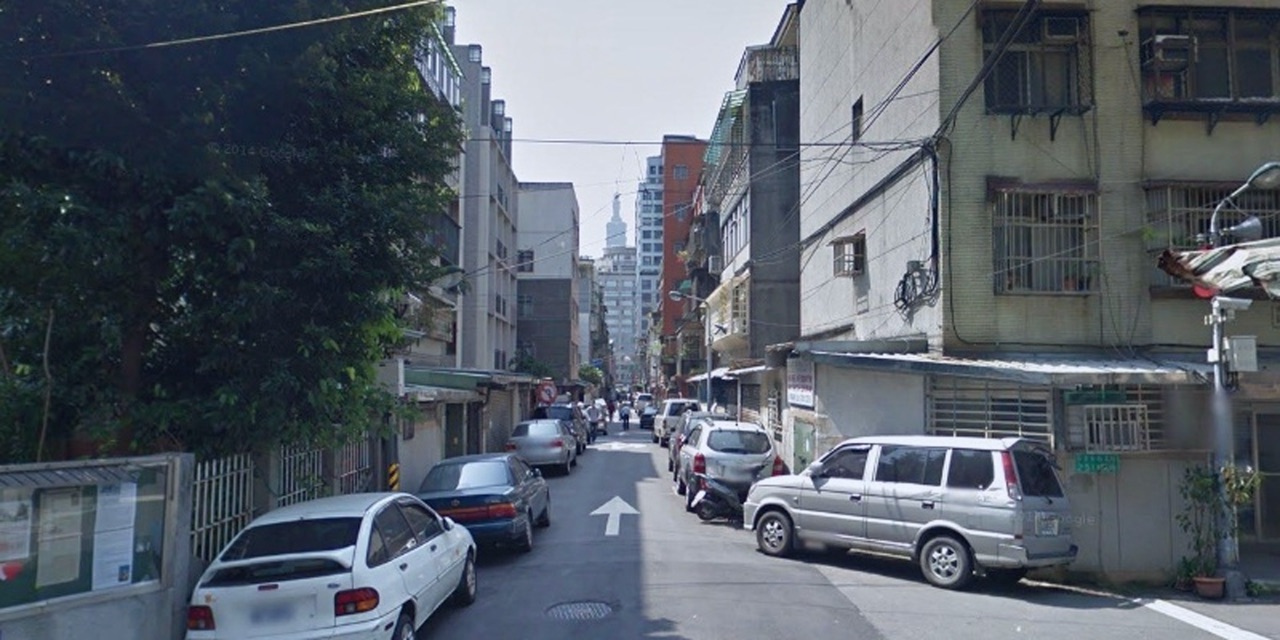}
\caption*{RGB}
\end{subfigure}%
\begin{subfigure}{\imgWidth}
\includegraphics[width=\textwidth]{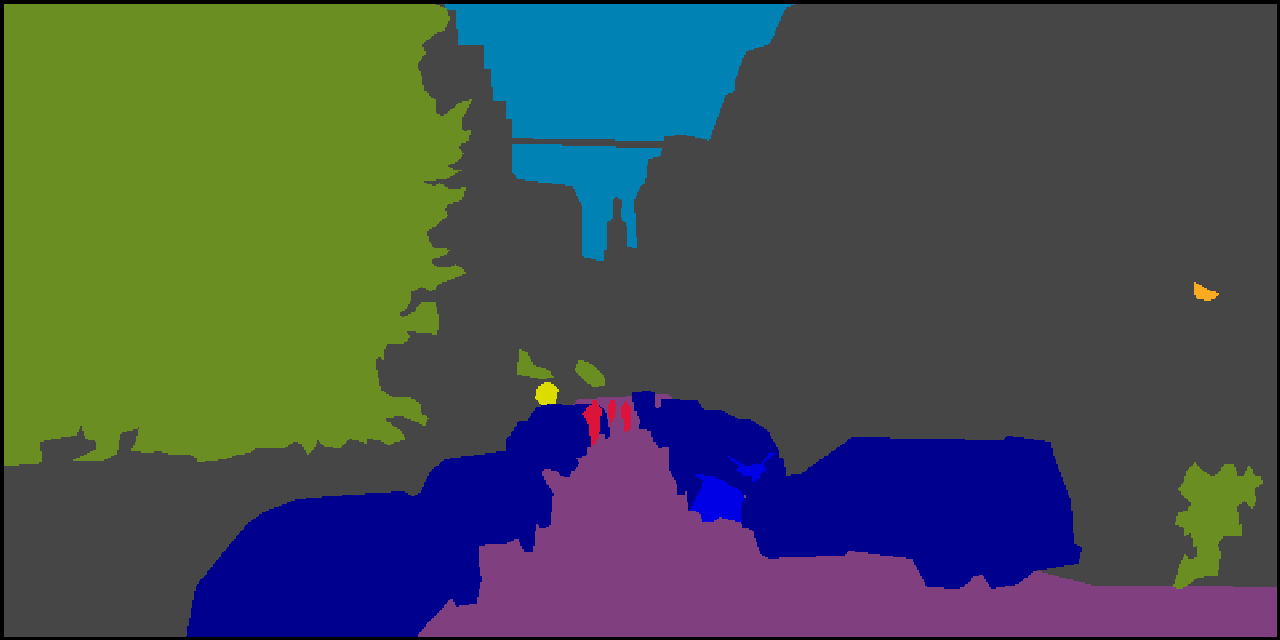}
\caption*{Ground truth}
\end{subfigure}%
\begin{subfigure}{\imgWidth}
\includegraphics[width=\textwidth]{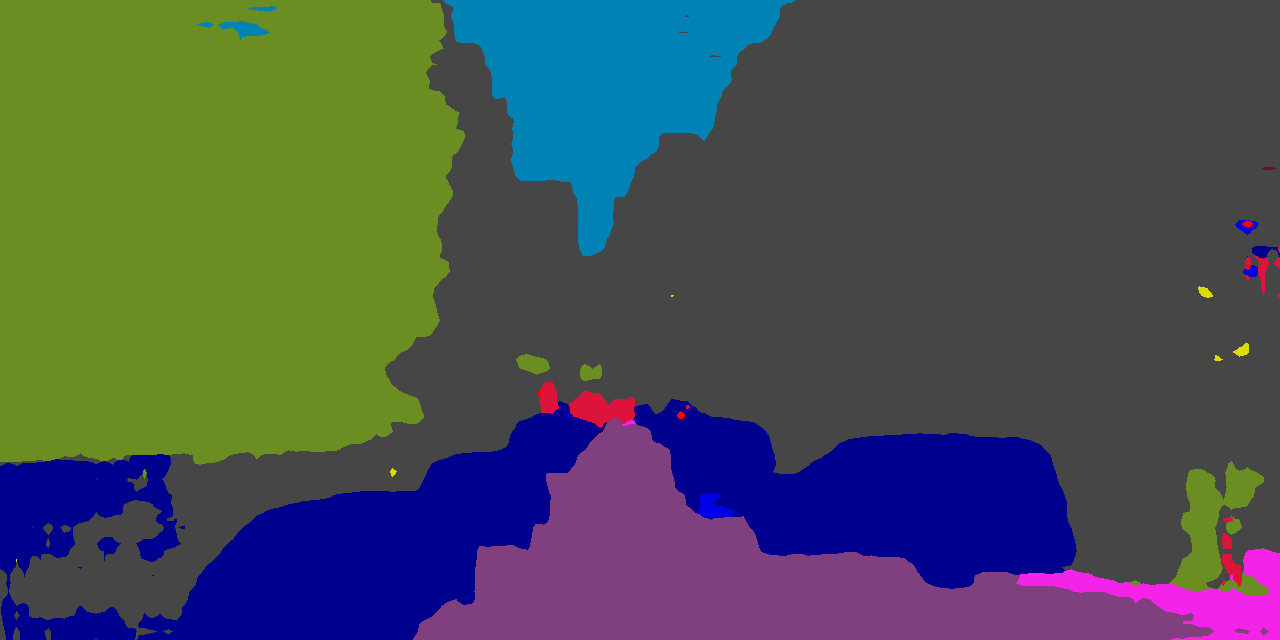}
\caption*{Source only}
\end{subfigure}%
\begin{subfigure}{\imgWidth}
\includegraphics[width=\textwidth]{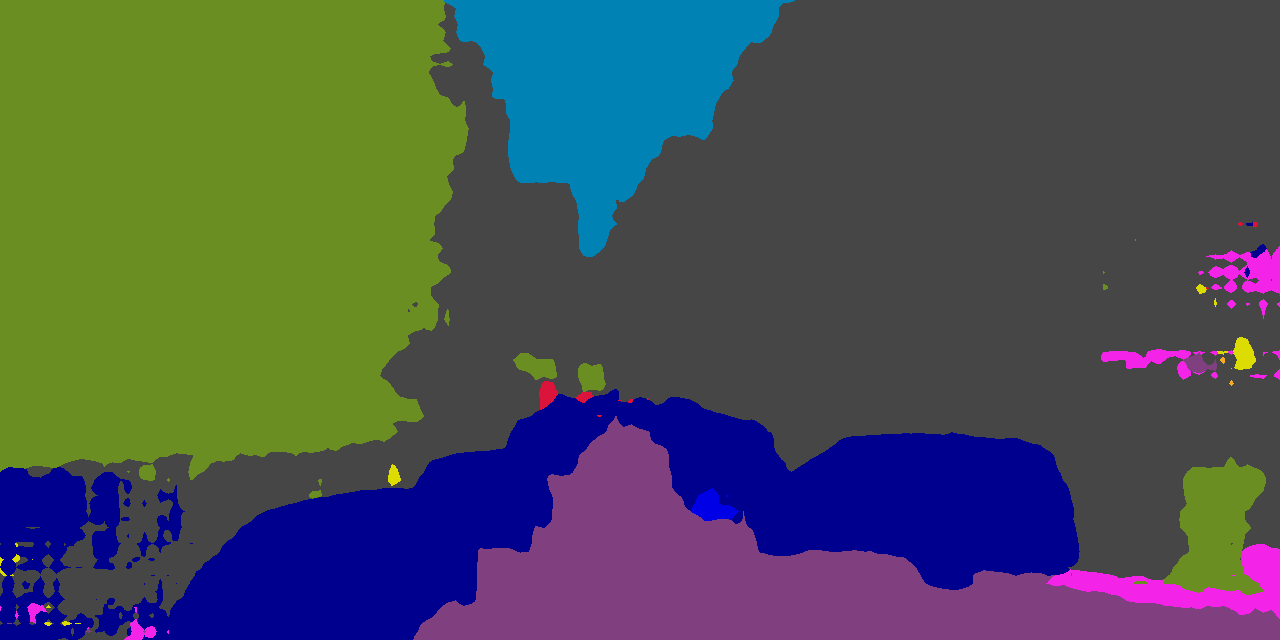}
\caption*{MaxSquareIW~\cite{Chen2019}}
\end{subfigure}%
\begin{subfigure}{\imgWidth}
\includegraphics[width=\textwidth]{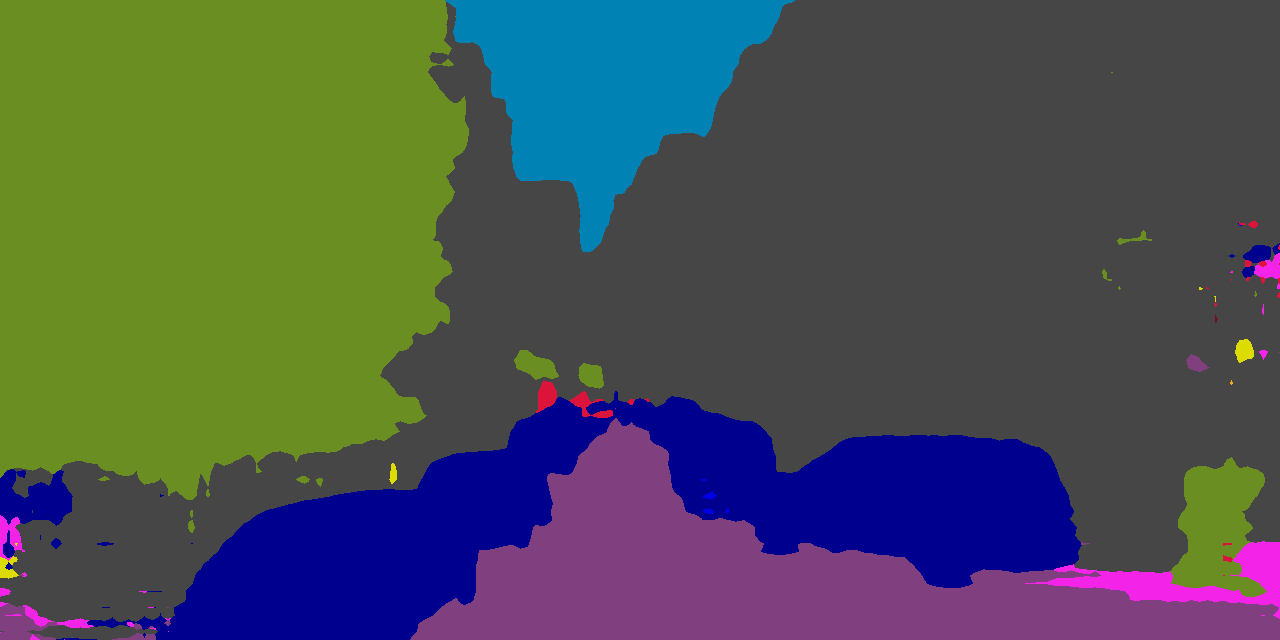}
\caption*{UDA OCE~\cite{toldo2020clustering}}
\end{subfigure}%
\begin{subfigure}{\imgWidth}
\includegraphics[width=\textwidth]{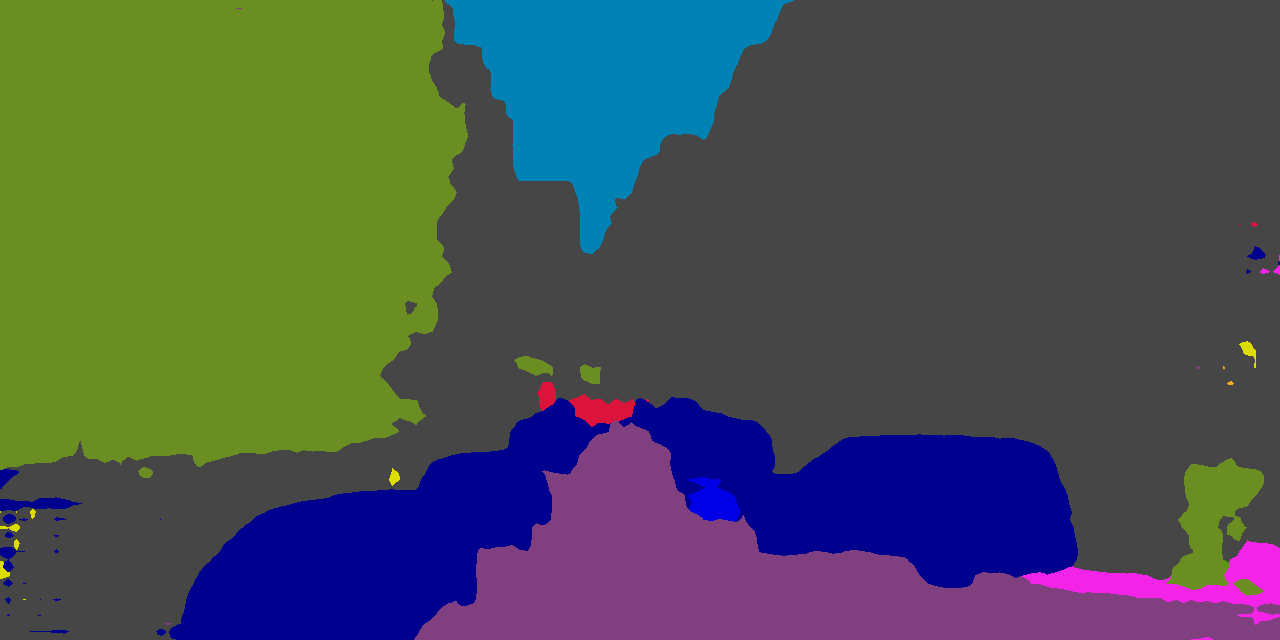}
\caption*{LSR$^+$ (ours)}
\end{subfigure}
\end{subfigure}

\caption{Qualitative results on the Cross-City benchmark.}
\label{fig:crosscity}
\end{figure*}

\renewcommand{\imgWidth}{0.142\textwidth}
\begin{figure*}
\newcolumntype{Y}{>{\centering\arraybackslash}X}
\centering
\begin{subfigure}{.6em}
\scriptsize\rotatebox{90}{~~~GTAV$\rightarrow$Cityscapes}
\end{subfigure}%
\begin{subfigure}{\textwidth-1em}
\hspace*{.2em}%
\begin{subfigure}{\imgWidth}
\includegraphics[width=\textwidth]{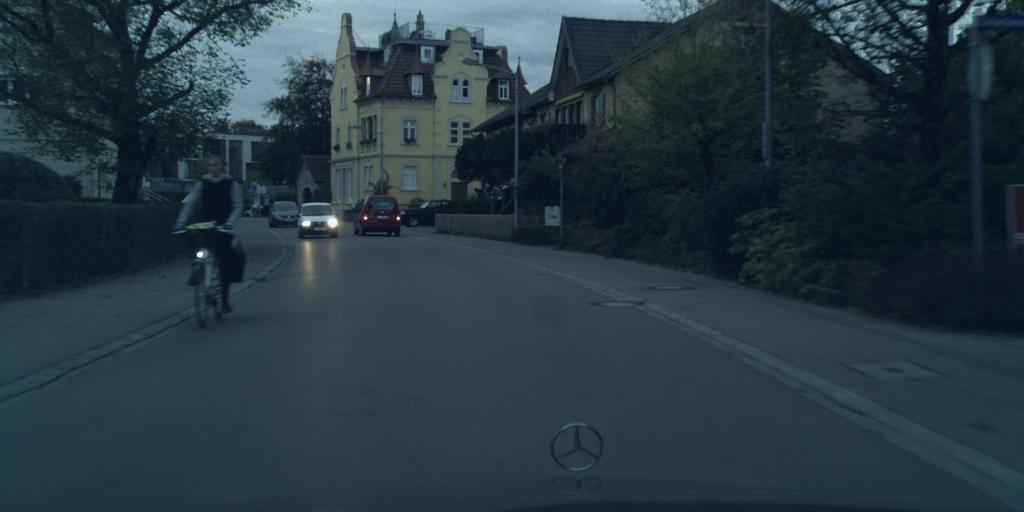}
\end{subfigure}%
\begin{subfigure}{\imgWidth}
\includegraphics[width=\textwidth]{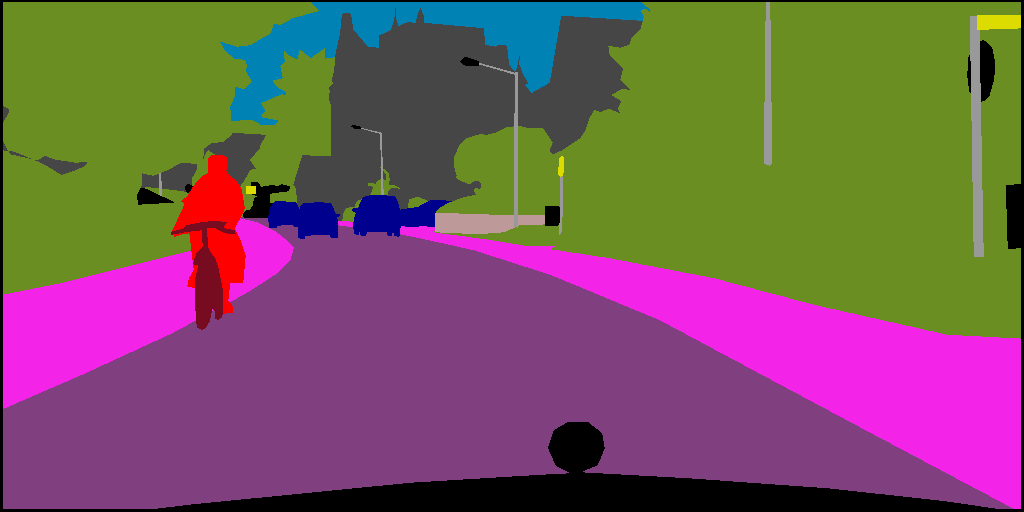}
\end{subfigure}%
\begin{subfigure}{\imgWidth}
\includegraphics[width=\textwidth]{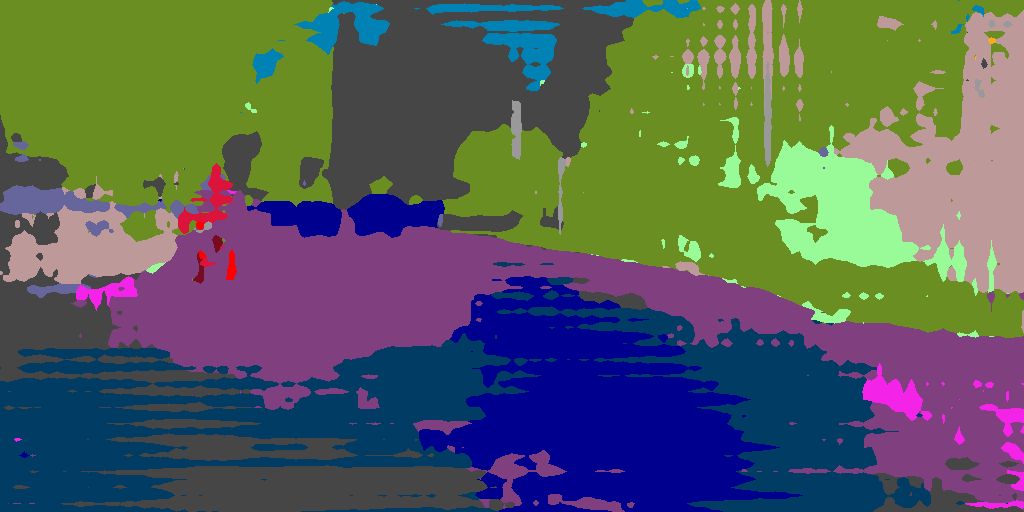}
\end{subfigure}%
\begin{subfigure}{\imgWidth}
\includegraphics[width=\textwidth]{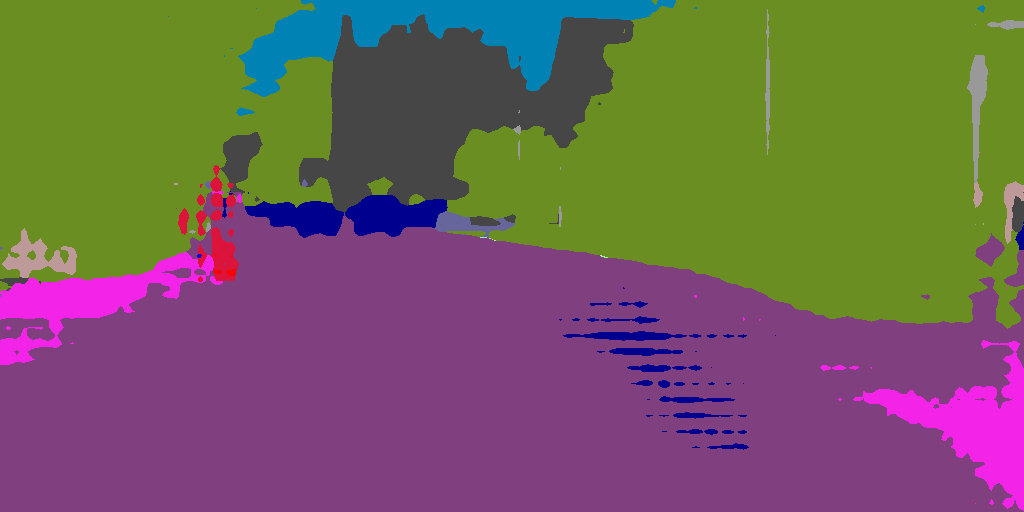}
\end{subfigure}%
\begin{subfigure}{\imgWidth}
\includegraphics[width=\textwidth]{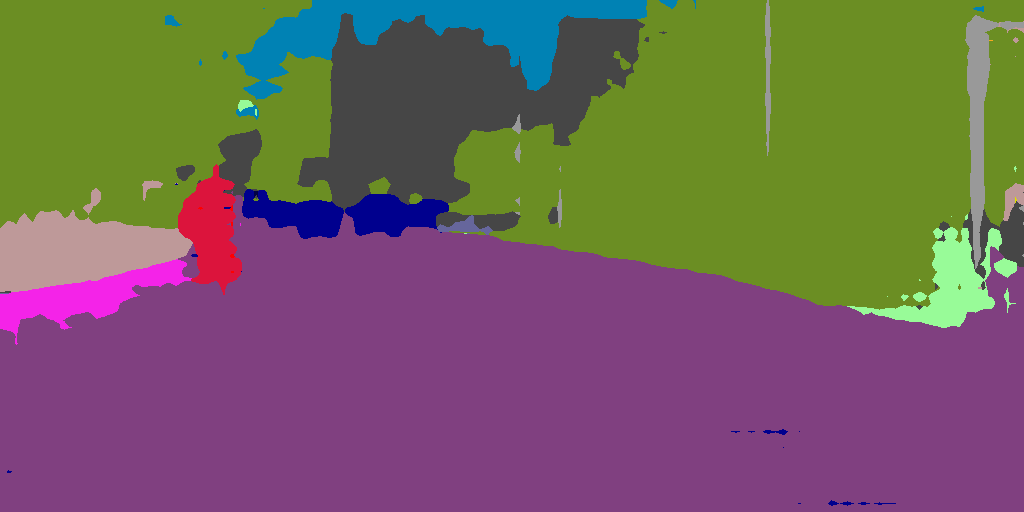}
\end{subfigure}%
\begin{subfigure}{\imgWidth}
\includegraphics[width=\textwidth]{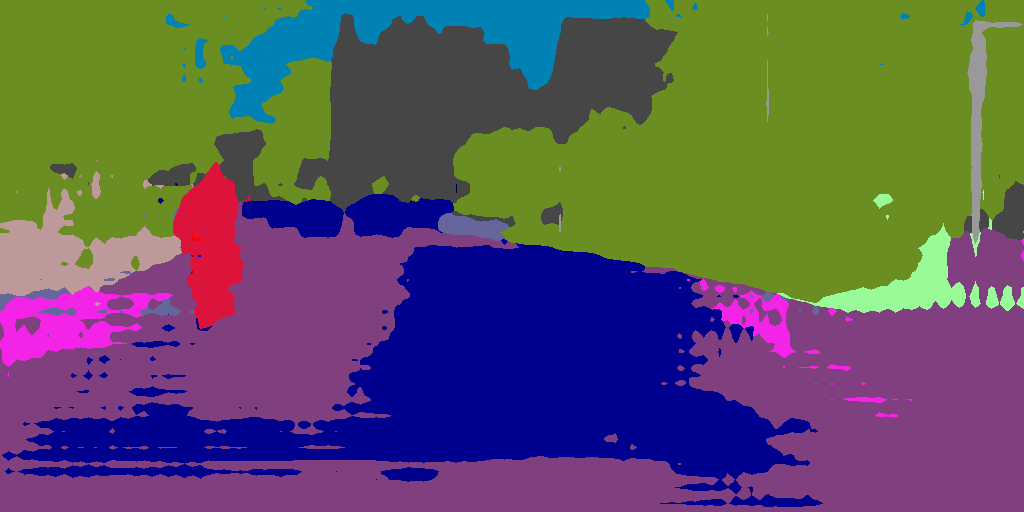}
\end{subfigure}%
\begin{subfigure}{\imgWidth}
\includegraphics[width=\textwidth]{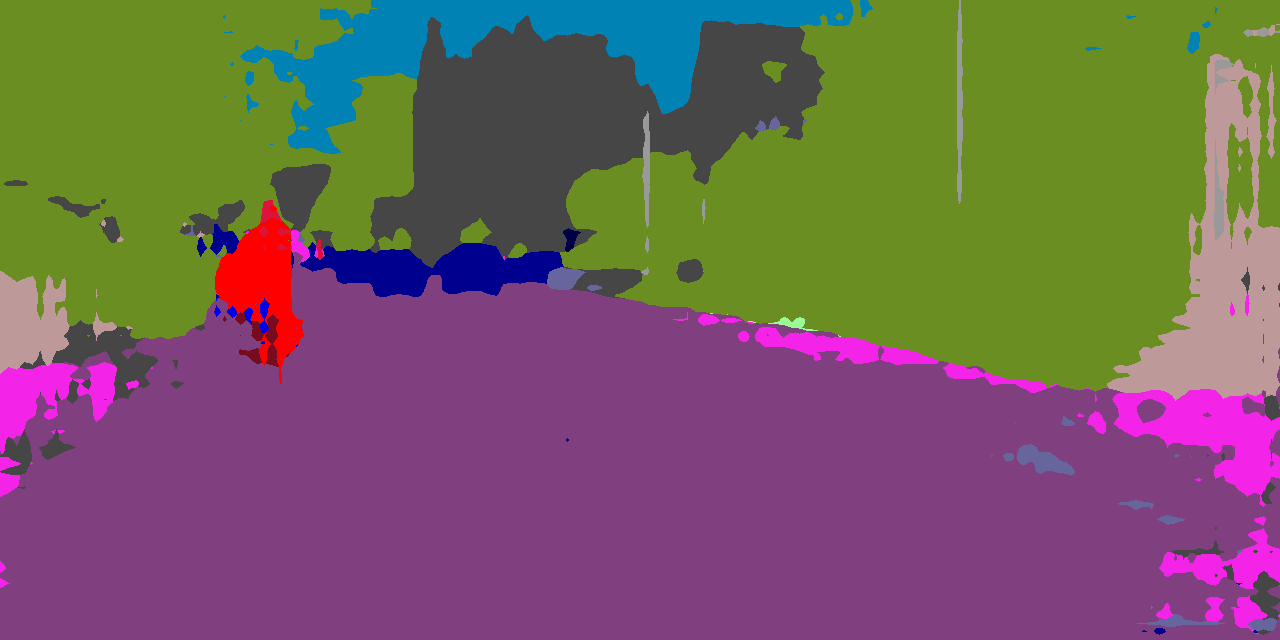}
\end{subfigure}

\hspace*{.2em}%
\begin{subfigure}{\imgWidth}
\includegraphics[width=\textwidth]{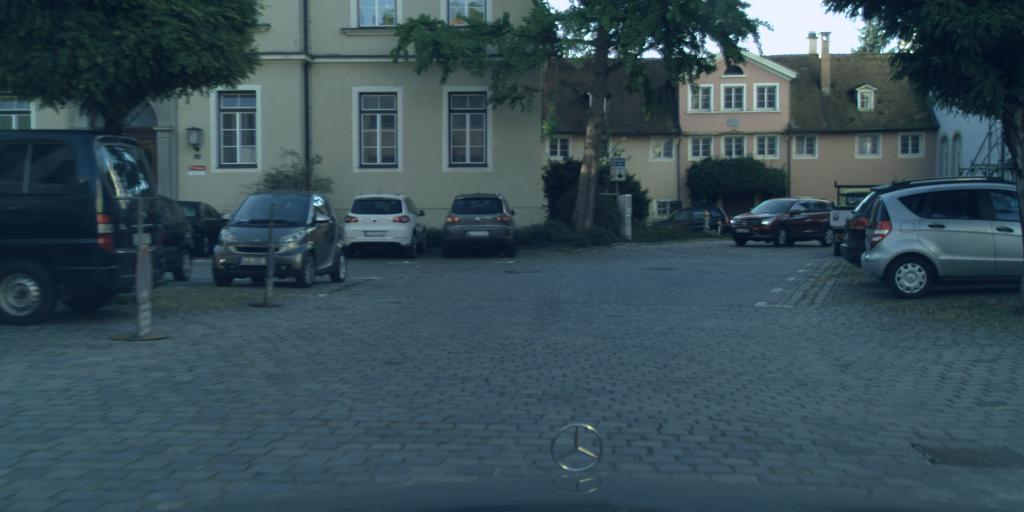}
\end{subfigure}%
\begin{subfigure}{\imgWidth}
\includegraphics[width=\textwidth]{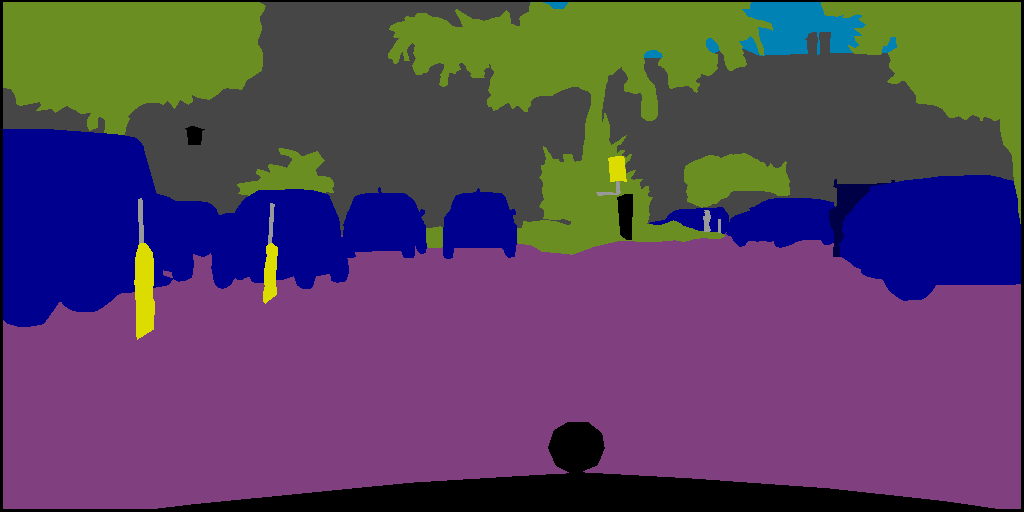}
\end{subfigure}%
\begin{subfigure}{\imgWidth}
\includegraphics[width=\textwidth]{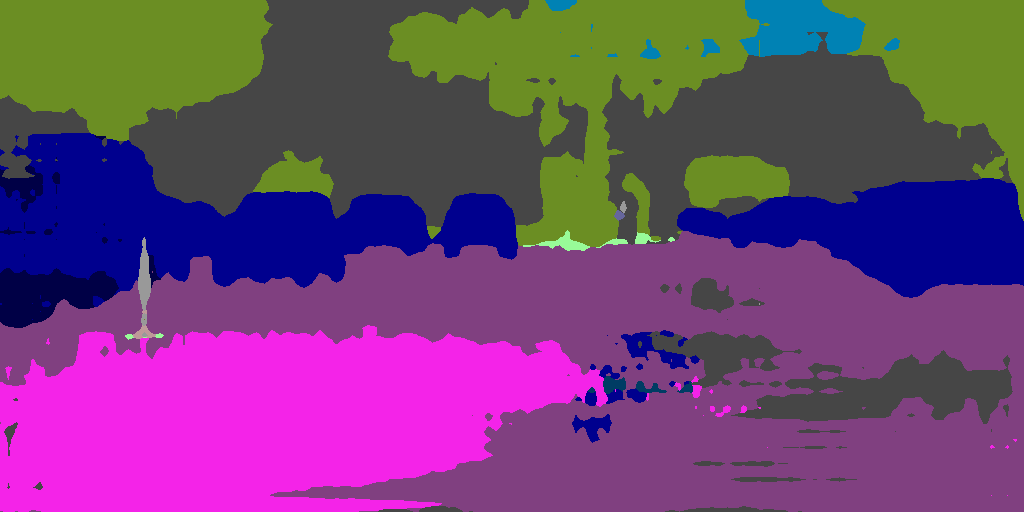}
\end{subfigure}%
\begin{subfigure}{\imgWidth}
\includegraphics[width=\textwidth]{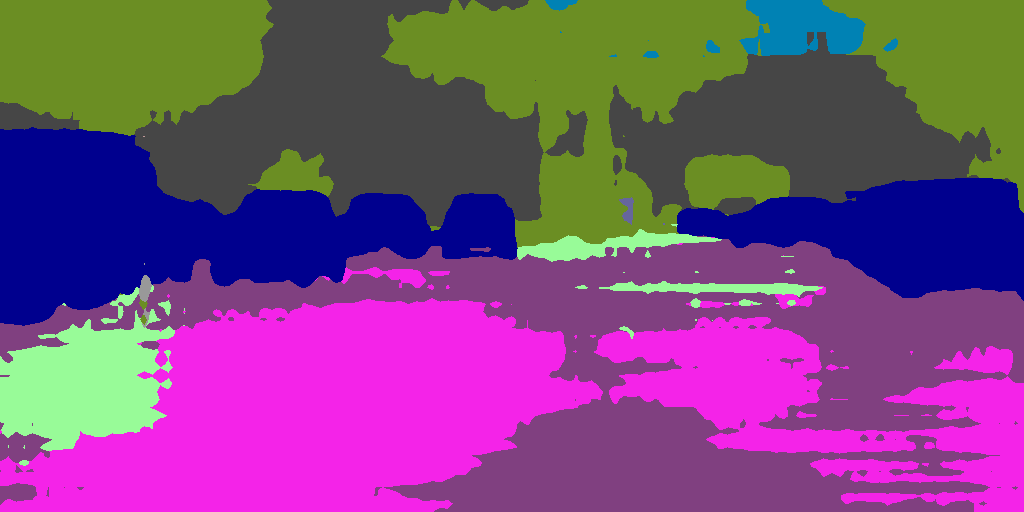}
\end{subfigure}%
\begin{subfigure}{\imgWidth}
\includegraphics[width=\textwidth]{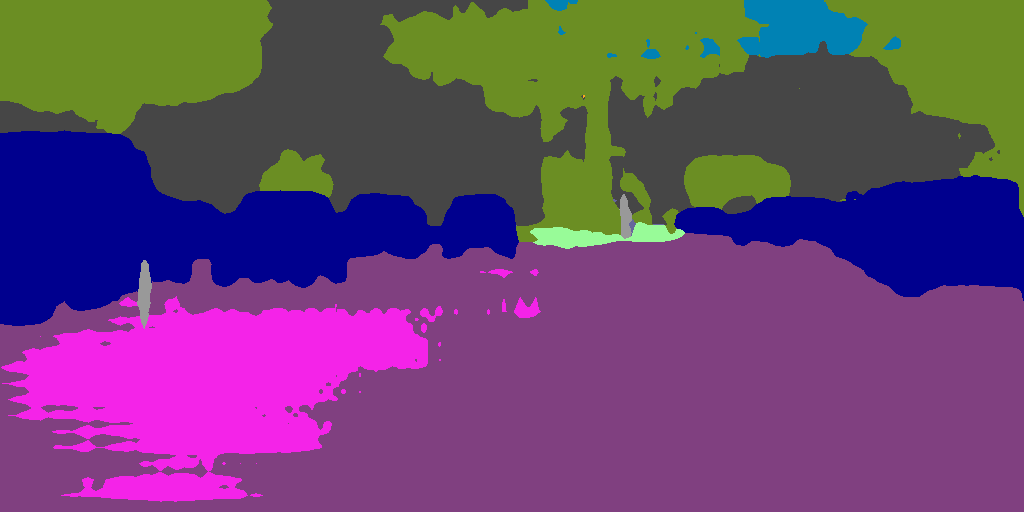}
\end{subfigure}%
\begin{subfigure}{\imgWidth}
\includegraphics[width=\textwidth]{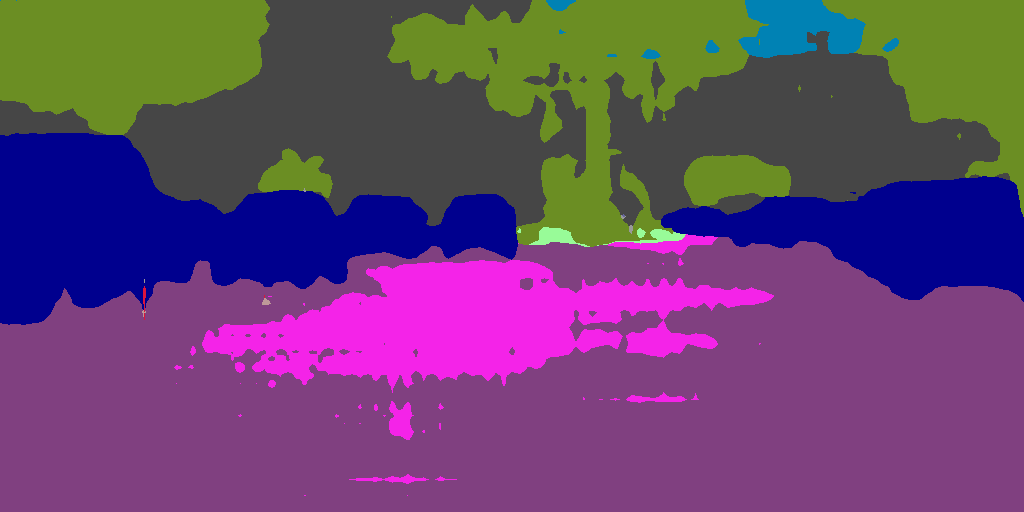}
\end{subfigure}%
\begin{subfigure}{\imgWidth}
\includegraphics[width=\textwidth]{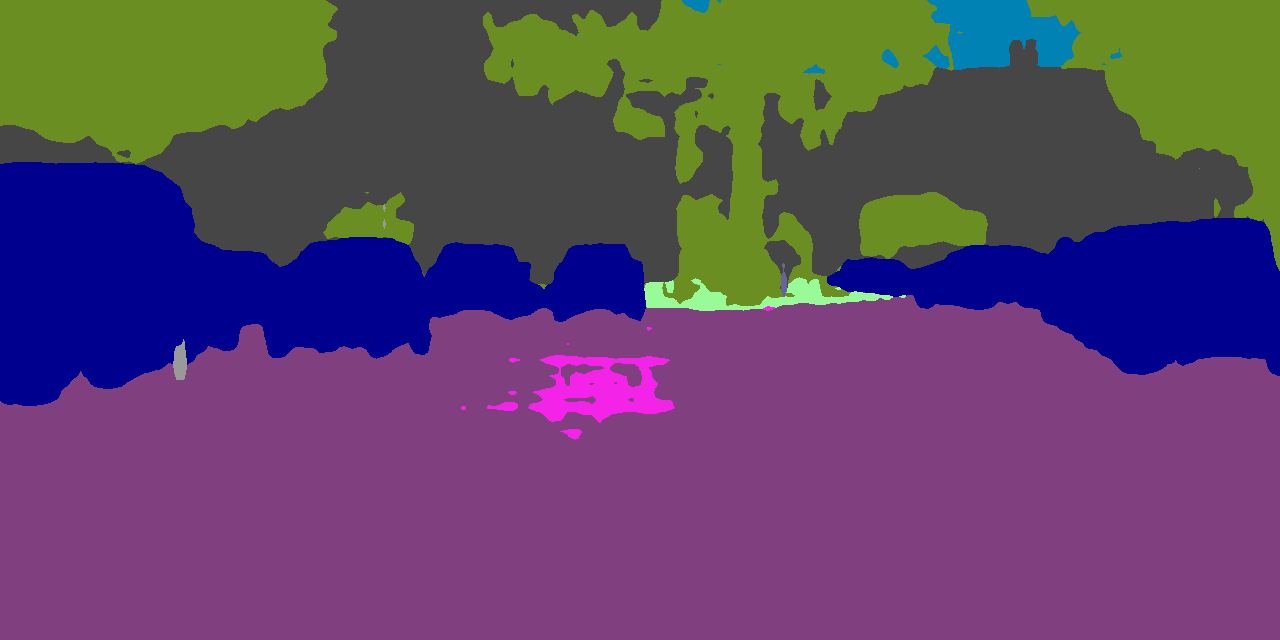}
\end{subfigure}

\hspace*{.2em}%
\begin{subfigure}{\imgWidth}
\includegraphics[width=\textwidth]{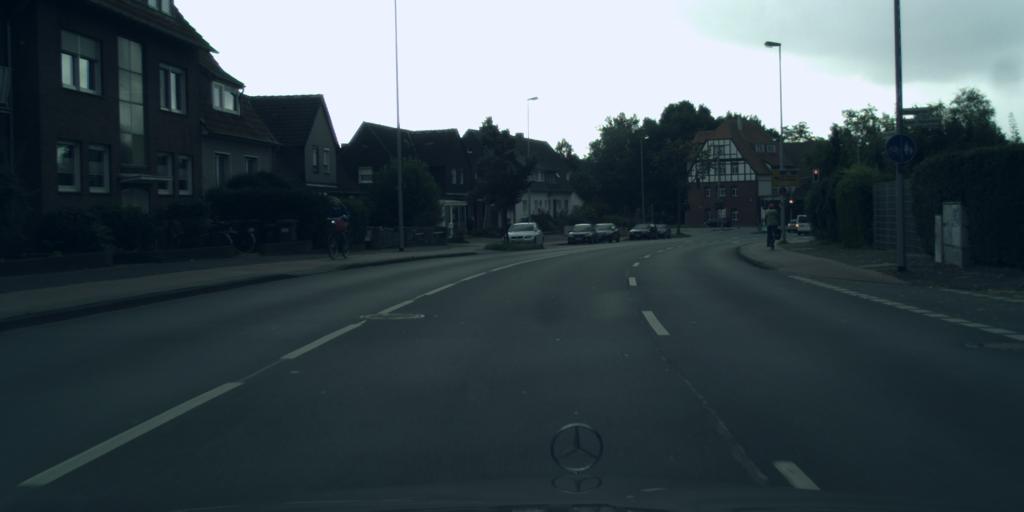}
\end{subfigure}%
\begin{subfigure}{\imgWidth}
\includegraphics[width=\textwidth]{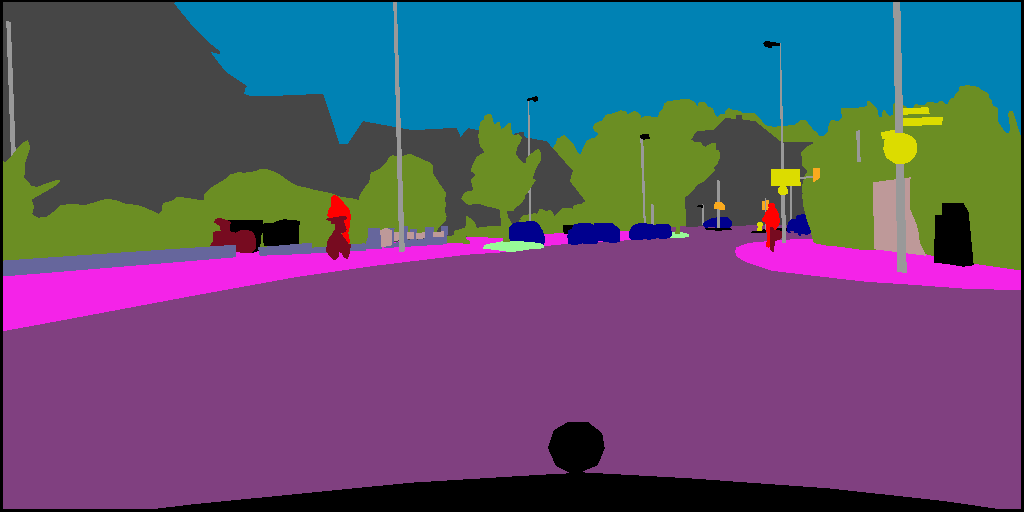}
\end{subfigure}%
\begin{subfigure}{\imgWidth}
\includegraphics[width=\textwidth]{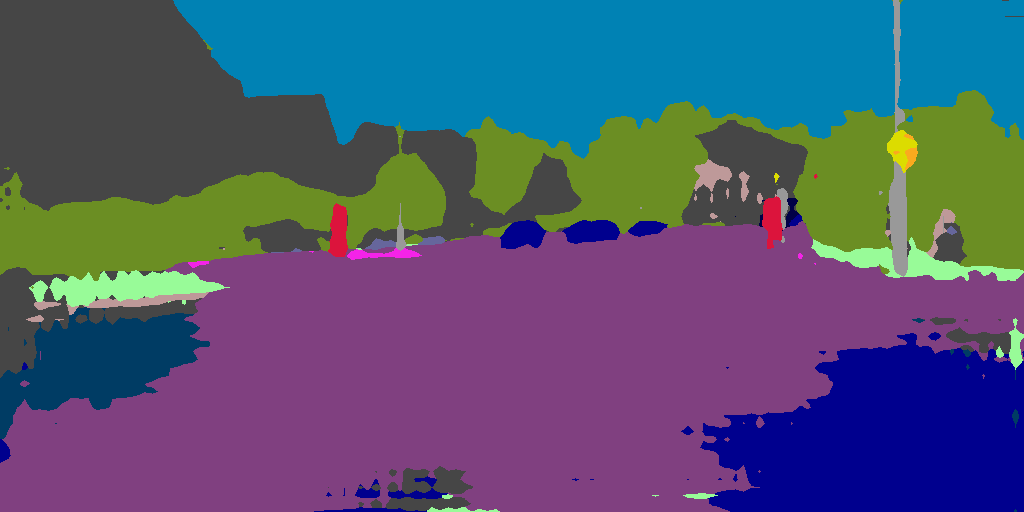}
\end{subfigure}%
\begin{subfigure}{\imgWidth}
\includegraphics[width=\textwidth]{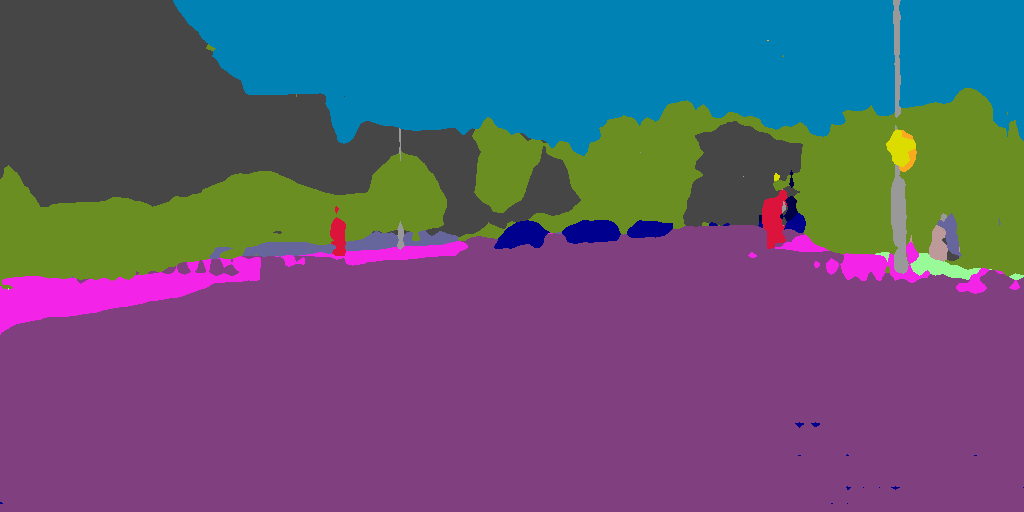}
\end{subfigure}%
\begin{subfigure}{\imgWidth}
\includegraphics[width=\textwidth]{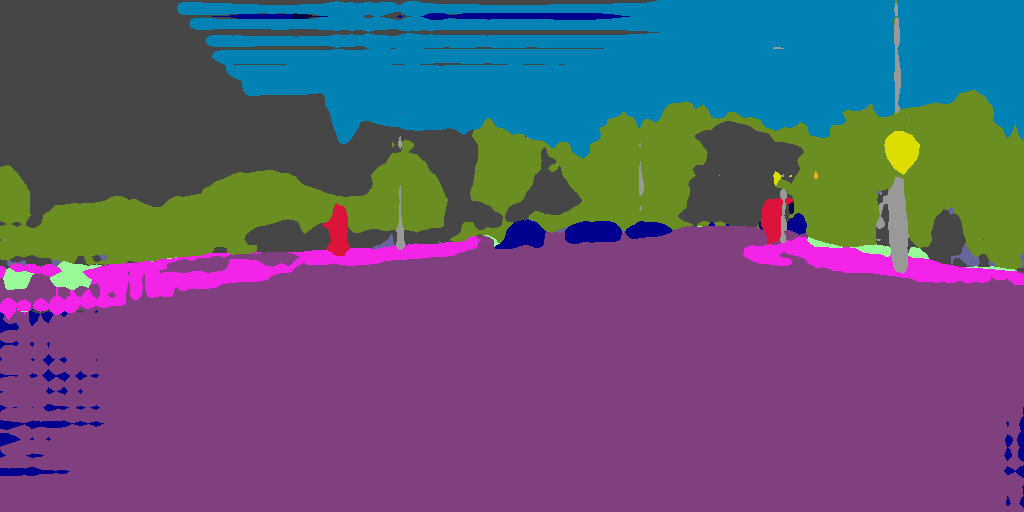}
\end{subfigure}%
\begin{subfigure}{\imgWidth}
\includegraphics[width=\textwidth]{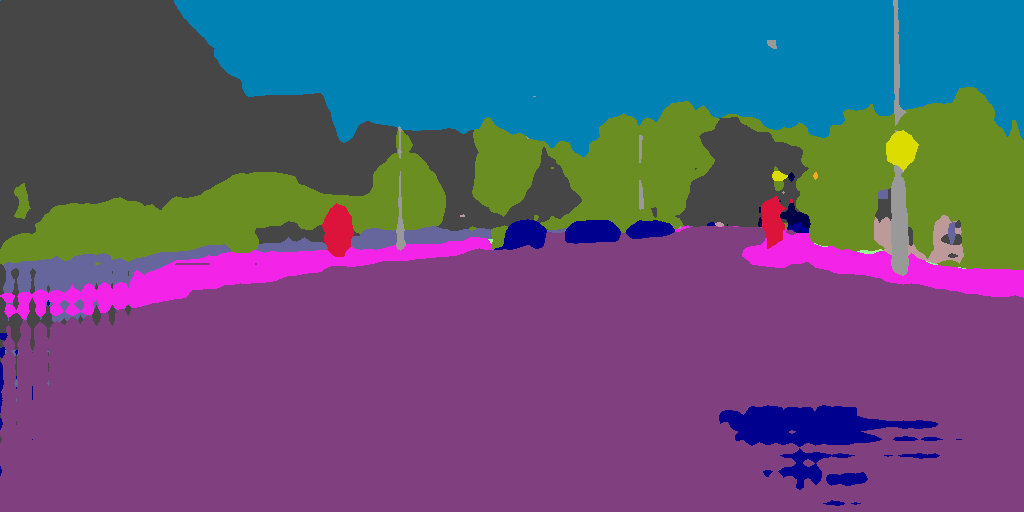}
\end{subfigure}%
\begin{subfigure}{\imgWidth}
\includegraphics[width=\textwidth]{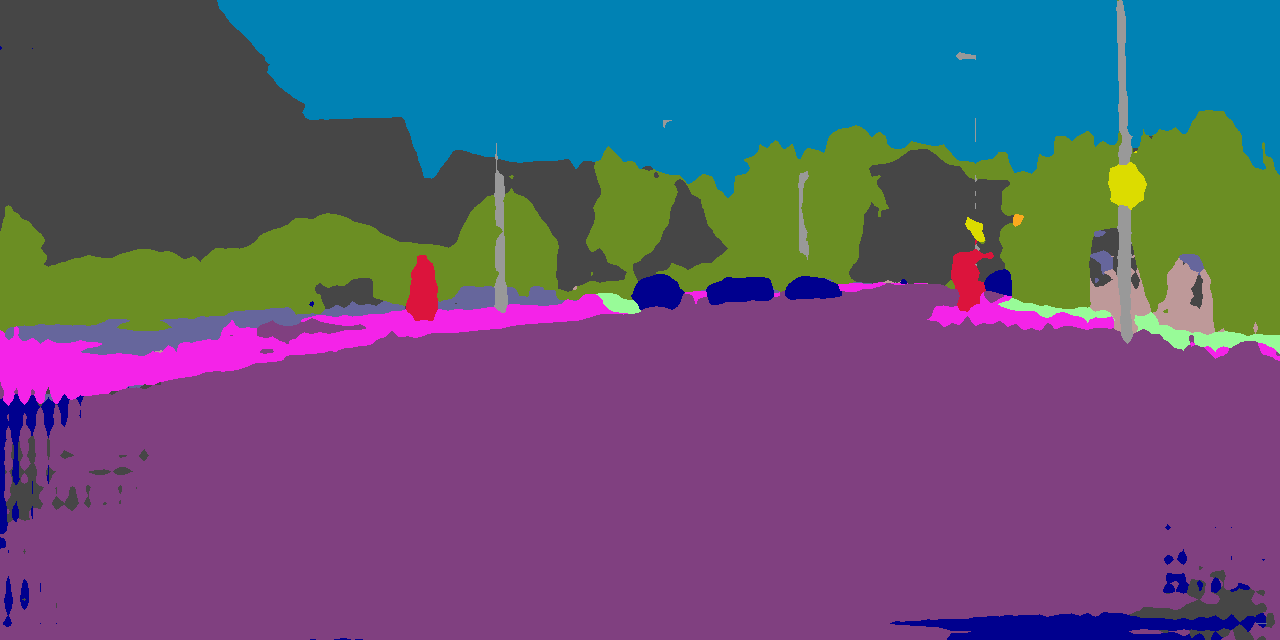}
\end{subfigure}
\end{subfigure}

\begin{subfigure}{\textwidth}
\small
\begin{tabularx}{\textwidth}{YYYYYYYYYY}
\cellcolor{road} \textcolor{white}{Road} & \cellcolor{sidewalk} Sidewalk & \cellcolor{building} \textcolor{white}{Building} & \cellcolor{wall} \textcolor{white}{Wall} & \cellcolor{fence} Fence & \cellcolor{pole} Pole & \cellcolor{tlight} T. Light & \cellcolor{tsign} T. Sign & \cellcolor{vegetation} \textcolor{white}{\footnotesize Vegetation} & \cellcolor{terrain} Terrain \\
\cellcolor{sky} Sky & \cellcolor{person} \textcolor{white}{Person} & \cellcolor{rider} \textcolor{white}{Rider} & \cellcolor{car} \textcolor{white}{Car} & \cellcolor{truck} \textcolor{white}{Truck} & \cellcolor{bus} \textcolor{white}{Bus} &  \cellcolor{train} \textcolor{white}{Train} & \cellcolor{motorbike} \footnotesize \textcolor{white}{Motorbike} & \cellcolor{bicycle} \textcolor{white}{Bicycle} & \cellcolor{unlabelled} \footnotesize \textcolor{white}{Unlabeled}
\end{tabularx}
\end{subfigure}
\begin{subfigure}{.6em}
\scriptsize\rotatebox{90}{~~~~~SYNTHIA$\rightarrow$Cityscapes}
\end{subfigure}%
\begin{subfigure}{\textwidth-1em}
\hspace*{.2em}%
\begin{subfigure}{\imgWidth}
\includegraphics[width=\textwidth]{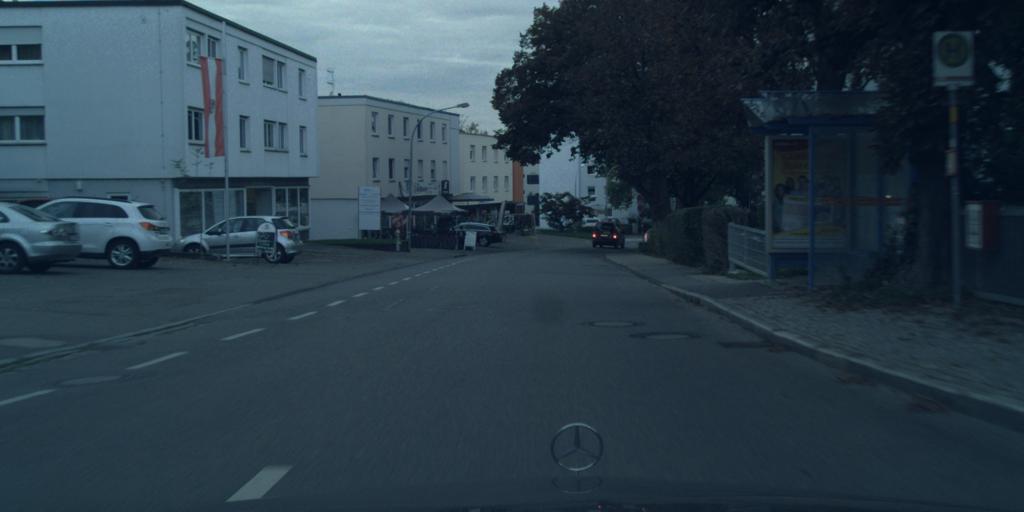}
\end{subfigure}%
\begin{subfigure}{\imgWidth}
\includegraphics[width=\textwidth]{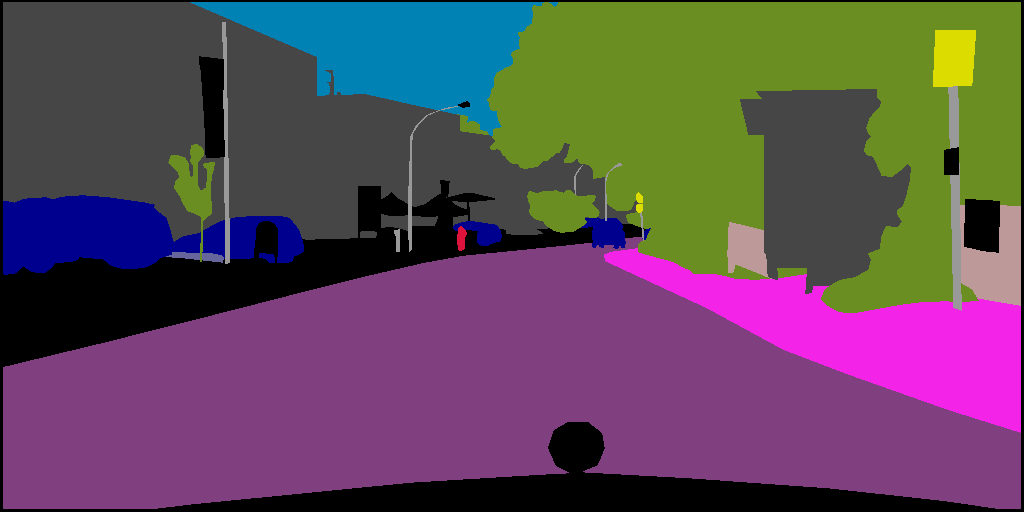}
\end{subfigure}%
\begin{subfigure}{\imgWidth}
\includegraphics[width=\textwidth]{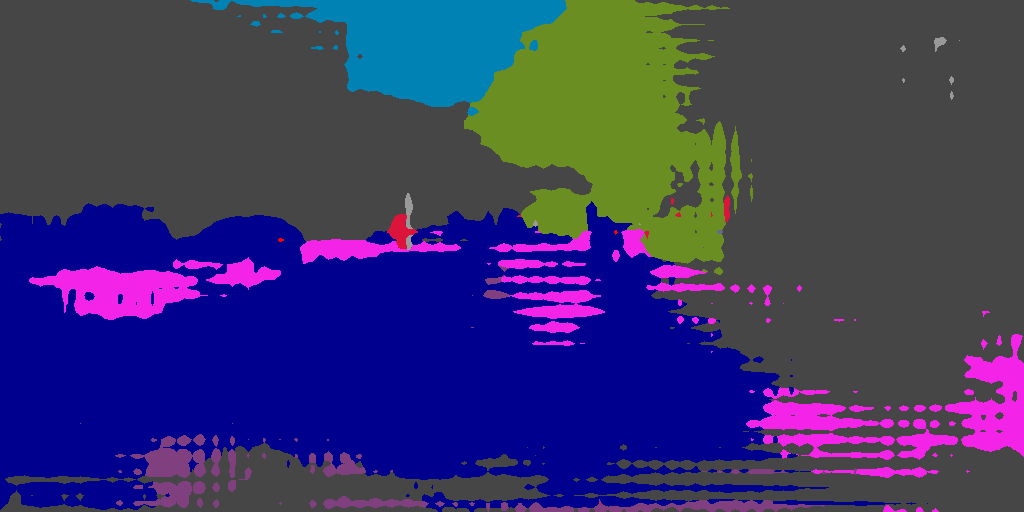}
\end{subfigure}%
\begin{subfigure}{\imgWidth}
\includegraphics[width=\textwidth]{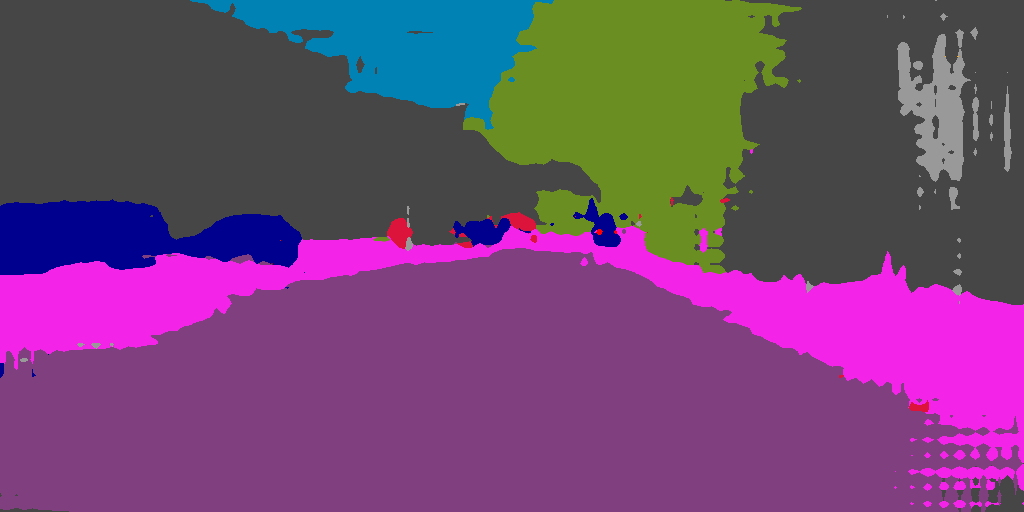}
\end{subfigure}%
\begin{subfigure}{\imgWidth}
\includegraphics[width=\textwidth]{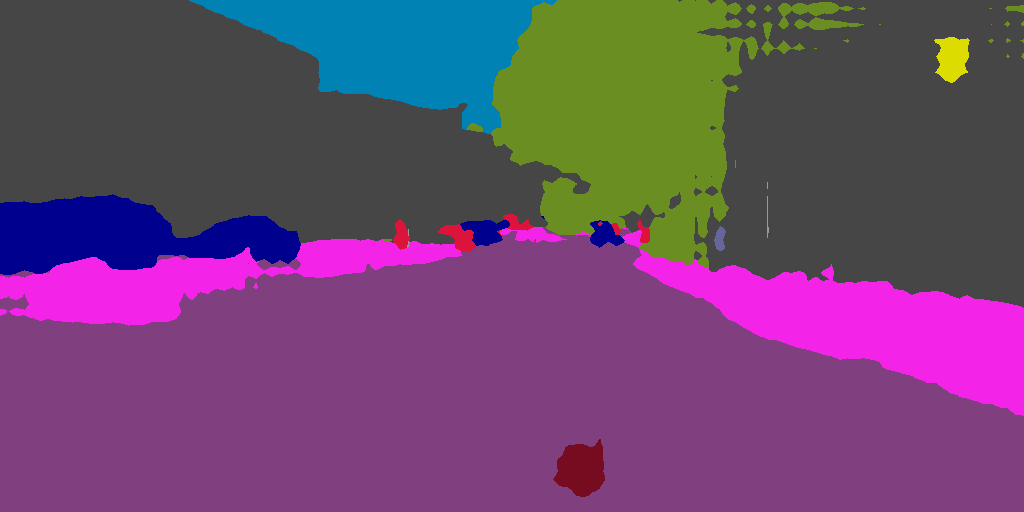}
\end{subfigure}%
\begin{subfigure}{\imgWidth}
\includegraphics[width=\textwidth]{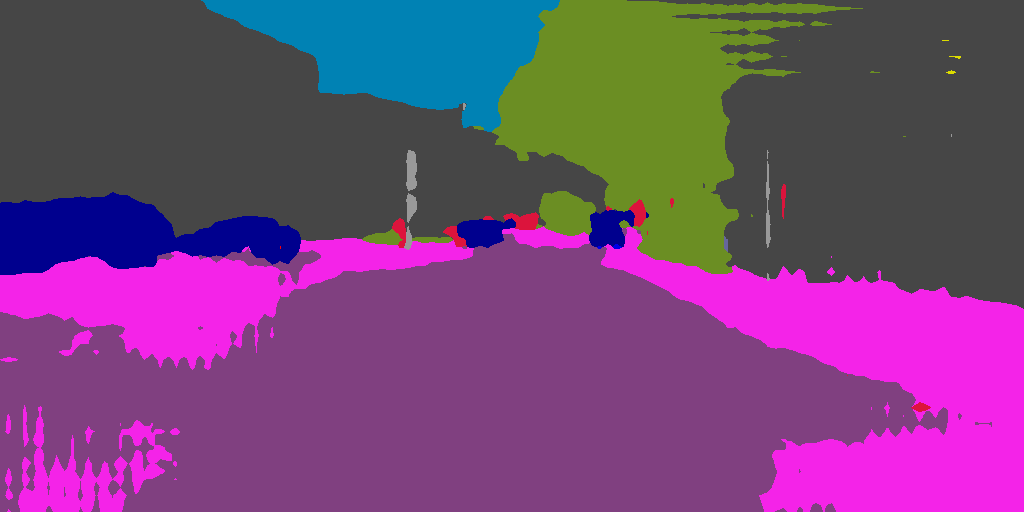}
\end{subfigure}%
\begin{subfigure}{\imgWidth}
\includegraphics[width=\textwidth]{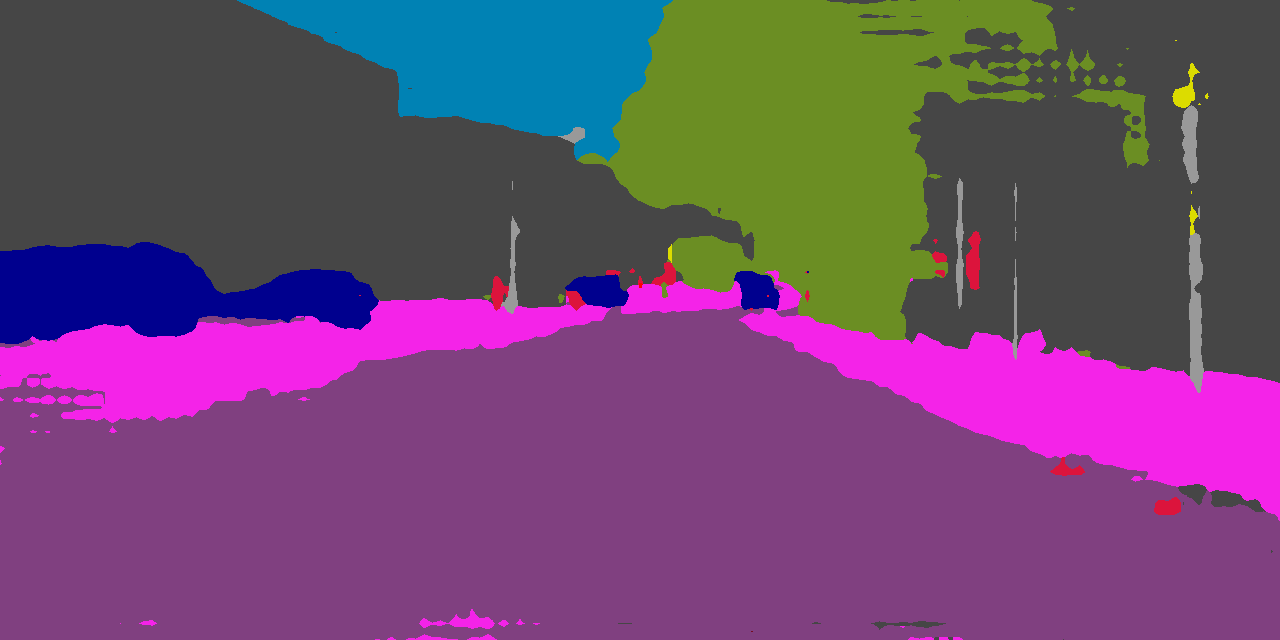}
\end{subfigure}

\hspace*{.2em}%
\begin{subfigure}{\imgWidth}
\includegraphics[width=\textwidth]{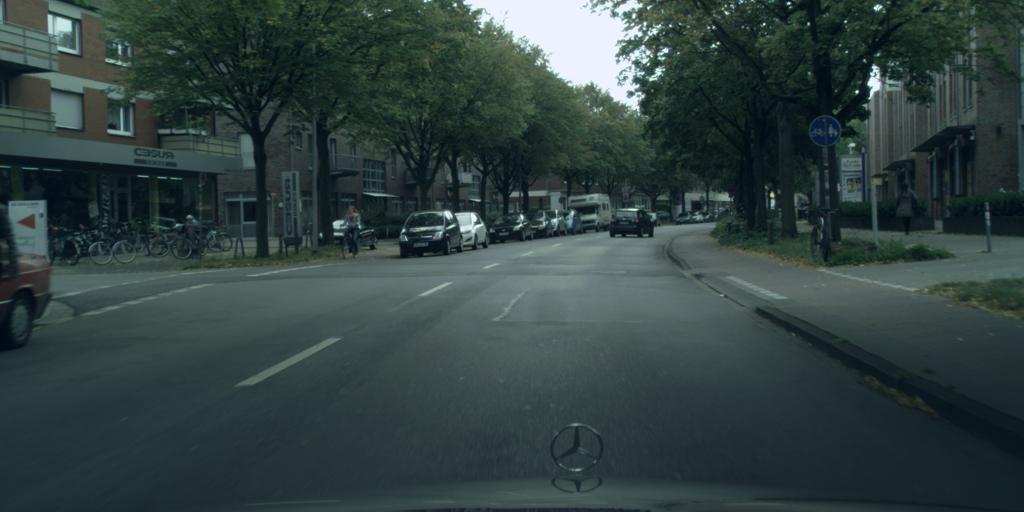}
\end{subfigure}%
\begin{subfigure}{\imgWidth}
\includegraphics[width=\textwidth]{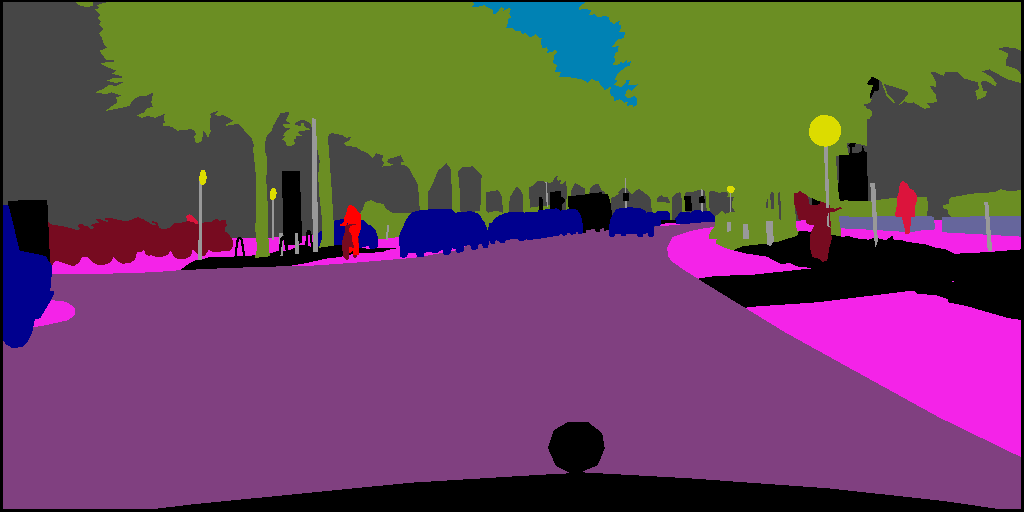}
\end{subfigure}%
\begin{subfigure}{\imgWidth}
\includegraphics[width=\textwidth]{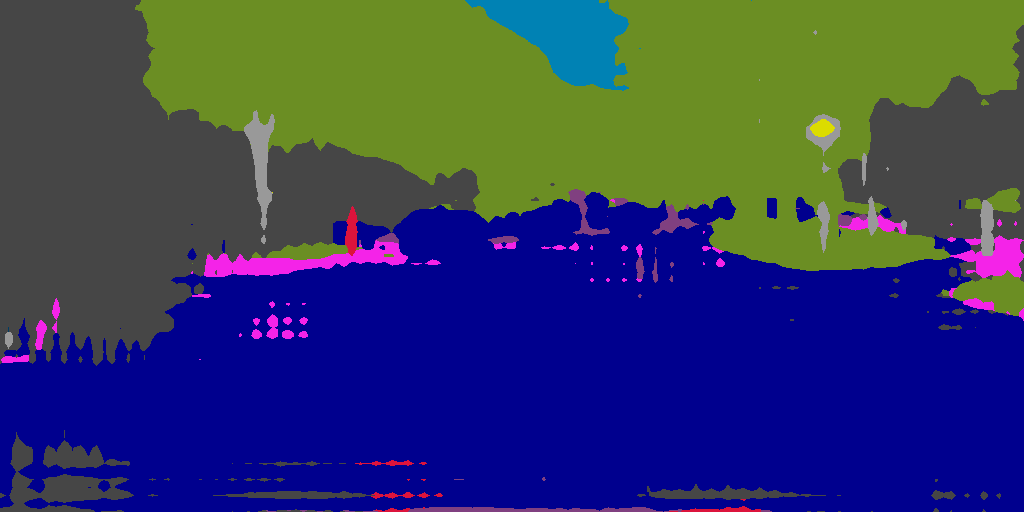}
\end{subfigure}%
\begin{subfigure}{\imgWidth}
\includegraphics[width=\textwidth]{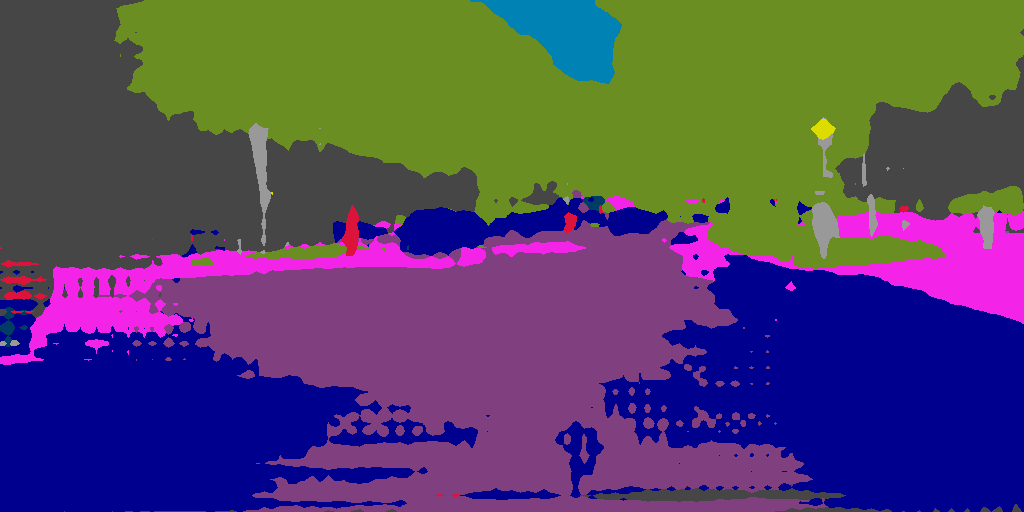}
\end{subfigure}%
\begin{subfigure}{\imgWidth}
\includegraphics[width=\textwidth]{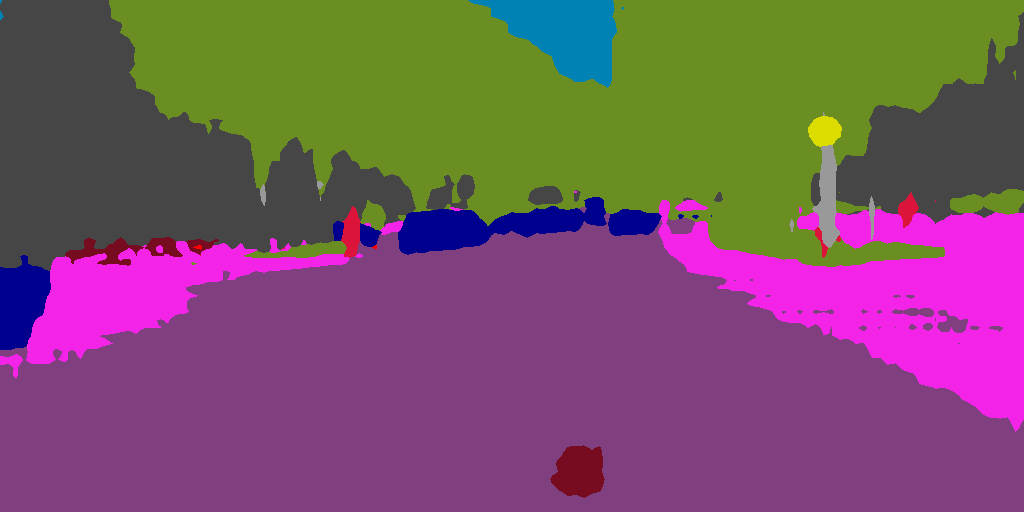}
\end{subfigure}%
\begin{subfigure}{\imgWidth}
\includegraphics[width=\textwidth]{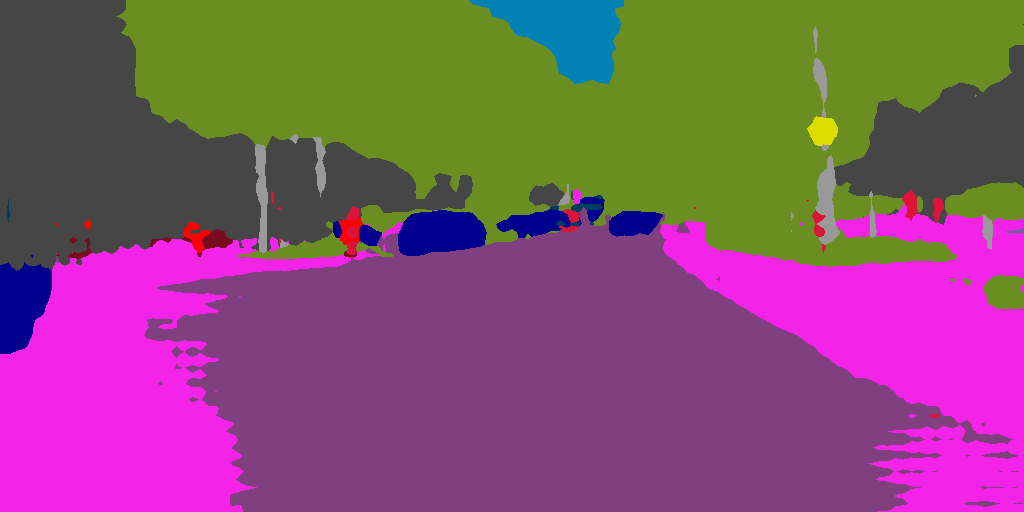}
\end{subfigure}%
\begin{subfigure}{\imgWidth}
\includegraphics[width=\textwidth]{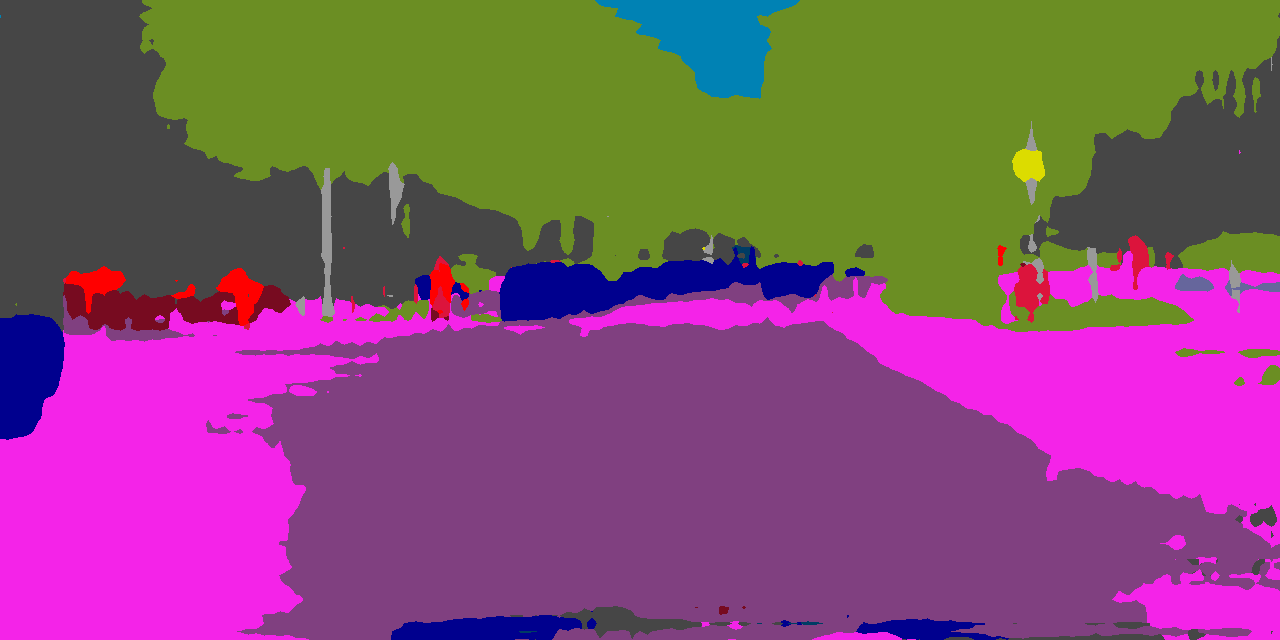}
\end{subfigure}

\hspace*{.2em}%
\begin{subfigure}{\imgWidth}
\includegraphics[width=\textwidth]{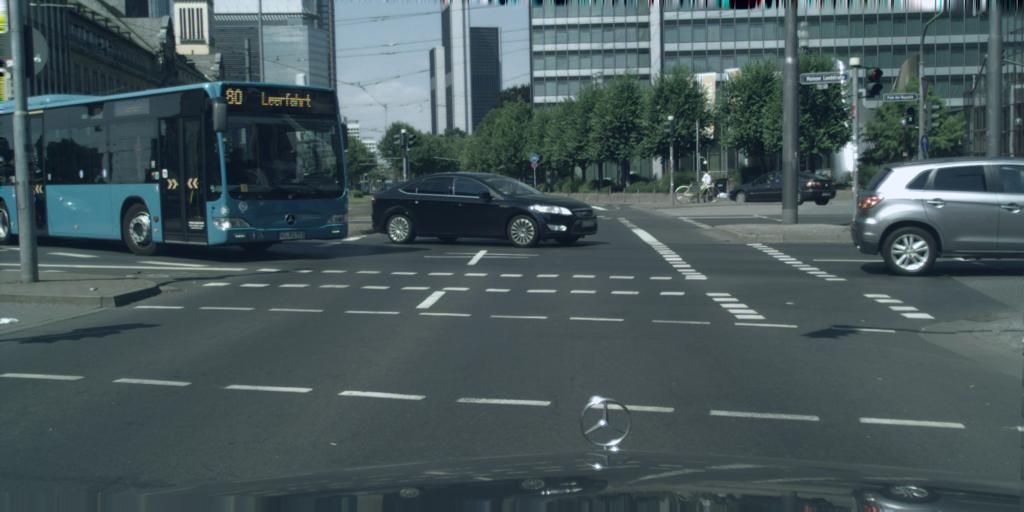}
\caption*{RGB}
\end{subfigure}%
\begin{subfigure}{\imgWidth}
\includegraphics[width=\textwidth]{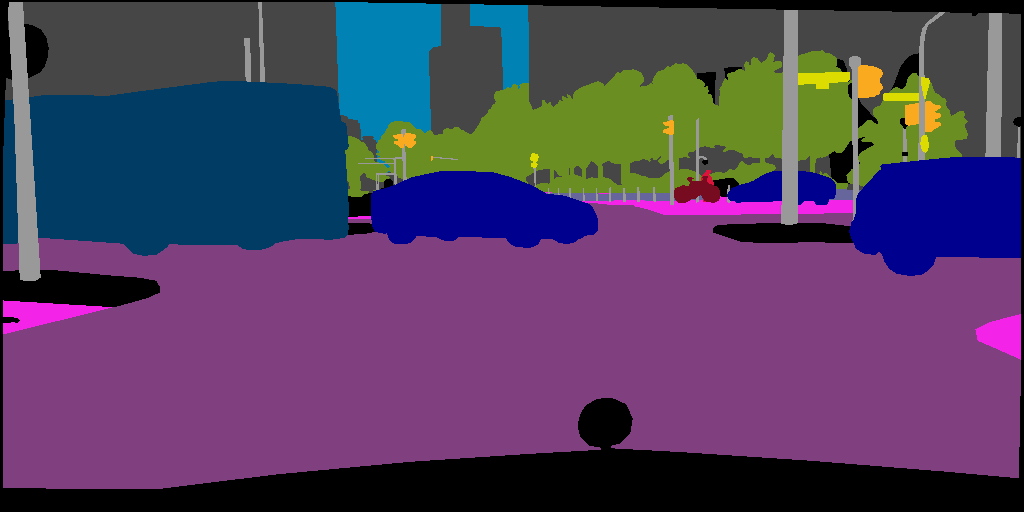}
\caption*{Ground truth}
\end{subfigure}%
\begin{subfigure}{\imgWidth}
\includegraphics[width=\textwidth]{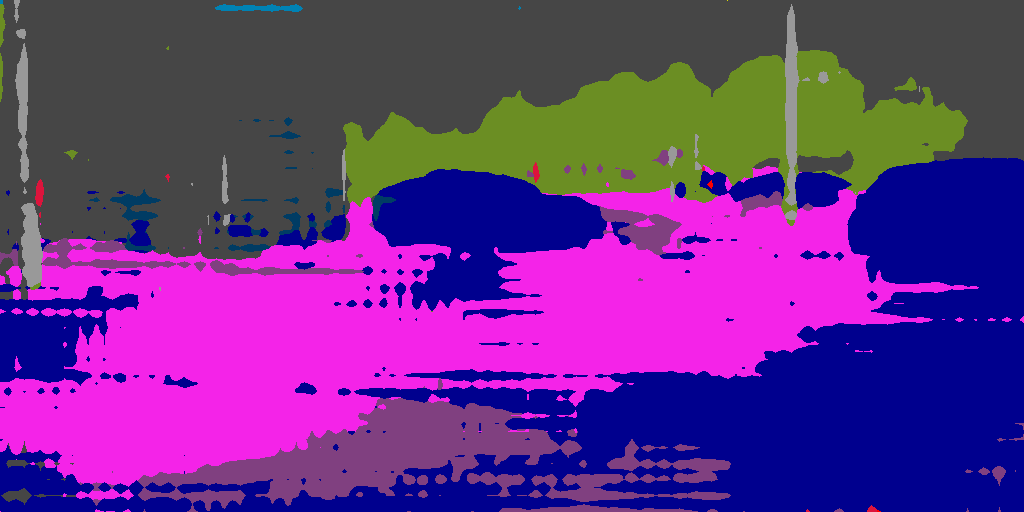}
\caption*{Source only}
\end{subfigure}%
\begin{subfigure}{\imgWidth}
\includegraphics[width=\textwidth]{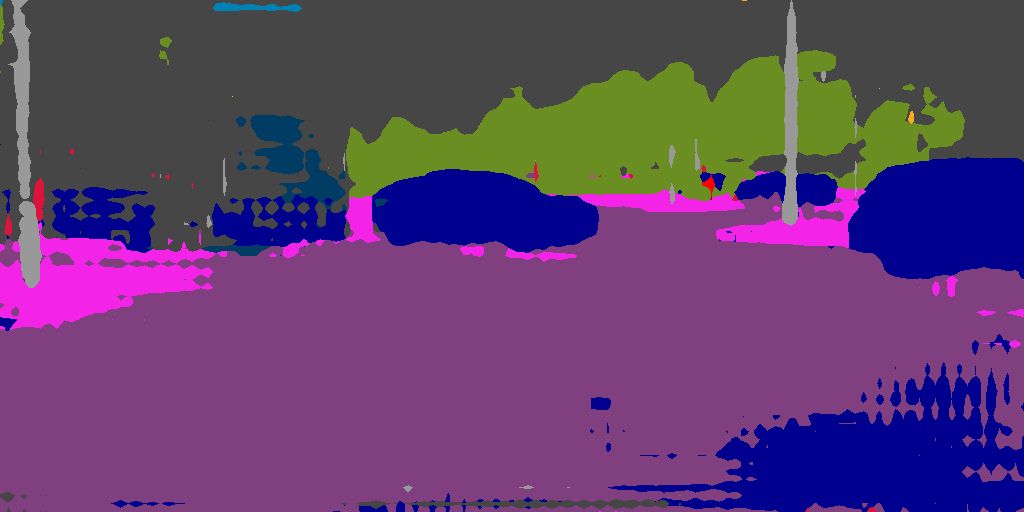}
\caption*{MaxSquareIW~\cite{Chen2019}}
\end{subfigure}%
\begin{subfigure}{\imgWidth}
\includegraphics[width=\textwidth]{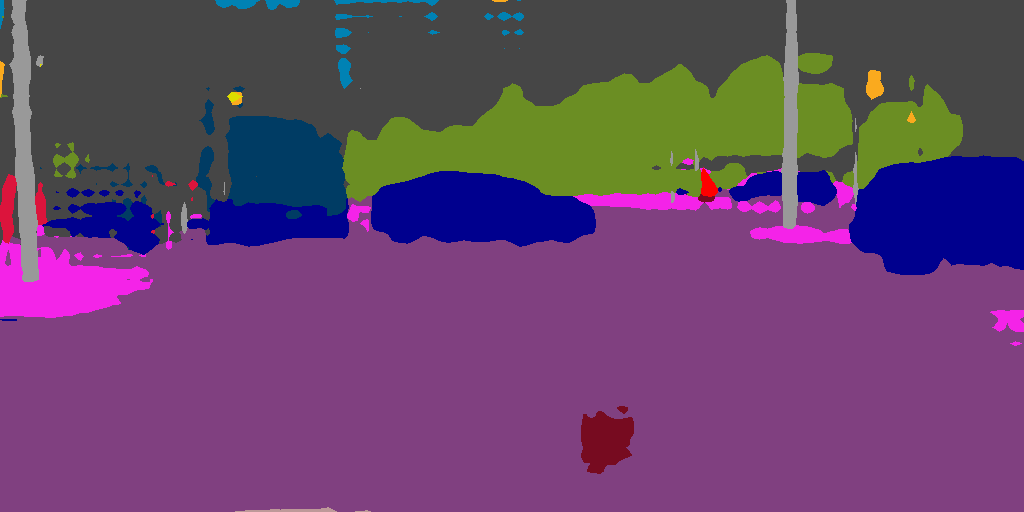}
\caption*{UDA OCE~\cite{toldo2020clustering}}
\end{subfigure}%
\begin{subfigure}{\imgWidth}
\includegraphics[width=\textwidth]{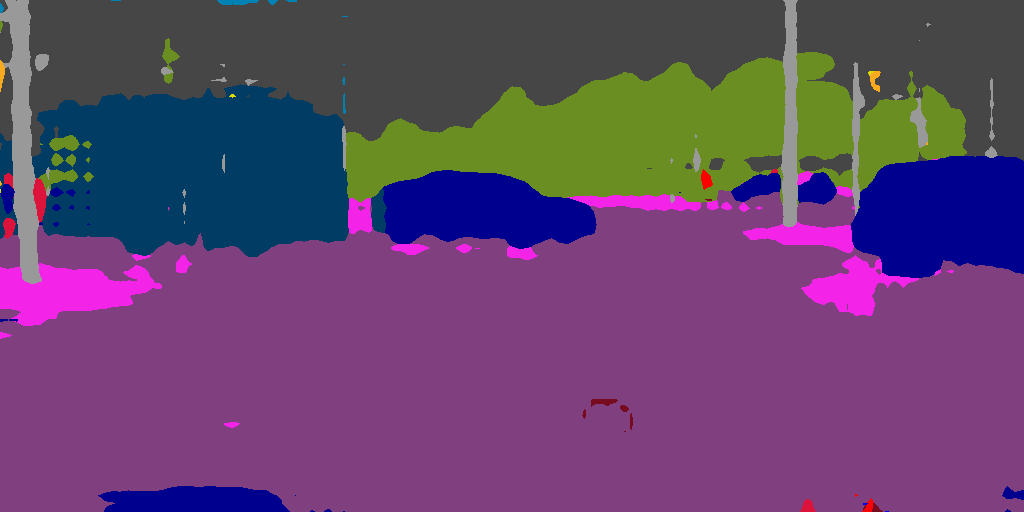}
\caption*{LSR~\cite{barbato2021latent}}
\end{subfigure}%
\begin{subfigure}{\imgWidth}
\includegraphics[width=\textwidth]{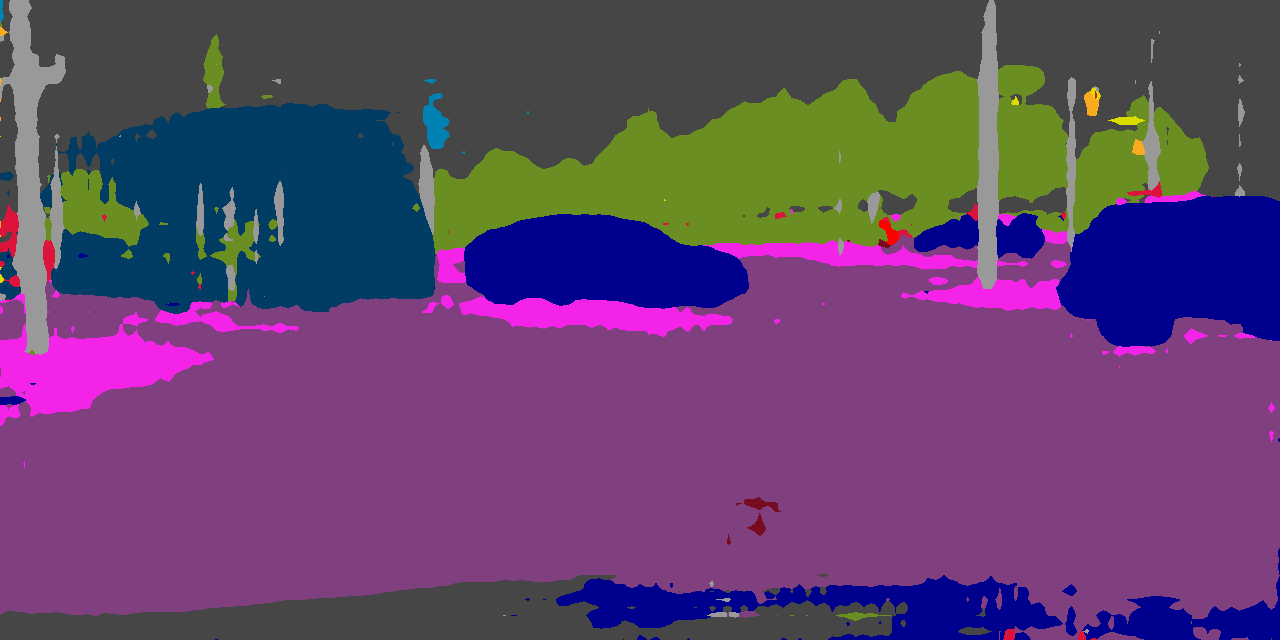}
\caption*{LSR$^{+}$ (ours)}
\end{subfigure}
\end{subfigure}
\caption{Qualitative results on sample scenes taken from the \textit{Cityscapes} validation split.}
\label{fig:cityscapes}
\end{figure*}

%% file: main.bbl
\begin{thebibliography}{1}
\providecommand{\url}[1]{#1}
\csname url@samestyle\endcsname
\providecommand{\newblock}{\relax}
\providecommand{\bibinfo}[2]{#2}
\providecommand{\BIBentrySTDinterwordspacing}{\spaceskip=0pt\relax}
\providecommand{\BIBentryALTinterwordstretchfactor}{4}
\providecommand{\BIBentryALTinterwordspacing}{\spaceskip=\fontdimen2\font plus
\BIBentryALTinterwordstretchfactor\fontdimen3\font minus
  \fontdimen4\font\relax}
\providecommand{\BIBforeignlanguage}[2]{{%
\expandafter\ifx\csname l@#1\endcsname\relax
\typeout{** WARNING: IEEEtran.bst: No hyphenation pattern has been}%
\typeout{** loaded for the language `#1'. Using the pattern for}%
\typeout{** the default language instead.}%
\else
\language=\csname l@#1\endcsname
\fi
#2}}
\providecommand{\BIBdecl}{\relax}
\BIBdecl

\bibitem{_he2016deep}
K.~He, X.~Zhang, S.~Ren, and J.~Sun, ``Deep residual learning for image
  recognition,'' in \emph{Proceedings of the IEEE Conference on Computer Vision
  and Pattern Recognition}, 2016, pp. 770--778.

\bibitem{_simonyan2015very}
K.~Simonyan and A.~Zisserman, ``Very deep convolutional networks for
  large-scale image recognition,'' in \emph{Proceedings of the International
  Conference on Learning Representations}, 2015.

\bibitem{_Chen2019}
M.~Chen, H.~Xue, and D.~Cai, ``Domain adaptation for semantic segmentation with
  maximum squares loss,'' in \emph{Proceedings of the International Conference
  on Computer Vision}, 2019, pp. 2090--2099.

\bibitem{_toldo2020clustering}
M.~Toldo, U.~Michieli, and P.~Zanuttigh, ``Unsupervised domain adaptation in
  semantic segmentation via orthogonal and clustered embeddings,'' in
  \emph{Proceedings of the IEEE/CVF Winter Conference on Applications of
  Computer Vision}, 2021, pp. 1358--1368.

\bibitem{_barbato2021latent}
F.~Barbato, M.~Toldo, U.~Michieli, and P.~Zanuttigh, ``Latent space
  regularization for unsupervised domain adaptation in semantic segmentation,''
  in \emph{Proceedings of the IEEE Conference on Computer Vision and Pattern
  Recognition Workshops}, 2021.

\end{thebibliography}


\begin{thebibliography}{10}
\providecommand{\url}[1]{#1}
\csname url@samestyle\endcsname
\providecommand{\newblock}{\relax}
\providecommand{\bibinfo}[2]{#2}
\providecommand{\BIBentrySTDinterwordspacing}{\spaceskip=0pt\relax}
\providecommand{\BIBentryALTinterwordstretchfactor}{4}
\providecommand{\BIBentryALTinterwordspacing}{\spaceskip=\fontdimen2\font plus
\BIBentryALTinterwordstretchfactor\fontdimen3\font minus
  \fontdimen4\font\relax}
\providecommand{\BIBforeignlanguage}[2]{{%
\expandafter\ifx\csname l@#1\endcsname\relax
\typeout{** WARNING: IEEEtran.bst: No hyphenation pattern has been}%
\typeout{** loaded for the language `#1'. Using the pattern for}%
\typeout{** the default language instead.}%
\else
\language=\csname l@#1\endcsname
\fi
#2}}
\providecommand{\BIBdecl}{\relax}
\BIBdecl

\bibitem{toldo2020unsupervised}
M.~Toldo, A.~Maracani, U.~Michieli, and P.~Zanuttigh, ``Unsupervised domain
  adaptation in semantic segmentation: a review,'' \emph{Technologies}, vol.~8,
  no.~2, 2020.

\bibitem{bengio2013representation}
Y.~Bengio, A.~Courville, and P.~Vincent, ``Representation learning: A review
  and new perspectives,'' \emph{IEEE Transactions on Pattern Analysis and
  Machine Intelligence}, vol.~35, no.~8, pp. 1798--1828, 2013.

\bibitem{girshick2014rich}
R.~Girshick, J.~Donahue, T.~Darrell, and J.~Malik, ``Rich feature hierarchies
  for accurate object detection and semantic segmentation,'' in
  \emph{Proceedings of the IEEE Conference on Computer Vision and Pattern
  Recognition}, 2014, pp. 580--587.

\bibitem{barbato2021latent}
F.~Barbato, M.~Toldo, U.~Michieli, and P.~Zanuttigh, ``Latent space
  regularization for unsupervised domain adaptation in semantic segmentation,''
  in \emph{Proceedings of the IEEE Conference on Computer Vision and Pattern
  Recognition Workshops}, 2021.

\bibitem{Cordts2016}
M.~Cordts, M.~Omran, S.~Ramos, T.~Rehfeld, M.~Enzweiler, R.~Benenson,
  U.~Franke, S.~Roth, and B.~Schiele, ``{The Cityscapes dataset for semantic
  urban scene understanding},'' in \emph{Proceedings of the IEEE Conference on
  Computer Vision and Pattern Recognition}, 2016, pp. 3213--3223.

\bibitem{long2015}
J.~Long, E.~Shelhamer, and T.~Darrell, ``Fully convolutional networks for
  semantic segmentation,'' in \emph{Proceedings of the IEEE Conference on
  Computer Vision and Pattern Recognition}, 2015, pp. 3431--3440.

\bibitem{ronneberger2015u}
O.~Ronneberger, P.~Fischer, and T.~Brox, ``U-net: Convolutional networks for
  biomedical image segmentation,'' in \emph{International Conference on Medical
  image computing and computer-assisted intervention}.\hskip 1em plus 0.5em
  minus 0.4em\relax Springer, 2015, pp. 234--241.

\bibitem{zhao2017}
H.~Zhao, J.~Shi, X.~Qi, X.~Wang, and J.~Jia, ``Pyramid scene parsing network,''
  in \emph{Proceedings of the IEEE Conference on Computer Vision and Pattern
  Recognition}, 2017, pp. 2881--2890.

\bibitem{chen2017rethinking}
L.~Chen, G.~Papandreou, F.~Schroff, and H.~Adam, ``Rethinking atrous
  convolution for semantic image segmentation,'' \emph{arXiv preprint
  arXiv:1706.05587}, 2017.

\bibitem{chen2018deeplab}
L.-C. Chen, G.~Papandreou, I.~Kokkinos, K.~Murphy, and A.~L. Yuille, ``Deeplab:
  Semantic image segmentation with deep convolutional nets, atrous convolution,
  and fully connected crfs,'' \emph{IEEE Transactions on Pattern Analysis and
  Machine Intelligence}, vol.~40, pp. 834--848, 2018.

\bibitem{chen2018encoder}
L.~Chen, Y.~Zhu, G.~Papandreou, F.~Schroff, and H.~Adam, ``Encoder-decoder with
  atrous separable convolution for semantic image segmentation,'' in
  \emph{Proceedings of the European Conference on Computer Vision}, 2018, pp.
  833--851.

\bibitem{neuhold2017mapillary}
G.~Neuhold, T.~Ollmann, S.~Rota~Bulo, and P.~Kontschieder, ``{The Mapillary
  vistas dataset for semantic understanding of street scenes},'' in
  \emph{Proceedings of the International Conference on Computer Vision}, 2017,
  pp. 4990--4999.

\bibitem{varma2019idd}
G.~Varma, A.~Subramanian, A.~Namboodiri, M.~Chandraker, and C.~Jawahar, ``Idd:
  A dataset for exploring problems of autonomous navigation in unconstrained
  environments,'' in \emph{Proceedings of the Winter Conference on Applications
  of Computer Vision}.\hskip 1em plus 0.5em minus 0.4em\relax IEEE, 2019, pp.
  1743--1751.

\bibitem{chen2017nomore}
Y.~Chen, W.~Chen, Y.~Chen, B.~Tsai, Y.~F. Wang, and M.~Sun, ``No more
  discrimination: Cross city adaptation of road scene segmenters,'' in
  \emph{Proceedings of the International Conference on Computer Vision}, 2017,
  pp. 2011--2020.

\bibitem{brostow2009semantic}
G.~J. Brostow, J.~Fauqueur, and R.~Cipolla, ``Semantic object classes in video:
  A high-definition ground truth database,'' \emph{Pattern Recognition
  Letters}, vol.~30, no.~2, pp. 88--97, 2009.

\bibitem{Dosovitskiy17}
A.~Dosovitskiy, G.~Ros, F.~Codevilla, A.~Lopez, and V.~Koltun, ``{CARLA}: {An}
  open urban driving simulator,'' in \emph{Proceedings of the 1st Annual
  Conference on Robot Learning}, 2017, pp. 1--16.

\bibitem{Richter2016}
S.~R. Richter, V.~Vineet, S.~Roth, and V.~Koltun, ``Playing for data: {G}round
  truth from computer games,'' in \emph{Proceedings of the European Conference
  on Computer Vision}, 2016, pp. 102--118.

\bibitem{ros2016}
G.~Ros, L.~Sellart, J.~Materzynska, D.~Vazquez, and A.~M. Lopez, ``The synthia
  dataset: A large collection of synthetic images for semantic segmentation of
  urban scenes,'' in \emph{Proceedings of the IEEE Conference on Computer
  Vision and Pattern Recognition}, 2016, pp. 3234--3243.

\bibitem{chen2019crdoco}
Y.-C. Chen, Y.-Y. Lin, M.-H. Yang, and J.-B. Huang, ``Crdoco: Pixel-level
  domain transfer with cross-domain consistency,'' in \emph{Proceedings of the
  IEEE Conference on Computer Vision and Pattern Recognition}, 2019, pp.
  1791--1800.

\bibitem{hoffman2018}
J.~Hoffman, E.~Tzeng, T.~Park, J.-Y. Zhu, P.~Isola, K.~Saenko, A.~Efros, and
  T.~Darrell, ``Cycada: Cycle-consistent adversarial domain adaptation,'' in
  \emph{Proceedings of the International Conference on Machine Learning}, 2018,
  pp. 1994--2003.

\bibitem{hoffman2016}
J.~Hoffman, D.~Wang, F.~Yu, and T.~Darrell, ``{FCNs in the wild: Pixel-level
  adversarial and constraint-based adaptation},'' \emph{arXiv preprint
  arXiv:1612.02649}, 2016.

\bibitem{MurezKKRK18}
Z.~Murez, S.~Kolouri, D.~J. Kriegman, R.~Ramamoorthi, and K.~Kim, ``Image to
  image translation for domain adaptation,'' in \emph{Proceedings of the IEEE
  Conference on Computer Vision and Pattern Recognition}, 2018, pp. 4500--4509.

\bibitem{toldo2020}
M.~Toldo, U.~Michieli, G.~Agresti, and P.~Zanuttigh, ``Unsupervised domain
  adaptation for mobile semantic segmentation based on cycle consistency and
  feature alignment,'' \emph{Image and Vision Computing}, vol.~95, 2020.

\bibitem{pizzati2020domain}
F.~Pizzati, R.~d. Charette, M.~Zaccaria, and P.~Cerri, ``Domain bridge for
  unpaired image-to-image translation and unsupervised domain adaptation,'' in
  \emph{Proceedings of the Winter Conference on Applications of Computer
  Vision}, 2020, pp. 2990--2998.

\bibitem{du2019ssfdan}
L.~Du, J.~Tan, H.~Yang, J.~Feng, X.~Xue, Q.~Zheng, X.~Ye, and X.~Zhang,
  ``{SSF-DAN:} separated semantic feature based domain adaptation network for
  semantic segmentation,'' in \emph{Proceedings of the International Conference
  on Computer Vision}, 2019, pp. 982--991.

\bibitem{sankaranarayanan2018}
S.~Sankaranarayanan, Y.~Balaji, A.~Jain, S.~Nam~Lim, and R.~Chellappa,
  ``Learning from synthetic data: Addressing domain shift for semantic
  segmentation,'' in \emph{Proceedings of the IEEE Conference on Computer
  Vision and Pattern Recognition}, 2018, pp. 3752--3761.

\bibitem{tsai2018}
Y.-H. Tsai, W.-C. Hung, S.~Schulter, K.~Sohn, M.-H. Yang, and M.~Chandraker,
  ``Learning to adapt structured output space for semantic segmentation,'' in
  \emph{Proceedings of the IEEE Conference on Computer Vision and Pattern
  Recognition}, 2018, pp. 7472--7481.

\bibitem{Tsai2019}
Y.-H. Tsai, K.~Sohn, S.~Schulter, and M.~Chandraker, ``Domain adaptation for
  structured output via discriminative patch representations,'' in
  \emph{Proceedings of the International Conference on Computer Vision}, 2019,
  pp. 1456--1465.

\bibitem{biasetton2019}
M.~Biasetton, U.~Michieli, G.~Agresti, and P.~Zanuttigh, ``{Unsupervised Domain
  Adaptation for Semantic Segmentation of Urban Scenes},'' in \emph{Proceedings
  of the IEEE Conference on Computer Vision and Pattern Recognition Workshops},
  2019, pp. 1211--1220.

\bibitem{michieli2020adversarial}
U.~Michieli, M.~Biasetton, G.~Agresti, and P.~Zanuttigh, ``Adversarial learning
  and self-teaching techniques for domain adaptation in semantic
  segmentation,'' \emph{IEEE Transaction on Intelligent Vehicles}, vol.~5, pp.
  508--518, 2020.

\bibitem{spadotto2020unsupervised}
T.~Spadotto, M.~Toldo, U.~Michieli, and P.~Zanuttigh, ``Unsupervised domain
  adaptation with multiple domain discriminators and adaptive self-training,''
  in \emph{Proceedings of the International Conference on Pattern Recognition},
  2020.

\bibitem{toldo2020clustering}
M.~Toldo, U.~Michieli, and P.~Zanuttigh, ``Unsupervised domain adaptation in
  semantic segmentation via orthogonal and clustered embeddings,'' in
  \emph{Proceedings of the IEEE/CVF Winter Conference on Applications of
  Computer Vision}, 2021, pp. 1358--1368.

\bibitem{Lee2019}
S.~Lee, D.~Kim, N.~Kim, and S.-G. Jeong, ``Drop to adapt: Learning
  discriminative features for unsupervised domain adaptation,'' in
  \emph{Proceedings of the International Conference on Computer Vision}, 2019,
  pp. 91--100.

\bibitem{Park2018}
S.~Park, J.~Park, S.~Shin, and I.~Moon, ``Adversarial dropout for supervised
  and semi-supervised learning,'' in \emph{Proceedings of the {AAAI} Conference
  on Artificial Intelligence}, 2018, pp. 3917--3924.

\bibitem{Saito2018ADR}
K.~Saito, Y.~Ushiku, T.~Harada, and K.~Saenko, ``Adversarial dropout
  regularization,'' in \emph{Proceedings of the International Conference on
  Learning Representations}, 2018.

\bibitem{zou2018}
Y.~Zou, Z.~Yu, B.~Vijaya~Kumar, and J.~Wang, ``Unsupervised domain adaptation
  for semantic segmentation via class-balanced self-training,'' in
  \emph{Proceedings of the European Conference on Computer Vision}, 2018, pp.
  289--305.

\bibitem{Zou2019}
Y.~Zou, Z.~Yu, X.~Liu, B.~V. Kumar, and J.~Wang, ``Confidence regularized
  self-training,'' in \emph{Proceedings of the International Conference on
  Computer Vision}, 2019, pp. 5982--5991.

\bibitem{zhang2017}
Y.~Zhang, P.~David, and B.~Gong, ``Curriculum domain adaptation for semantic
  segmentation of urban scenes,'' in \emph{Proceedings of the International
  Conference on Computer Vision}, 2017, pp. 2020--2030.

\bibitem{zhang2020curriculum}
Y.~Zhang, P.~David, H.~Foroosh, and B.~Gong, ``A curriculum domain adaptation
  approach to the semantic segmentation of urban scenes,'' \emph{IEEE
  Transactions on Pattern Analysis and Machine Intelligence}, 2020.

\bibitem{Chen2019}
M.~Chen, H.~Xue, and D.~Cai, ``Domain adaptation for semantic segmentation with
  maximum squares loss,'' in \emph{Proceedings of the International Conference
  on Computer Vision}, 2019, pp. 2090--2099.

\bibitem{vu2019advent}
T.-H. Vu, H.~Jain, M.~Bucher, M.~Cord, and P.~P{\'e}rez, ``Advent: Adversarial
  entropy minimization for domain adaptation in semantic segmentation,'' in
  \emph{Proceedings of the IEEE Conference on Computer Vision and Pattern
  Recognition}, 2019, pp. 2517--2526.

\bibitem{kang2019contrastive}
G.~Kang, L.~Jiang, Y.~Yang, and A.~G. Hauptmann, ``Contrastive adaptation
  network for unsupervised domain adaptation,'' in \emph{Proceedings of the
  IEEE Conference on Computer Vision and Pattern Recognition}, 2019, pp.
  4893--4902.

\bibitem{tian2020domain}
L.~Tian, Y.~Tang, L.~Hu, Z.~Ren, and W.~Zhang, ``Domain adaptation by class
  centroid matching and local manifold self-learning,'' \emph{arXiv preprint
  arXiv:2003.09391}, 2020.

\bibitem{michieli2021continual}
U.~Michieli and P.~Zanuttigh, ``Continual semantic segmentation via
  repulsion-attraction of sparse and disentangled latent representations,'' in
  \emph{Proceedings of the IEEE Conference on Computer Vision and Pattern
  Recognition}, 2021.

\bibitem{dong2018few}
N.~Dong and E.~P. Xing, ``Few-shot semantic segmentation with prototype
  learning,'' in \emph{Proceedings of the British Machine Vision Conference},
  vol.~3, 2018.

\bibitem{wang2019panet}
K.~Wang, J.~H. Liew, Y.~Zou, D.~Zhou, and J.~Feng, ``Panet: Few-shot image
  semantic segmentation with prototype alignment,'' in \emph{Proceedings of the
  International Conference on Computer Vision}, 2019, pp. 9197--9206.

\bibitem{liang2019distant}
J.~Liang, R.~He, Z.~Sun, and T.~Tan, ``Distant supervised centroid shift: {A}
  simple and efficient approach to visual domain adaptation,'' in
  \emph{Proceedings of the IEEE Conference on Computer Vision and Pattern
  Recognition}, 2019, pp. 2975--2984.

\bibitem{wang2019unsupervised}
Q.~Wang and T.~P. Breckon, ``Unsupervised domain adaptation via structured
  prediction based selective pseudo-labeling,'' in \emph{Proceedings of the
  {AAAI} Conference on Artificial Intelligence}, 2020, pp. 6243--6250.

\bibitem{choi2020role}
H.~Choi, A.~Som, and P.~Turaga, ``Role of orthogonality constraints in
  improving properties of deep networks for image classification,'' \emph{arXiv
  preprint arXiv:2009.10762}, 2020.

\bibitem{pinheiro2018unsupervised}
P.~O. Pinheiro, ``Unsupervised domain adaptation with similarity learning,'' in
  \emph{Proceedings of the IEEE Conference on Computer Vision and Pattern
  Recognition}, 2018, pp. 8004--8013.

\bibitem{wu2019improving}
S.~Wu, J.~Zhong, W.~Cao, R.~Li, Z.~Yu, and H.-S. Wong, ``Improving
  domain-specific classification by collaborative learning with adaptation
  networks,'' in \emph{Proceedings of the AAAI Conference on Artificial
  Intelligence}, 2019, pp. 5450--5457.

\bibitem{790410}
D.~G. {Lowe}, ``Object recognition from local scale-invariant features,'' in
  \emph{Proceedings of the Seventh IEEE International Conference on Computer
  Vision}, vol.~2, 1999, pp. 1150--1157 vol.2.

\bibitem{chen2017no}
Y.-H. Chen, W.-Y. Chen, Y.-T. Chen, B.-C. Tsai, Y.-C. Frank~Wang, and M.~Sun,
  ``No more discrimination: Cross city adaptation of road scene segmenters,''
  in \emph{Proceedings of the IEEE International Conference on Computer
  Vision}, 2017, pp. 1992--2001.

\bibitem{tranheden2020dacs}
W.~Tranheden, V.~Olsson, J.~Pinto, and L.~Svensson, ``Dacs: Domain adaptation
  via cross-domain mixed sampling,'' in \emph{Proceedings of the IEEE/CVF
  Winter Conference on Applications of Computer Vision}, 2021, pp. 1379--1389.

\bibitem{he2016deep}
K.~He, X.~Zhang, S.~Ren, and J.~Sun, ``Deep residual learning for image
  recognition,'' in \emph{Proceedings of the IEEE Conference on Computer Vision
  and Pattern Recognition}, 2016, pp. 770--778.

\bibitem{hendrycks2019benchmarking}
D.~Hendrycks and T.~G. Dietterich, ``Benchmarking neural network robustness to
  common corruptions and surface variations,'' 2019.

\bibitem{li2020spatial}
C.~Li, D.~Du, L.~Zhang, L.~Wen, T.~Luo, Y.~Wu, and P.~Zhu, ``Spatial attention
  pyramid network for unsupervised domain adaptation,'' in \emph{Proceedings of
  the European Conference on Computer Vision}, 2020.

\bibitem{maaten2008visualizing}
L.~Van~der Maaten and G.~Hinton, ``Visualizing data using t-sne.''
  \emph{Journal of machine learning research}, vol.~9, no.~11, 2008.

\bibitem{simonyan2015very}
K.~Simonyan and A.~Zisserman, ``Very deep convolutional networks for
  large-scale image recognition,'' in \emph{Proceedings of the International
  Conference on Learning Representations}, 2015.

\end{thebibliography}
